%% file: acml26_submission_template.tex
\documentclass[wcp]{jmlr}

\usepackage{longtable}

\usepackage{booktabs}

\usepackage{lineno}

\makeatletter
\let\Ginclude@graphics\@org@Ginclude@graphics 
\makeatother

\jmlryear{2026}
\jmlrworkshop{ACML 2026}

\input{math_commands.tex}

\usepackage{amsmath}
\usepackage{amssymb}
\usepackage{mathtools}

\usepackage{url}            
\usepackage{booktabs}       
\usepackage{amsfonts}       
\usepackage{nicefrac}       
\usepackage{microtype}      
\usepackage{xcolor}         
\usepackage{bm}             
\usepackage{graphicx}
\usepackage{algorithm}

\newtheorem{problem}{Problem}[section]

\newtheorem{assumption}{Assumption}[section]

\usepackage{algorithm}
\usepackage{algorithmic}   
\usepackage[normalem]{ulem}

\title[Convergence of GSAM]{Convergence of Sharpness-Aware Minimization Algorithms using Increasing Batch Size and Decaying Learning Rate}

 \author{\Name{Hinata Harada} \Email{haradahinata1011@gmail.com}\\
  \Name{Hideaki Iiduka} \Email{iiduka@cs.meiji.ac.jp}\\
\addr Meiji University, Japan}

\editors{}

\begin{document}

\makeatletter
\let \@jmlrpages \@empty
\makeatother

\maketitle

\begin{abstract}
The sharpness-aware minimization (SAM) algorithm and its variants, including gap guided SAM (GSAM), have been successful at improving the generalization capability of deep neural network models by finding flat local minima of the empirical loss in training. Meanwhile, it has been shown theoretically and practically that increasing the batch size or decaying the learning rate avoids sharp local minima of the empirical loss. In this paper, we consider the GSAM algorithm with increasing batch sizes or decaying learning rates, such as cosine annealing or linear learning rate, and theoretically show its convergence. Moreover, we numerically compare SAM (GSAM) with and without an increasing batch size and conclude that using an increasing batch size { achieves a lower worst-case $\ell_\infty$ adaptive sharpness} than compared with using a constant batch size and learning rate.
\end{abstract}

\section{Introduction}
One way to train a deep neural network (DNN) is to find an optimal parameter $\bm{x}^\star$ of the network in the sense of minimizing the empirical loss $f_S (\bm{x}) = \frac{1}{n} \sum_{i \in [n]} f_i (\bm{x})$ given by the training set $S = (z_1, z_2, \cdots, z_n)$ and a nonconvex loss function $f(\bm{x};z_i) = f_i(\bm{x})$ corresponding to the $i$-th training data $z_i \in S$ ($i \in [n] := \{1,2,\cdots,n\}$). Our main concern is whether a DNN trained by an algorithm for empirical risk minimization (ERM), wherein the empirical loss $f_S$ is minimized, has a strong generalization capability. The sharpness-aware minimization (SAM) problem \citep{foret2021sharpnessaware} was proposed as a way to improve a DNN's generalization capability. The SAM problem is to minimize a perturbed empirical loss defined as the maximum empirical loss $f_{S,\rho} (\bm{x}) := \max_{\|\bm{\epsilon}\| \leq \rho} f_S (\bm{x} + \bm{\epsilon})$ over a certain neighborhood of a parameter $\bm{x} \in \mathbb{R}^d$ of the DNN, where $\rho \geq 0$ and $\bm{\epsilon} \in \mathbb{R}^d$. From the definition of the perturbed empirical loss $f_{S,\rho}$, the SAM problem is specialized to finding flat local minima of the empirical loss $f_S$, which may lead to a better generalization capability than finding sharp minima \citep{keskar2017on,Jiang2020Fantastic}. Although \citep{pmlr-v202-andriushchenko23a} reported that the relationship between sharpness and generalization would be weak, the SAM algorithm and its variants for solving the SAM problem have high generalization capabilities and superior performance, as shown in, e.g., \citep{chen2022when,du2022efficient,andriushchenko2023sharpnessaware,wen2023how,chen2023why,mollenhoff2023sam,wang2024efficient,sherborne2024tram,springer2024sharpnessaware}.

Meanwhile, an algorithm using a large batch size falls into sharp local minima of the empirical loss $f_S$ and the algorithm would experience a drop in generalization performance \citep{NIPS2017_a5e0ff62,goyal2018accuratelargeminibatchsgd,You2020Large}. It has been shown that increasing the batch size \citep{Byrd:2012aa,balles2016coupling,pmlr-v54-de17a,l.2018dont,goyal2018accuratelargeminibatchsgd} or decaying the learning rate \citep{6952943,pmlr-v37-ioffe15,loshchilov2017sgdr,DBLP:journals/corr/abs-1903-09900} avoids sharp local minima of the empirical loss. Hence, we are interested in verifying whether the SAM algorithm with an increasing batch size or decaying learning rate performs well in training DNNs. In this paper, we focus on the SAM algorithm called gap guided SAM (GSAM) algorithm \citep{zhuang2022surrogate} (see Algorithm \ref{algo:1} for details). 
{
GSAM is an interesting method in that it retains the core idea of sharpness-aware optimization in SAM while having a relatively simple structure, and it introduces a decomposition of the gradient into orthogonal and parallel components. Furthermore, previous work \citep{zhuang2022surrogate} has shown that GSAM performs well in practice, making it a promising method from both practical and theoretical perspectives.
}

\textbf{Contribution:} The main contribution of this paper is to show an $\epsilon$-approximation of the GSAM algorithm \textbf{with an increasing batch size and constant learning rate} ((7) in Table \ref{table:1}; Theorem \ref{thm:3_1}) and \textbf{with a constant batch size and decaying learning rate} ((8) in Table \ref{table:1}; Theorem \ref{thm:3_2}).

\begin{table*}[ht]
\centering
\caption{Convergence of SAM and its variants to minimize $\hat{f}_{S,\rho}^{\mathrm{SAM}}(\bm{x}) = f_S (\bm{x}) + \rho \|\nabla f_S (\bm{x})\|$ over the number of steps $T$. ``Noise" in the Gradient column means that algorithm uses noisy observation, i.e., $\bm{g}(\bm{x}) = \nabla f(\bm{x}) + \text{(Noise)}$, of the full gradient $\nabla f(\bm{x})$, while ``Mini-batch" in the Gradient column means that algorithm uses a mini-batch gradient $\nabla f_B (\bm{x}) = \frac{1}{b} \sum_{i\in [b]} \nabla f_{\xi_i} (\bm{x})$ with a batch size $b$. Here, we let $\mathbb{E}[\|\nabla \hat{f}_{S,\rho}^{\mathrm{SAM}*}\|] := \min_{t \in [T]} \mathbb{E}[\|\nabla \hat{f}_{S,\rho}^{\mathrm{SAM}} (\bm{x}_{t})\|]$, where $(\bm{x}_t)_{t=0}^{T}$ is the sequence generated by Algorithm. Results (1)--(6) were presented in
(1) \citep[Theorem 2]{pmlr-v162-andriushchenko22a},
(2) \citep[Theorem 2]{mi2022make},
(3) \citep[Theorem 5.1]{zhuang2022surrogate},
(4) \citep[Theorem 4.6]{si2023practical},
(5) \citep[Corollary 1]{li2023enhancing},
and
(6) \citep[Theorem 2]{li2024friendly}.
{
$^{\dagger}$See Appendix~\ref{sec:rate} for the derivation of the rates and the conditions on $\epsilon$, $\rho$, and $b$.}}\label{table:1}
\begin{tabular}{lllll}
\toprule
Algorithm & Gradient  & Leaning Rate & Perturbation  & Convergence Analysis\\
\midrule
(1) SAM   
& Mini-batch $b$     
& $\eta_T = \Theta(\frac{1}{T^{1/2}})$ 
& $\rho_T = \Theta(\frac{1}{T^{1/4}})$ 
& $\mathbb{E}[\|\nabla f_S^* \|] = O(\frac{1}{T^{1/4}} + \frac{1}{bT^{1/4}} )$ \\
\midrule
(2) SSAM 
& Noise 
& $\eta_t = \Theta(\frac{1}{t^{1/2}} )$
& $\rho_t = \Theta(\frac{1}{t^{1/2}} )$ 
& $\mathbb{E}[\|\nabla f_S^* \|] = O(\frac{\sqrt{\log T}}{T^{1/4}} )$ \\
\midrule
(3) GSAM 
& Noise  
& $\eta_t = \Theta(\frac{1}{t^{1/2}} )$ 
& $\rho_t = \Theta(\frac{1}{t^{1/2}} )$ 
& $\mathbb{E}[\|\nabla \hat{f}_{S,\rho_t}^{\mathrm{SAM}*}\|] = 
O\left(\frac{\sqrt{\log T}}{T^{1/4}} \right)$ \\
\midrule
(4) $m$-SAM 
& Noise  
& $\eta_T = O( \frac{1}{T^{1/2}} )$ 
& $\rho$ 
& $\mathbb{E}[\|\nabla f_S^* \|] = O(\sqrt{\frac{1}{T^{1/2}} + \rho^2} )$ \\
\midrule
(5) VaSSO
& Noise  
& $\eta_T = \Theta(\frac{1}{T^{1/2}} )$ 
& $\rho_T = \Theta(\frac{1}{T^{1/2}} )$ 
& $\mathbb{E}[\|\nabla \hat{f}_{S,\rho}^{\mathrm{SAM}*}\|] = O(\frac{1}{T^{1/4}} )$\\
\midrule
(6) FSAM
& Noise
& $\eta_T = \Theta(\frac{1}{T^{1/2}} )$ 
& $\rho_t = \Theta(\frac{1}{t^{1/2}} )$ 
& $\mathbb{E}[\|\nabla f_S^* \|] = O(\frac{\sqrt{\log T}}{T^{1/4}})$ \\
\midrule
(7) GSAM 
& Increasing   
& Constant 
& $\rho$  
& {$\mathbb{E}[\|\nabla \hat{f}_{S,\rho}^{\mathrm{SAM}*}\|] = O(\frac{1}{T^{1/4}})^{\dagger}$} \\
\quad\textbf{[Ours]}
& mini-batch $b_t$
& $\eta = O(n\epsilon^2)$
& {$= O( \epsilon^4)$}
& \\
\midrule
(8) GSAM 
& Mini-batch $b$   
& Cosine/Linear  
& $\rho$  
& {$\mathbb{E}[\|\nabla \hat{f}_{S,\rho}^{\mathrm{SAM}*}\|] = O(\frac{1}{T^{1/4}})^{\dagger}$} \\
\quad\textbf{[Ours]}
& 
& $\eta_t \to \underline{\eta}$ ($\geq 0$)
& {$= O( \epsilon^4)$}
& \\
\bottomrule
\end{tabular}
\end{table*}
Our convergence analyses of GSAM are based on the search direction noise $\eta_t \bm{\omega}_t$ (defined by (\ref{noise})) between GSAM and gradient descent (GD) (Theorems \ref{thm:1} and \ref{thm:2} in Section \ref{sec:2.3}). The norm of the noise is approximately $\Theta (\frac{\eta_t}{\sqrt{b_t}})$ (see also (\ref{noise_gsam})). Since this implies that GSAM using a large batch size $b$ or a small learning rate $\eta$ behaves approximately the same as GD in solving the SAM problem, GSAM eventually needs to use a large batch size or a small learning rate. 
Accordingly, it will be useful to use increasing batch sizes or decaying learning rates, as the previous results presented in the second paragraph of this section point out.
Our other contribution is to provide numerical results on training ResNets on multiple datasets, in particular showing that using a doubly increasing batch size {achieves lower test error and lower measured sharpness} than using a constant batch size and learning rate (Section \ref{sec:3}).

\textbf{Related work:} Convergence analyses of SGD \citep{robb1951} with a fixed batch size have been presented in \citep{doi:10.1137/120880811,Ghadimi:2016aa,NEURIPS2019_2557911c,feh2020,chen2020,sca2020,loizou2021,wang2021on,Arjevani:2023aa,khaled2022better}. Our analyses found that SGD (an example of GSAM) using increasing batch sizes or a cosine-annealing (linear) learning rate is an $\epsilon$-approximation. The linear scaling rule \citep{goyal2018accuratelargeminibatchsgd,l.2018dont,xie2021a} based on $\frac{\eta}{b}$ coincides with our rule based on the noise norm $\eta \|\bm{\omega}_t\|^2 = \Theta(\frac{\eta_t}{b_t})$. In \citep{pmlr-v48-hazanb16,sato2023using}, it was shown that SGD with an increasing batch size reaches the global optimum under the strong convexity assumption of the smoothed function of $f_S$. This paper shows that, with nonconvex loss functions, GSAM with an increasing batch size achieves an $\epsilon$-approximation. 

\section{SAM problem and GSAM}
\subsection{SAM problem and its approximation problem}
Given a parameter $\bm{x} \in \mathbb{R}^d$ and a data point $z$, a machine-learning model provides a prediction whose quality can be measured by a differentiable nonconvex loss function $f(\bm{x};z)$. For a training set $S = (z_1, z_2, \ldots, z_n)$, $f_i (\cdot) := f(\cdot;z_i)$ is the loss function corresponding to the $i$-th training data $z_i$. The empirical risk minimization (ERM) is to minimize the empirical loss defined for all $\bm{x} \in \mathbb{R}^d$ by 
\begin{align}\label{erm}
f_{S} (\bm{x}) = \frac{1}{n} \sum_{i\in [n]} f(\bm{x};z_i)
= \frac{1}{n} \sum_{i\in [n]} f_i(\bm{x}).
\end{align} 
Given $\rho \geq 0$ and a training set $S$, the SAM problem \citep[(1)]{foret2021sharpnessaware} is to minimize 
\begin{align}\label{prob:sam}
f_{S,\rho}^{\mathrm{SAM}}(\bm{x}) 
:= \max_{\|\bm{\epsilon}\|_2 \leq \rho} f_{S}(\bm{x} + \bm{\epsilon}).
\end{align}
Let $\bm{x}\in \mathbb{R}^d$ and $\rho \geq 0$. Taylor's theorem thus implies that there exists $\tau = \tau (\bm{x},\rho) \in (0,1)$ such that the maximizer $\bm{\epsilon}_{S,\rho}^\star (\bm{x})$ of $f_S (\bm{x} + \bm{\epsilon})$ over $B_2 (\bm{0};\rho) := \{\bm{\epsilon} \in \mathbb{R}^d \colon \|\bm{\epsilon}\|_2 \leq \rho \}$ is as follows:
\begin{align*}
\bm{\epsilon}_{S,\rho}^\star (\bm{x}) 
&:= \argmax_{\|\bm{\epsilon}\|_2 \leq \rho}
f_{S}(\bm{x} + \bm{\epsilon})
= \argmax_{\|\bm{\epsilon}\|_2 \leq \rho} 
\bigg\{ f_S(\bm{x}) + \langle \nabla f_S (\bm{x}), \bm{\epsilon} \rangle_2
+ \frac{1}{2} \langle \bm{\epsilon}, \nabla^2 f_S (\bm{x} + \tau \bm{\epsilon}) \bm{\epsilon} \rangle_2
\bigg\},
\end{align*}
where we suppose that $f_S$ is twice differentiable on $\mathbb{R}^d$. Then, assuming $\|\bm{\epsilon}\|_2^2 \approx 0$ (i.e., a small enough $\rho^2$), $\bm{\epsilon}_{S,\rho}^\star (\bm{x})$ can be approximated as follows $\hat{\bm{\epsilon}}_{S,\rho} (\bm{x})$ \citep[(2)]{foret2021sharpnessaware}:
\begin{align}\label{hat_epsilon}
\bm{\epsilon}_{S,\rho}^\star (\bm{x}) 
\approx 
\hat{\bm{\epsilon}}_{S,\rho} (\bm{x})
:=
\argmax_{\|\bm{\epsilon}\|_2 \leq \rho} \langle \nabla f_S (\bm{x}), \bm{\epsilon} \rangle_2
=  
\begin{cases}
\left\{ \rho \frac{\nabla f_S (\bm{x})}{\|\nabla f_S (\bm{x})\|_2} \right\} &\text{ }(\nabla f_S (\bm{x}) \neq \bm{0})\\
B_2 (\bm{0};\rho) &\text{ }(\nabla f_S (\bm{x}) = \bm{0}).
\end{cases} 
\end{align}
Here, our goal is to solve the following approximation of the SAM problem of minimizing $f_{S,\rho}^{\mathrm{SAM}} (\bm{x}) = \max_{\|\bm{\epsilon}\|_2 \leq \rho} f_{S}(\bm{x} + \bm{\epsilon})$ (see (\ref{prob:sam}) and (\ref{hat_epsilon})).

\begin{problem}[
  Approximated SAM problem \citep{foret2021sharpnessaware}
]
\label{prob:1}
Let $f_S$ be the empirical loss defined by (\ref{erm}) with the training set $S = (z_1,z_2,\cdots,z_n)$. Given $\rho \geq 0$,
\begin{align*}
\text{minimize } 
\hat{f}_{S,\rho}^{\mathrm{SAM}} (\bm{x})
&:= 
\max_{\|\bm{\epsilon}\|_2 \leq \rho} 
\left\{ f_{S}(\bm{x}) 
+ \langle \nabla f_S (\bm{x}), \bm{\epsilon} \rangle_2
\right\}= 
f_{S}(\bm{x}) + \rho \|\nabla f_S (\bm{x})\|_2
\text{ subject to } \bm{x} \in \mathbb{R}^d.
\end{align*}
\end{problem}
We use the following approximation \citep[(3)]{foret2021sharpnessaware} of the gradient of $\hat{f}_{S,\rho}^{\mathrm{SAM}}$ at $\bm{x} \in \mathbb{R}^d$:
\begin{align}\label{sam_s}
\nabla \hat{f}_{S,\rho}^{\mathrm{SAM}}(\bm{x})
&:= 
\nabla f_{S}(\bm{x})|_{\bm{x} + \hat{\bm{\epsilon}}_{S,\rho} (\bm{x})}
= 
\begin{cases}
\nabla f_{S} \left(\bm{x} + \rho \frac{\nabla f_{S} (\bm{x})}{\|\nabla f_{S} (\bm{x})\|_2} \right) &\text{ }(\nabla f_{S} (\bm{x}) \neq \bm{0})\\
\nabla f_{S} \left(\bm{x} + \bm{u} \right) &\text{ }(\nabla f_{S} (\bm{x}) = \bm{0}),
\end{cases}
\end{align}
where $\hat{\bm{\epsilon}}_{S,\rho}(\bm{x})$ is denoted by (\ref{hat_epsilon}) and $\bm{u}$ is an arbitrary point in $B_2 (\bm{0};\rho)$ (e.g., we may set $\bm{u} = \bm{0}$ before implementing algorithms).

\subsection{Mini-batch GSAM algorithm}
As a way of solving Problem \ref{prob:1}, we will study the GSAM algorithm \citep[Algorithm 1]{zhuang2022surrogate} using $b$ loss functions $f_{\xi_{t,1}}, f_{\xi_{t,2}}, \cdots, f_{\xi_{t,b}} \in \{f_1,f_2,\cdots,f_n\}$ that are randomly chosen at each time $t$, where $b$ is a batch size satisfying $b \leq n$. We suppose that the loss functions satisfy the following conditions.

\begin{assumption}\label{assum:1}
{\em (A1)} $f_i \colon \mathbb{R}^d \to \mathbb{R}$ ($i\in [n]$) is twice differentiable and $L_i$-smooth.

{\em (A2)} $\nabla f_{\xi} \colon \mathbb{R}^d \to \mathbb{R}^d$ is the stochastic gradient of $\nabla f_S$; i.e., {\em (i)} for all $\bm{x} \in \mathbb{R}^d$, $\mathbb{E}_{\xi}[\nabla f_{\xi}(\bm{x})] = \nabla f_S(\bm{x})$, {\em (ii)} there exists $\sigma \geq 0$ such that, for all $\bm{x} \in \mathbb{R}^d$, $\mathbb{V}_{\xi}[\nabla f_{\xi}(\bm{x})] = \mathbb{E}_{\xi}[\| \nabla f_{\xi}(\bm{x}) - \nabla f_S(\bm{x})\|_2^2] \leq \sigma^2$, where $\xi$ is a random variable which is independent of $\bm{x}$ and $\mathbb{E}_\xi[\cdot]$ stands for the expectation with respect to $\xi$. 

{\em (A3)} Let $t\in \mathbb{N}$ and suppose that $b_t \in \mathbb{N}$ and $b_t \leq n$. Let $\bm{\xi}_t = (\xi_{t,1}, \xi_{t,2}, \cdots, \xi_{t,b_t})^\top$ be a random variable {that consists of $b_t$ variables sampled uniformly at random without replacement from $[n]$.} The full gradient $\nabla f_S (\bm{x})$ is estimated as the following mini-batch gradient at $\bm{x} \in \mathbb{R}^d$:
\begin{align}\label{minibatch_gradeint}
\nabla f_{S_t}(\bm{x}) := \frac{1}{b_t} \sum_{i\in [b_t]} \nabla f_{\xi_{t,i}}(\bm{x}),
\end{align}
where $\bm{\xi}_t$ is independent of $\bm{x}$, $b_t$, and $\bm{\xi}_{t'}$ ($t \neq t'$). 
\end{assumption}

We define $\hat{\bm{\epsilon}}_{S_t,\rho}$ by replacing $S$ in (\ref{hat_epsilon}) with $S_t$ in (A3), i.e.,
\begin{align}\label{hat_epsilon_t} 
\hat{\bm{\epsilon}}_{S_t, \rho} (\bm{x})
&:=
\argmax_{\|\bm{\epsilon}\|_2 \leq \rho} \langle \nabla f_{S_t} (\bm{x}), \bm{\epsilon} \rangle_2
=  
\begin{cases}
\left\{ \rho \frac{\nabla f_{S_t} (\bm{x})}{\|\nabla f_{S_t} (\bm{x})\|_2} \right\} &\text{ }(\nabla f_{S_t} (\bm{x}) \neq \bm{0})\\
B_2 (\bm{0};\rho) &\text{ }(\nabla f_{S_t} (\bm{x}) = \bm{0}),
\end{cases}
\end{align}
where $\nabla f_{S_t}$ is defined as in (\ref{minibatch_gradeint}). Accordingly, an approximation of a mini-batch gradient of $\hat{f}_{S,\rho}^{\mathrm{SAM}}$ (see Problem \ref{prob:1} and (\ref{sam_s})) at $\bm{x} \in \mathbb{R}^d$ can be defined as
\begin{align}\label{sam_s_t}
\nabla \hat{f}_{S_t,\rho}^{\mathrm{SAM}}(\bm{x})
&:= 
\nabla f_{S_t}(\bm{x})|_{\bm{x} + \hat{\bm{\epsilon}}_{S_t,\rho} (\bm{x})}
= 
\begin{cases}
\nabla f_{S_t} \left(\bm{x} + \rho \frac{\nabla f_{S_t} (\bm{x})}{\|\nabla f_{S_t} (\bm{x})\|_2} \right) &\text{ }(\nabla f_{S_t} (\bm{x}) \neq \bm{0})\\
\nabla f_{S_t} \left(\bm{x} + \bm{u} \right) &\text{ }(\nabla f_{S_t} (\bm{x}) = \bm{0}),
\end{cases}
\end{align}
where $\hat{\bm{\epsilon}}_{S_t,\rho} (\bm{x})$ is denoted by (\ref{hat_epsilon_t}) and $\bm{u}$ is an arbitrary point in $B_2 (\bm{0};\rho)$. Accordingly, the SAM algorithm \citep[Algorithm 1]{foret2021sharpnessaware} can be obtained by applying SGD to the objective function $\hat{f}_{S,\rho}^{\mathrm{SAM}}$ in Problem \ref{prob:1}, as described in Algorithm \ref{algo:1}. GD for Problem \ref{prob:1} coincides with Algorithm \ref{algo:1} with $S_t = S$ (i.e., $b_t = n$), as follows:
\begin{align}\label{gd}
\bm{x}_{t+1} 
:= \bm{x}_t - \eta_t \nabla \hat{f}_{S,\rho}^{\mathrm{SAM}}(\bm{x}_t),
\end{align}
where $\nabla \hat{f}_{S,\rho}^{\mathrm{SAM}}$ is defined as in (\ref{sam_s}). The GSAM algorithm uses an ascent step in the orthogonal direction that is obtained by using stochastic gradient decomposition $\nabla f_{S_t}(\bm{x}) = \nabla f_{S_t \parallel}(\bm{x}) + \nabla f_{S_t \perp}(\bm{x})$ to minimize a surrogate gap $h_t (\bm{x}) := \hat{f}^{\mathrm{SAM}}_{S_t, \rho}(\bm{x}) - f_{S_t}(\bm{x})$ (see \citep[Section 4]{zhuang2022surrogate}). 

\begin{algorithm*}
\caption{Mini-batch GSAM algorithm}
\label{algo:1}
\begin{algorithmic}
\REQUIRE
$\rho \geq 0$ (hyperparameter),
$\bm{u} \in B_2 (\bm{0};\rho)$,
$\bm{x}_0 \in \mathbb{R}^d$ (initial point),
$b_t > 0$ (batch size),
$\eta_t > 0$ (learning rate),
$\alpha \in \mathbb{R}$ (control parameter of ascent step),
$T \geq 1$ (steps)
\ENSURE
$(\bm{x}_t)_{t=0}^{T} \subset \mathbb{R}^d$
\FOR{$t=0,1,\ldots,T-1$}
\STATE
$\nabla \hat{f}_{S_t,\rho}^{\mathrm{SAM}}(\bm{x}_t)
:=
\begin{cases}
\nabla f_{S_t} \left(\bm{x}_t + \rho \frac{\nabla f_{S_t} (\bm{x}_t)}{\|\nabla f_{S_t} (\bm{x}_t)\|_2} \right) &\text{ }(\nabla f_{S_t} (\bm{x}_t) \neq \bm{0})\\
\nabla f_{S_t} (\bm{x}_t + \bm{u}) &\text{ }(\nabla f_{S_t} (\bm{x}_t) = \bm{0})
\end{cases}
\triangleleft \text{See } (\ref{minibatch_gradeint}) \text{ for }
\nabla f_{S_t}$
\STATE
$\bm{d}_t :=
\begin{cases}
- (\nabla \hat{f}_{S_t,\rho}^{\mathrm{SAM}}(\bm{x}_t) - \alpha \nabla f_{S_t \perp}(\bm{x}_t)) &\text{ (GSAM)}\\
- \nabla \hat{f}_{S_t,\rho}^{\mathrm{SAM}}(\bm{x}_t) &\text{ (SAM; $\alpha = 0$)}\\
- \nabla \hat{f}_{S_t,0}^{\mathrm{SAM}}(\bm{x}_t)
= - \nabla f_{S_t} (\bm{x}_t) &\text{ (SGD; $\alpha = \rho = 0$)}
\end{cases}$
\STATE
$\bm{x}_{t+1}
:= \bm{x}_t + \eta_t \bm{d}_t$
\ENDFOR
\end{algorithmic}
\end{algorithm*}

\subsection{Search direction noise between GSAM and GD}\label{sec:2.3}
GSAM can find local minima of Problem \ref{prob:1} (by using $- \nabla \hat{f}_{S_t,\rho}^{\mathrm{SAM}}(\bm{x}_t)$) that are flatter local minima of $f_{S}$ (by using $\alpha \nabla f_{S_t \perp}(\bm{x}_t)$) (see \citep[Section 4]{zhuang2022surrogate} for details). Meanwhile, GD defined as (\ref{gd}) (i.e., GSAM with $b_t = n$ and $\alpha = 0$) is the simplest algorithm for solving Problem \ref{prob:1}. Although this GD can minimize $\hat{f}_{S,\rho}^{\mathrm{SAM}}$ by using the full gradient $\nabla \hat{f}_{S,\rho}^{\mathrm{SAM}}(\bm{x}_t)$, it is not guaranteed that it converges to a flatter minimum of $f_S$ compared with the one of GSAM. Here, let us compare GSAM with GD. Let $\bm{x}_t \in \mathbb{R}^d$ be the $t$-th approximation of Problem \ref{prob:1} and $\eta_t > 0$. The $\bm{x}_{t+1}$ generated by GSAM is as follows:
\begin{align}\label{noise}
\bm{x}_{t+1} 
&= \bm{x}_t + \eta_t \{- (\nabla \hat{f}_{S_t,\rho}^{\mathrm{SAM}}(\bm{x}_t) - \alpha \nabla f_{S_t \perp}(\bm{x}_t))\}\\
&= \underbrace{\bm{x}_t - \eta_t \nabla \hat{f}_{S,\rho}^{\mathrm{SAM}}(\bm{x}_t)}_{\text{GD}} \nonumber
+ \underbrace{\eta_t (
\overbrace{\nabla \hat{f}_{S,\rho}^{\mathrm{SAM}}(\bm{x}_t)
- \nabla \hat{f}_{S_t,\rho}^{\mathrm{SAM}}(\bm{x}_t)}^{\hat{\bm{\omega}}_t} + \alpha \nabla f_{S_t \perp}(\bm{x}_t))}_{\text{Search Direction Noise } \eta_t \bm{\omega}_t} \nonumber
\end{align}
This implies that, if $\eta_t \bm{\omega}_t := \eta_t (\nabla \hat{f}_{S,\rho}^{\mathrm{SAM}}(\bm{x}_t) - \nabla \hat{f}_{S_t,\rho}^{\mathrm{SAM}}(\bm{x}_t) + \alpha \nabla f_{S_t \perp}(\bm{x}_t))$ is approximately zero, i.e., $b_t \approx n$ and $\alpha \approx 0$, then GSAM is approximately GD in the sense of the norm of $\mathbb{R}^d$, and if $\eta_t \bm{\omega}_t$ is not zero under $\alpha \neq 0$, i.e., $b_t < n$, then the behavior of GSAM with $b_t < n$ differs from the one of GD. We call $\eta_t \bm{\omega}_t$ the {\em search direction noise} of GSAM, since $\eta_t \bm{\omega}_t$ is noise from the viewpoint of the search direction of GD. Theorem \ref{thm:1} below provides an upper bound on the norm of the search direction noise of GSAM (see Appendix \ref{appendix_a} for the proof).

\begin{theorem}
[Upper bound of $\mathbb{E}\eta_t \|\bm{\omega}_t\|_2$]
\label{thm:1}
Suppose that Assumption \ref{assum:1} holds and define $\bm{\omega}_t \in \mathbb{R}^d$ for all $t \in \mathbb{N} \cup \{0\}$ by $\bm{\omega}_t := \hat{\bm{\omega}}_t + \alpha \nabla f_{S_t \perp}(\bm{x}_t)$, where $\bm{x}_t$ is generated by Algorithm \ref{algo:1} and we assume that $G_{\perp} := \sup_{t\in \mathbb{N}\cup \{0\}}\|\nabla f_{S_t \perp}(\bm{x}_t)\|_2 < + \infty$. Then, for all $t \in \mathbb{N} \cup \{0\}$, {if $b_t = n$, $\mathbb{E}[\eta_t \|\bm{\omega}_t\|_2] \leq \eta_t |\alpha| G_{\perp}$, and if $b_t < n$, }
\begin{align*}
\mathbb{E}[\eta_t \|\bm{\omega}_t\|_2]
\leq  
\eta_t \Bigg\{
\sqrt{4 \rho^2 \left( \frac{1}{b_t^2} + \frac{1}{n^2} \right)
\big(\sum_{i \in [n]} L_i \big)^2 
+ \frac{2 \sigma^2}{b_t}}
+ |\alpha| G_{\perp} 
\Bigg\},
\end{align*}
where $\mathbb{E}[\cdot]$ stands for the total expectation defined by $\mathbb{E} = \mathbb{E}_{\bm{\xi}_0} \mathbb{E}_{\bm{\xi}_1} \cdots \mathbb{E}_{\bm{\xi}_t}$. 
\end{theorem}

In the case of GSAM with $b_t = n$ and $\alpha \neq 0$, we have that $\eta_t \bm{\omega}_t = \eta_t (\nabla \hat{f}_{S,\rho}^{\mathrm{SAM}}(\bm{x}_t) - \nabla \hat{f}_{S,\rho}^{\mathrm{SAM}}(\bm{x}_t) + \alpha \nabla f_{S \perp}(\bm{x}_t)) = \eta_t \alpha \nabla f_{S \perp}(\bm{x}_t)$. Hence, an upper bound of $\mathbb{E}[\eta_t \|\bm{\omega}_t\|_2]$ is $\eta_t |\alpha| G_{\perp}$ (Theorem \ref{thm:1} ($b_t = n$)). For simplicity, let us consider the case of $\alpha = 0$. The search direction noise $\eta_t \bm{\omega}_t$ of GSAM with $b_t < n$ is not zero, from $\nabla \hat{f}_{S,\rho}^{\mathrm{SAM}}(\bm{x}_t) \neq \nabla \hat{f}_{S_t,\rho}^{\mathrm{SAM}}(\bm{x}_t)$ (see (\ref{noise})). Meanwhile, the search direction noise $\eta_t \bm{\omega}_t$ of GD (GSAM with $b_t = n$ and $\alpha = 0$) is $\eta_t \bm{\omega}_t = \eta_t (\nabla \hat{f}_{S,\rho}^{\mathrm{SAM}}(\bm{x}_t) - \nabla \hat{f}_{S,\rho}^{\mathrm{SAM}}(\bm{x}_t)) = \bm{0}$, which implies that $\mathbb{E}[\eta_t \|\bm{\omega}_t\|_2] = 0$ (This result coincides with Theorem \ref{thm:1} ($b_t = n$ and $\alpha = 0$)). Accordingly, the noise norm $\mathbb{E}[\eta_t \|\bm{\omega}_t\|_2]$ of GSAM will decrease as the batch size $b_t$ increases. In fact, from Theorem \ref{thm:1} ($b_t < n$), the upper bound $U(\eta_t, b_t)$ of $\mathbb{E}[\eta_t \|\bm{\omega}_t\|_2]$,
\begin{align*}
\mathbb{E}[\eta_t \|\bm{\omega}_t\|_2]
&\leq
\eta_t
\sqrt{4 \rho^2 \left( \frac{1}{b_t^2} + \frac{1}{n^2} \right)
\big(\sum_{i \in [n]} L_i \big)^2 
+ \frac{2 \sigma^2}{b_t}}
\leq 
\eta_t \frac{\sqrt{8\rho^2 (\sum_{i\in [n]} L_i)^2 + 2 \sigma^2}}{\sqrt{b_t}}
=: U (\eta_t, b_t),
\end{align*}
is a monotone decreasing function of $b_t$. As a result, $\mathbb{E}[\eta_t \|\bm{\omega}_t\|_2]$ decreases as $b_t$ increases. Theorem \ref{thm:1} also indicates that the smaller $\eta_t$ is, the smaller $\mathbb{E}[\eta_t \|\bm{\omega}_t\|_2]$ becomes. 

Next, we provide a lower bound of the norm of the search direction noise of GSAM. Theorem \ref{thm:2} is proven in Appendix \ref{appendix_a}.

\begin{theorem}
[Lower bound of $\mathbb{E}\eta_t \|\bm{\omega}_t\|_2$]
\label{thm:2}
Under the assumptions in Theorem \ref{thm:1}, for all $t \in \mathbb{N} \cup \{0\}$, {if $b_t = n$, $\mathbb{E}[\eta_t \|\bm{\omega}_t\|_2] \geq \eta_t |\alpha| \mathbb{E}[\|\nabla f_{S \perp}(\bm{x}_t)\|_2]$, and if $b_t < n \land A_t \geq 0$, }
\begin{align*}
\mathbb{E}[\eta_t \|\bm{\omega}_t\|_2]
\geq 
\eta_t \left\{ \frac{c_t \sigma}{\sqrt{b_t}}
- \rho \left(\frac{1}{b_t} + \frac{1}{n}  \right) \sum_{i\in [n]} L_i
- |\alpha| G_{\perp} \right\},
\end{align*}
{and if $b_t < n \land A_t < 0$, then}
\begin{align*}
\mathbb{E}[\eta_t \|\bm{\omega}_t\|_2]
\geq
\eta_t \left\{ \rho \left(\frac{d_t}{b_t} - \frac{1}{n}  \right) \sum_{i\in [n]} L_i
- \frac{\sigma}{\sqrt{b_t}}
- |\alpha| G_{\perp} \right\},
\end{align*}
where $A_t$ is defined by (\ref{key:2_3}), $c_t, d_t \in (0,1]$, and $|\alpha|$ is small such that, for $b_t < n$,\\  $|\alpha| \|\nabla f_{S_t \perp}(\bm{x}_t)\|_2 \leq \|\hat{\bm{\omega}}_t\|_2$. 
\end{theorem}

From the definition (\ref{noise}) of the search direction noise, the noise norm $\mathbb{E}[\eta_t \|\bm{\omega}_t\|_2]$ of GSAM will increase as the batch size $b_t$ decreases. We can verify this fact from Theorem \ref{thm:2} ($b_t < n \land A_t \geq 0$). For simplicity, let us consider the case where $\alpha = 0$. We set $T \geq 1$, $c := \min_{t\in [0:T]} c_t$, and $\rho \leq \frac{c \sigma}{2\sum_{i\in[n]} L_i}$ (this setting implies that $\rho$, which is used in the definition of Problem \ref{prob:1}, will be a small parameter (see also (\ref{hat_epsilon})). Then, the lower bound $L (\eta_t, b_t)$ of $\mathbb{E}[\eta_t \|\bm{\omega}_t\|_2]$ satisfies 
\begin{align*}
\mathbb{E}[\eta_t \|\bm{\omega}_t\|_2]
&\geq
\eta_t \left\{ \frac{c_t \sigma}{\sqrt{b_t}}
- \rho \left(\frac{1}{b_t} + \frac{1}{n}  \right) \sum_{i\in [n]} L_i
\right\}
\geq 
\eta_t \frac{c_t \sigma - 2 \rho \sum_{i\in [n]} L_i}{\sqrt{b_t}} =: L (\eta_t, b_t) 
\text{ } (\geq 0),
\end{align*}
which implies that the smaller $b_t$ is, the larger the lower bound $L(\eta_t, b_t)$ of $\mathbb{E}[\eta_t \|\bm{\omega}_t\|_2]$ becomes (We can verify this result from Theorem \ref{thm:2} ($b_t < n \land A_t < 0$)). Therefore, $\mathbb{E}[\eta_t \|\bm{\omega}_t\|_2]$ increases as $b_t$ decreases. 

To solve Problem \ref{prob:1}, we consider a mini-batch scheduler and a learning rate scheduler based on Theorems \ref{thm:1} and \ref{thm:2}. To apply not only GSAM but also SAM ($\alpha = 0$) to Problem \ref{prob:1}, we will assume that $|\alpha|$ is approximately zero. Theorems \ref{thm:1} and \ref{thm:2} (see also the definitions of $U (\eta_t, b_t)$ and $L (\eta_t, b_t)$) indicate that, for a given small $\rho$ and for all $t \in \mathbb{N} \cup \{0\}$,
\begin{align}\label{noise_gsam}
\mathbb{E}[\eta_t \|\bm{\omega}_t\|_2]
&\approx 
\mathbb{E}\left[\eta_t \left\|\nabla \hat{f}_{S,\rho}^{\mathrm{SAM}}(\bm{x}_t)
- \nabla \hat{f}_{S_t,\rho}^{\mathrm{SAM}}(\bm{x}_t) \right\|_2 \right]
\approx
\begin{cases}
\Theta \left( \frac{\eta_t}{\sqrt{b_t}} \right) &\text{ } (b_t < n) \\
0 &\text{ } (b_t = n).
\end{cases} 
\end{align}
Equation (\ref{noise_gsam}) indicates that the full gradient $\nabla \hat{f}_{S,\rho}^{\mathrm{SAM}}(\bm{x}_0)$ substantially differs from $\nabla \hat{f}_{S_0,\rho}^{\mathrm{SAM}}(\bm{x}_0)$ with a small batch size $b_0$ or a large learning rate $\eta_0$. Meanwhile, GSAM eventually needs to use a large batch size $b$ or a small learning rate $\eta$, since the behavior of GSAM using a large $b$ or small $\eta$ is approximately like that of GD in minimizing $\hat{f}_{S,\rho}^{\mathrm{SAM}}$. Accordingly, it would be useful to use increasing batch sizes or decaying learning rates in the process of training DNNs.

\subsection{Convergence analysis of GSAM}\label{sec:2.4}
\subsubsection{Increasing batch size and constant learning rate}
Motivated by \citep{l.2018dont}, we focus on using a constant learning rate defined for all $t\in \mathbb{N} \cup \{0\}$ by $\eta_t = \eta \in (0, + \infty)$ and a mini-batch scheduler that gradually increases the batch size:
\begin{align}\label{mini-batch}
\begin{split}
\underbrace{b_0 = \cdots = b_0}_{E_0 \text{ epochs}} 
\leq 
\underbrace{b_1 = \cdots = b_1}_{E_1 \text{ epochs}} 
\leq \cdots \leq 
\underbrace{b_M = \cdots = b_M = n}_{E_M \text{ epochs}},
\end{split}
\end{align}
where $M \in \mathbb{N}$ and $E_i \in \mathbb{N}$ ($i\in [0:M]$). Accordingly, we have that the total number of steps for training is $T = \sum_{i \in [0:M]} \lceil \frac{n}{b_i} \rceil E_i$.

Theorem \ref{thm:1} leads us to the following theorem, the proof of which is given in Appendix \ref{proof_thm_3_1}.

\begin{theorem}
\label{thm:3_1}
\textup{\textbf{($\epsilon$--approximation of GSAM with an increasing batch size and constant learning rate)}}
Consider the sequence $(\bm{x}_t)$ generated by the mini-batch GSAM algorithm (Algorithm \ref{algo:1}) with an increasing batch size $b_t \in (0,n]$ defined by (\ref{mini-batch}) and a constant learning rate, $\eta_t = \eta \in (0, +\infty)$. Furthermore, let us assume that there exists a positive number $G$ such that $\max \{\sup_{t\in \mathbb{N} \cup \{0\}} \|\nabla f_S (\bm{x}_{t} + \hat{\bm{\epsilon}}_{S_{t},\rho}(\bm{x}_{t}))\|_2, \sup_{t\in \mathbb{N} \cup \{0\}} \|\nabla \hat{f}^{\mathrm{SAM}}_{S_t,\rho} (\bm{x}_t)\|_2, \sup_{t\in \mathbb{N} \cup \{0\}} \\ \|\nabla \hat{f}^{\mathrm{SAM}}_{S,\rho} (\bm{x}_t)\|_2, G_{\perp} \} \leq G$, where $G_{\perp} := \sup_{t\in \mathbb{N}\cup \{0\}}\|\nabla f_{S_t \perp}(\bm{x}_t)\|_2 < + \infty$ (see Theorem \ref{thm:1}). Let $\epsilon > 0$ be the precision and let $b_0 > 0$, $\eta > 0$, $\alpha \in \mathbb{R}$, and $\rho \geq 0$ such that 
{
\begin{align}
&\eta
\in
\Bigg[
\frac{12 \sigma \rho C}{\epsilon^2}
\left(
\frac{G}{\sqrt{b_0}}
+ \frac{3 \sigma \rho C}{n b_0} \sum_{i\in [n]} L_i
\right),\
\frac{(|\alpha| + 1)^{-2} n^3 \epsilon^2}
{6 G^2 \sum_{i\in [n]} L_i \left\{ n^2 + 4 \rho^2 C^2 \big( \sum_{i\in [n]} L_i \big)^2 \right\}}
\Bigg], \label{eta_2}\\
&\rho (|\alpha| + 1) 
\leq 
\frac{n \sqrt{b_0} \epsilon^2}{6G(CG \sqrt{b_0} + B \sigma)\sum_{i\in [n]} L_i}, \text{ }
\rho
\leq 
\frac{n b_0 \epsilon^2}{24 G \sqrt{n^2 + b_0^2}\sum_{i \in [n]} L_i}, \label{eta_3}
\end{align}
}
where $B$ and $C$ are nonnegative constants. Then, there exists $t_0 \in \mathbb{N}$ such that, for all $T \geq t_0$, 
\begin{align*}
&\min_{t \in [0:T-1]}\mathbb{E}\left[ \left\|\nabla \hat{f}_{S,\rho}^{\mathrm{SAM}}(\bm{x}_t) \right\|_2 \right]
\leq \epsilon.
\end{align*}
\end{theorem} 

Theorem \ref{thm:3_1} indicates that the parameters $|\alpha|$ and $\rho$ in (\ref{eta_3}) become small and thereby achieve an $\epsilon$--approximation of GSAM. The setting of the small parameter $\rho$ is consistent with the definition of Problem \ref{prob:1} (see also (\ref{hat_epsilon})). Moreover, the setting also matches the numerical results in \citep{zhuang2022surrogate} that used small $|\alpha|$ and $\rho$. { Since C appears in (\ref{eta_2}) only through $\rho C$, using a small $\rho$ makes the lower bound of $\eta$ in (\ref{eta_2}) approximately zero and term 4 $\rho^2 C^2 \big( \sum_{i\in [n]} L_i \big)^2$ in the upper bound approximately zero. Hence, a constant learning rate $\eta$ satisfying (\ref{eta_2}) is approximately
\begin{align}\label{eta_small}
\eta 
\in
\left(0,  
\frac{n \epsilon^2}{6(|\alpha| + 1)^2 G^2 \sum_{i\in [n]} L_i}
\right].
\end{align}
From (\ref{eta_small}), it would be appropriate to set a small $\eta$ in order to achieve an $\epsilon$-approximation of GSAM. In fact, the numerical results in \citep{zhuang2022surrogate} used small learning rates, such as $10^{-2}$, $10^{-3}$, and $10^{-5}$. 

Since SGD (i.e., GSAM with $\alpha = \rho=0$) satisfies (\ref{eta_3}), Theorem \ref{thm:3_1} guarantees that SGD is an $\epsilon$-approximation in the sense of  $\min_{t\in [0:T-1]} \mathbb{E}[\|\nabla f_S (\bm{x}_t)\|_2] \leq \epsilon$. Moreover, using $\alpha = \rho = 0$ makes the upper bound of $\min_{t\in [0:T-1]} \mathbb{E}[\|\nabla f_S (\bm{x}_t)\|_2]$ ($= \min_{t\in [0:T-1]} \mathbb{E}[\|\nabla \hat{f}_{S,\rho}^{\text{SAM}} (\bm{x}_t)\|_2]$) smaller than using $\alpha \neq 0 \lor \rho \neq 0$. Hence, SGD using $\alpha = \rho = 0$ would minimize $\|\nabla f_S (\bm{x}_t)\|_2$ more quickly than would SAM/GSAM using $\alpha \neq 0 \lor \rho \neq 0$ (see also Figure \ref{fig3} (Left) indicating that SGD minimizes $f_S$ more quickly than SAM/GSAM). Meanwhile, the previous results in \citep{foret2021sharpnessaware,zhuang2022surrogate} indicate that using $\alpha \neq 0 \lor \rho \neq 0$ leads to a better generalization than using $\alpha = \rho = 0$ (see Figure \ref{fig3} (Right) and Tables \ref{table:CIFAR and T-ImageNet} and \ref{table:ImageNet} indicating that SAM/GSAM with an increasing batch size has a higher generalization capability than SGD has with an increasing batch size).

\subsubsection{Constant batch size and decaying learning rate}
Motivated by \citep{loshchilov2017sgdr}, we focus on a constant batch size defined for all $t \in \mathbb{N} \cup \{0\}$ by $b_t = b$ and examine a cosine-annealing rate scheduler defined by 
\begin{align}\label{cosine_lr}
\eta_t = 
\underline{\eta} + \frac{\overline{\eta} - \underline{\eta}}{2} \left( 1 + \cos \left\lfloor \frac{t}{K} \right\rfloor \frac{\pi}{E} \right)
\quad (t \in [0 : KE]),
\end{align}
where $\underline{\eta}$ and $\overline{\eta}$ are such that $0 \leq \underline{\eta} \leq \overline{\eta}$, $E$ is the number of epochs, and $K = \lceil \frac{n}{b} \rceil$ is the number of steps per epoch. Accordingly, we have that the total number of steps for training is $T = KE$. The cosine-annealing learning rate (\ref{cosine_lr}) is updated per epoch and remains unchanged during $K$ steps.

Moreover, for a constant batch size $b_t = b$ ($t \in \mathbb{N} \cup \{0\}$), we examine a linear learning rate scheduler \citep{Liu2020On} defined by 
\begin{align}\label{linear_lr}
\eta_t 
= 
\frac{\underline{\eta} - \overline{\eta}}{T} t  + \overline{\eta} \quad (t \in [0 :T]),
\end{align}
where $\underline{\eta}$ and $\overline{\eta}$ are such that $0 \leq \underline{\eta} \leq \overline{\eta}$ and $T$ is the number of steps. The linear learning rate scheduler (\ref{linear_lr}) is updated with a linearly decaying step size from $0$ to $T$.

Theorem \ref{thm:1} leads us to the following theorem, the proof which is given in Appendix \ref{proof_thm_3_2} (The case of $\underline{\eta} > 0$ is shown in Appendix \ref{proof_thm_3_2}).

\begin{theorem}
\label{thm:3_2}
\textup{\textbf{($\epsilon$--approximation of GSAM with a constant batch size and decaying learning rate)}}
Consider the sequence $(\bm{x}_t)$ generated by the mini-batch GSAM algorithm (Algorithm \ref{algo:1}) with a constant batch size $b_t = b \in (0,n]$ and a decaying learning rate $\eta_t \in [\underline{\eta},\overline{\eta}]$ defined by (\ref{cosine_lr}) or (\ref{linear_lr}). Furthermore, let us assume that there exists a positive number $G$ defined as in Theorem \ref{thm:3_1}. Let $\epsilon > 0$ be the precision and let $b > 0$, $\overline{\eta} > 0$ ($= \underline{\eta}$), $\alpha \in \mathbb{R}$, and $\rho \geq 0$ such that 
{
\begin{align}
&\overline{\eta} 
\in
\begin{cases}
\left[
\frac{24 \sigma \rho C}{\epsilon^2}
\left(
\frac{G}{\sqrt{b}} + \frac{3 \sigma \rho C}{n b} \sum_{i\in [n]} L_i
\right),\
\frac{2 (|\alpha| + 1)^{-2} n^3 \epsilon^2}
{9 G^2 \sum_{i\in [n]} L_i \left\{ n^2 + 4 \rho^2 C^2 \big( \sum_{i\in [n]} L_i \big)^2 \right\}}
\right] &\text{ if } (\ref{cosine_lr}) \text{ is used},\\
\left[
\frac{24 \sigma \rho C}{\epsilon^2}
\left(
\frac{G}{\sqrt{b}} + \frac{3 \sigma \rho C}{n b} \sum_{i\in [n]} L_i
\right),\
\frac{(|\alpha| + 1)^{-2} n^3 \epsilon^2}
{4 G^2 \sum_{i\in [n]} L_i \left\{ n^2 + 4 \rho^2 C^2 \big( \sum_{i\in [n]} L_i \big)^2 \right\}}
\right] &\text{ if } (\ref{linear_lr}) \text{ is used},
\end{cases}
\label{eta_2_1}\\ 
&\rho (|\alpha| + 1) 
\leq 
\frac{n \sqrt{b} \epsilon^2}{6G(CG \sqrt{b} + B \sigma)\sum_{i\in [n]} L_i},
\text{ }
\rho
\leq 
\frac{n b \epsilon^2}{24 G \sqrt{n^2 + b^2}\sum_{i \in [n]} L_i}, \label{eta_3_1}
\end{align}
}
where $B$ and $C$ are nonnegative constants.
Then, there exists $t_0 \in \mathbb{N}$ such that, for all $T \geq t_0$, 
\begin{align*}
&\min_{t \in [0:T -1]}\mathbb{E}\left[ \left\|\nabla \hat{f}_{S,\rho}^{\mathrm{SAM}}(\bm{x}_t) \right\|_2 \right]
\leq \epsilon.
\end{align*}
\end{theorem} 

Theorem \ref{thm:3_2} indicates that the parameters $|\alpha|$ and $\rho$ in (\ref{eta_3_1}) become small and thereby achieve an $\epsilon$--approximation of GSAM, as is also seen in Theorem \ref{thm:3_1}. A discussion similar to the one showing (\ref{eta_small}) implies that the maximum learning rate $\overline{\eta}$ satisfying (\ref{eta_2_1}) using a small $\rho$ is approximately
\begin{align}\label{eta_small_1}
&\overline{\eta} 
\in
\begin{cases}
\left(
0,  
\frac{2 (|\alpha| + 1)^{-2} n \epsilon^2}{9 G^2 \sum_{i\in [n]} L_i}
\right] &\text{ if } (\ref{cosine_lr}) \text{ is used},\\
\left(
0,  
\frac{(|\alpha| + 1)^{-2} n \epsilon^2}{4 G^2 \sum_{i\in [n]} L_i}
\right] &\text{ if } (\ref{linear_lr}) \text{ is used}.
\end{cases}
\end{align}
From (\ref{eta_small_1}), it would be appropriate to set a small $\eta$ in order to achieve an $\epsilon$-approximation of GSAM. In fact, the numerical results in \citep{zhuang2022surrogate} used small values of $\overline{\eta}$, such as $1.6$ and $3 \times 10^{-3}$. 

Theorem \ref{thm:3_2} guarantees that SGD is an $\epsilon$-approximation in the sense of $\min_{t\in [0:T-1]} \mathbb{E}[\|\nabla f_S (\bm{x}_t)\|_2] \leq \epsilon$. Moreover, using $\alpha = \rho = 0$ makes the upper bound of $\min_{t\in [0:T-1]} \mathbb{E}[\|\nabla f_S (\bm{x}_t)\|_2]$ smaller than when using $\alpha \neq 0 \lor \rho \neq 0$. Hence, SGD using $\alpha = \rho = 0$ would minimize $\|\nabla f_S (\bm{x}_t)\|_2$ more quickly than SAM/GSAM using $\alpha \neq 0 \lor \rho \neq 0$ (see also Figure \ref{fig4} (Left) indicating that SGD minimizes $f_S$ more quickly than SAM/GSAM). Meanwhile, the previous results in \citep{foret2021sharpnessaware,zhuang2022surrogate} indicate that using $\alpha \neq 0 \lor \rho \neq 0$ leads to a higher generalization capability compared with using $\alpha = \rho = 0$ (see Tables \ref{table:CIFAR and T-ImageNet} and \ref{table:ImageNet} which show that the generalization capability of SAM/GSAM+C has a higher than that of SGD+C).

\section{Numerical results}\label{sec:3}
We used a computer equipped with NVIDIA GeForce RTX 4090$\times$2GPUs and an Intel Core i9 13900KF CPU. The software environment was Python 3.10.12, PyTorch 2.1.0, and CUDA 12.2. The solid lines in the figures represent the mean value and the shaded areas represent the maximum and minimum over three runs. 
The code used in the experiments is available at {\url{https://github.com/iiduka-researches/GSAM_batch_inc}.}
{For all experiments, the hyperparameters $\alpha$ and $\rho$ were determined by conducting a grid search of $\alpha \in \{0.01, 0.02, 0.03 \}$ and $\rho \in \{0.01 , 0.02 , 0.03 , 0.04 , 0.05\}$.}
{For each model and dataset, the grid search was conducted in the baseline setting 
with a constant batch size and a constant learning rate, and the selected values of $\alpha$ and $\rho$ 
were used without modification in the experiments with an increasing batch size (+B) and 
a cosine-annealing learning rate (+C).
}
\begin{table*}[ht]
\caption{{Mean value of the test error (Test Error) and worst-case $\ell_{\infty}$ adaptive sharpness (Sharpness) on the CIFAR-10/100 and Tiny-ImageNet datasets. ``(algorithm) + B" refers to ``(algorithm) + increasing\_batch" and ``(algorithm) + C" refers to ``(algorithm) + Cosine".}}
\label{table:CIFAR and T-ImageNet}

\centering
\resizebox{\textwidth}{!}{%
\begin{tabular}{c|c|ccc|ccc|ccc}
\hline
\textbf{Dataset / Model} & Metric & SGD & SAM & GSAM & SGD + B & SAM + B & GSAM + B & SGD + C & SAM + C & GSAM + C \\ \hline
\textbf{CIFAR-10 } & Test Error & 7.17 & 6.34 & 6.21 & 6.39 & \textbf{6.03} & 6.06 & 6.63 & 6.08 & 6.2 \\
\textbf{/ ResNet-18} & Sharpness & 106.46 & 51.82 & 48.39 & 8.73 & 4.27 & \textbf{4.08} & 141.37 & 88.53 & 89.3 \\ \hline
\textbf{CIFAR-10 } & Test Error & 7.04 & 6.13 & 5.98 & 5.53 & 5.10 & \textbf{5.05} & 6.67 & 5.87 & 5.92 \\
\textbf{/ WRN-28-10} & Sharpness & 1124.45 & 546.93 & 543.27 & 48.25 & \textbf{38.79} & 40.28 & 1467.67 & 771.79 & 794.45 \\ \hline
\textbf{CIFAR-100 } & Test Error & 26.61 & 26.39 & 26.61 & 25.58 & \textbf{25.10} & 25.18 & 26.63 & 25.87 & 26.12 \\
\textbf{/ ResNet-18} & Sharpness & 154.27 & 46.23 & 47.55 & 1.33 & 0.94 & \textbf{0.90} & 155.88 & 72.70 & 71.86 \\ \hline
\textbf{CIFAR-100 } & Test Error & 25.62 & 24.78 & 24.94 & 22.65 & \textbf{21.10} & 21.50 & 25.57 & 24.16 & 24.00 \\
\textbf{/ WRN-28-10} & Sharpness & 1113.26 & 456.20 & 435.17 & 22.72 & \textbf{10.99} & 12.37 & 1148.09 & 687.44 & 665.13 \\ \hline
\textbf{T-ImageNet } & Test Error & 39.62 & 40.28 & 40.13 & 39.36 & 38.40 & \textbf{38.25} & 39.84 & 39.07 & 39.05 \\
\textbf{/ ResNet-18}& Sharpness & 279.07 & 97.52 & 97.23 & 13.33 & 10.65 & \textbf{10.31} & 333.73 & 183.59 & 192.29 \\ \hline
\textbf{T-ImageNet  } & Test Error & 38.75 & 38.60 & 38.92 & 36.55 & 35.36 & \textbf{35.14} & 39.30 & 37.86 & 37.78 \\
 \textbf{  / WRN-28-10}& Sharpness & 2550.16 & 917.93 & 926.94 & 176.80 & 128.35 & \textbf{119.84} & 3066.71 & 1657.00 & 1803.57 \\ \hline
\end{tabular}
}
\end{table*}

\begin{table*}[ht]
\caption{{Mean value of the test error (Test Error) and worst-case $\ell_{\infty}$ adaptive sharpness (Sharpness) on the ImageNet dataset. ``(algorithm) + B" refers to ``(algorithm) + increasing\_batch" and ``(algorithm) + C" refers to ``(algorithm) + Cosine".}}
\label{table:ImageNet}
\centering

\resizebox{\textwidth}{!}{%
\begin{tabular}{c|c|ccc|ccc|ccc}
\hline
\textbf{Dataset / Model} & Metric & SGD & SAM & GSAM & SGD + B & SAM + B & GSAM + B & SGD + C & SAM + C & GSAM + C \\ \hline
\textbf{ImageNet } & Test Error & 32.23 & 30.62 & 30.63 & 29.19 & \textbf{29.06} & 29.23 & 30.58 & 29.96 & 29.72 \\
\textbf{/ ResNet-50}& Sharpness & 492.12 & 268.60 & 297.82 & 55.31 & 41.83 & \textbf{38.13} & 1504.53 & 853.04 & 670.51 \\ \hline
\end{tabular}
}
\end{table*}

\subsection{Training ResNets using algorithms}
{\textbf{CIFAR}\citep{krizhevsky2009learning} We set $E = 200$, $\eta = \overline{\eta} = 0.1$, $\underline{\eta} = 0.001$, $\alpha = 0.02$ and $\rho = 0.05$. We trained ResNet-18 and Wide-ResNet-28-10 on the CIFAR-10/100 dataset. SGD/SAM/ GSAM used a fixed batch size of 128 and a fixed learning rate of 0.1. SGD/SAM/GSAM + Cosine used a fixed batch size of 128 and a cosine-annealing learning rate defined by (\ref{cosine_lr}). SGD/SAM/GSAM + increasing\_batch used increasing batch sizes [16, 32, 64, 128, 256] with a fixed learning rate of 0.1. For the increasing batch sizes [16, 32, 64, 128, 256], the batch size was set to 16 for the first 40 epochs, and then it was doubled every 40 epochs (See Appendix \ref{train curves} for training curves).
}

\noindent\textbf{Tiny-ImageNet}\citep{Le2015TinyIV} We set $E = 100$, $\eta = \overline{\eta} = 0.1$, $\underline{\eta} = 0.001$, $\alpha = 0.02$ and $\rho = 0.05$. We trained ResNet-18 and Wide-ResNet-28-10 on the Tiny-ImageNet dataset. SGD/SAM/GSAM + Cosine used a fixed batch size of 128 and a cosine-annealing learning rate defined by (\ref{cosine_lr}). SGD/SAM/GSAM + increasing\_batch used increasing batch sizes [16, 32, 64, 128] with a fixed learning rate of 0.1. For the increasing batch sizes [16, 32, 64, 128], the batch size was set to 16 for the first 25 epochs, and then it was doubled every 25 epochs (See Appendix \ref{train curves} for training curves).


Table \ref{table:CIFAR and T-ImageNet} summarizes the mean values of the test errors and the worst-case $\ell_{\infty}$ adaptive sharpness defined by \citep[(1)]{pmlr-v202-andriushchenko23a} for the parameters $\bm{c} = (1,1,\cdots,1)^\top$ and $\rho = 0.0002$ obtained by the algorithm after $E$ epochs. {The sharpness reported here is one diagnostic measure of the flatness of the loss around the obtained solution, and it is not equivalent to the test error (i.e., generalization). Indeed, sharpness can correlate with training hyperparameters such as the learning rate rather than with the test error (\cite{pmlr-v202-andriushchenko23a}). Accordingly, we interpret the test error and the sharpness separately.} SAM+B (SAM + increasing\_batch) and GSAM+B (GSAM + increasing\_batch) achieved lower test errors and {lower measured sharpness than those of the algorithms using a constant batch size and either a constant or a cosine-annealing learning rate.}


\noindent\textbf{ImageNet}\citep{ILSVRC15} We set $E = 100$, $\eta = \overline{\eta} = 0.1$, $\underline{\eta} = 0.001$, $\alpha = 0.01$ and $\rho = 0.05$. We trained ResNet-50 on the ImageNet dataset.  SGD/SAM/GSAM + Cosine used a fixed batch size of 256 and a cosine-annealing learning rate defined by (\ref{cosine_lr}). SGD/SAM/GSAM + increasing\_batch used increasing batch sizes [32, 64, 128, 256] with a fixed learning rate of 0.1. For the increasing batch sizes [32, 64, 128, 256], the batch size was set to 32 for the first 25 epochs, and then it was doubled every 25 epochs. From Table \ref{table:ImageNet}, SAM+B (SAM + increasing\_batch) and GSAM+B (GSAM + increasing\_batch) achieved lower test errors and {lower measured sharpness}, consistent with the results observed on CIFAR-10/100 and T-ImageNet.

\section{Conclusion}
We gave upper and lower bounds on the search direction noise of GSAM for solving the SAM problem. Then, we examined GSAM with two mini-batch and learning rate schedulers based on the bounds: an increasing batch size and constant learning rate scheduler and a constant batch size and decaying learning rate scheduler. We performed convergence analyses on GSAM for the two schedulers. We also provided numerical results to support the analyses. {The numerical results showed that SAM/GSAM with an increasing batch size achieves a lower test error and lower measured sharpness than those of SAM/GSAM with a constant batch size.}

\acks{
{
We are sincerely grateful to the four anonymous reviewers for helping us improve the original manuscript. This research is partly supported by the computational resources of the DGX A100 named TAIHO at Meiji University. 
This work was supported by the Japan Society for the
Promotion of Science (JSPS) KAKENHI Grant Number 24K14846 awarded to Hideaki Iiduka.
}
}

\bibliography{acml26}

\clearpage
\appendix
\section{Proofs of Theorems \ref{thm:1} and \ref{thm:2}}
\label{appendix_a}

Let $\mathbb{N}$ be the set of natural numbers. Let $[n] := \{1,2,\cdots,n\}$ and $[0:n] := \{0,1,\cdots,n\}$ for $n \in \mathbb{N}$. Let $\mathbb{R}^d$ be a $d$-dimensional Euclidean space with inner product $\langle \bm{x}, \bm{y} \rangle_2 = \bm{x}^\top \bm{y}$ $(\bm{x},\bm{y} \in \mathbb{R}^d)$ and its induced norm $\|\bm{x}\|_2 := \sqrt{\langle \bm{x}, \bm{x} \rangle_2}$ $(\bm{x} \in \mathbb{R}^d)$. The gradient and Hessian of a twice differentiable function $f \colon \mathbb{R}^d \to \mathbb{R}$ at $\bm{x} \in \mathbb{R}^d$ are denoted by $\nabla f(\bm{x})$ and $\nabla^2 f(\bm{x})$, respectively. Let $L > 0$. A differentiable function $f \colon \mathbb{R}^d \to \mathbb{R}$ is said to be $L$--smooth if the gradient $\nabla f \colon \mathbb{R}^d \to \mathbb{R}^d$ is Lipschitz continuous; i.e., for all $\bm{x},\bm{y} \in \mathbb{R}^d$, $\| \nabla f(\bm{x}) - \nabla f(\bm{y}) \|_2 \leq L \|\bm{x} - \bm{y}\|_2$. Let $O$ and $\Theta$ be Landau's symbols, i.e., $y_t = O (x_t)$ (resp. $y_t = \Theta (x_t)$) if there exist $c > 0$ (resp. $c_1, c_2 > 0$) and $t_0 \in \mathbb{N}$ such that, for all $t \geq t_0$, $y_t \leq c x_t$ (resp. $c_1 x_t \leq y_t \leq c_2 x_t$). 

\subsection{Propositions}
\label{appendix_a_1}
We first give an upper bound of the variance of the stochastic gradient $\nabla f_{S_t}(\bm{x})$.

\begin{proposition}\label{prop:1}
Under Assumption \ref{assum:1}, we have that, for all $\bm{x} \in \mathbb{R}^d$ and all $t \in \mathbb{N} \cup \{0\}$,
{
\begin{align*}
&\mathbb{E}_{\bm{\xi}_t} \left[\nabla f_{S_t}(\bm{x}) \Big| \hat{\bm{\xi}}_{t-1} \right] = \nabla f_S (\bm{x}),\\ 
&\mathbb{V}_{\bm{\xi}_t} \left[\nabla f_{S_t}(\bm{x}) \Big| \hat{\bm{\xi}}_{t-1} \right] 
= \mathbb{E}_{\bm{\xi}_t} \left[\| \nabla f_{S_t}(\bm{x}) - \nabla f_S (\bm{x})  \|_2^2 \Big| \hat{\bm{\xi}}_{t-1} \right]
\leq \frac{n - b_t}{n-1} \cdot \frac{\sigma^2}{b_t}
\leq \frac{\sigma^2}{b_t},
\end{align*}
}
where $\mathbb{E}_{\bm{\xi}_t}[\cdot|\hat{\bm{\xi}}_{t-1}]$ stands for the expectation with respect to $\bm{\xi}_t$ conditioned on $\bm{\xi}_{t-1} = \hat{\bm{\xi}}_{t-1}$.
\end{proposition}

{\em Proof:} Let $\bm{x} \in \mathbb{R}^d$ and $t \in \mathbb{N} \cup \{0\}$. Assumption \ref{assum:1}(A3) ensures that 
\begin{align*}
\mathbb{E}_{\bm{\xi}_t} \left[\nabla f_{S_t}(\bm{x}) \Big| \hat{\bm{\xi}}_{t-1} \right] 
= 
\mathbb{E}_{\bm{\xi}_t} \left[\frac{1}{b_t} \sum_{i\in [b_t]} \nabla f_{\xi_{t,i}} (\bm{x}) \Bigg| \hat{\bm{\xi}}_{t-1} \right] 
=
\frac{1}{b_t} \sum_{i\in [b_t]}
\mathbb{E}_{\xi_{t,i}} \left[\nabla f_{\xi_{t,i}} (\bm{x}) \Big| \hat{\bm{\xi}}_{t-1} \right],
\end{align*}
which, together with Assumption \ref{assum:1}(A2)(i) and the independence of $\bm{\xi}_t$ and $\bm{\xi}_{t-1}$, implies that
\begin{align*}
\mathbb{E}_{\bm{\xi}_t} \left[\nabla f_{S_t}(\bm{x}) \Big| \hat{\bm{\xi}}_{t-1} \right] = \nabla f_S (\bm{x}).
\end{align*}
Assumption \ref{assum:1}(A3) implies that
\begin{align*}
\mathbb{V}_{\bm{\xi}_t} \left[ \nabla f_{S_t}(\bm{x}) \Big| \hat{\bm{\xi}}_{t-1} \right]
&= 
\mathbb{E}_{\bm{\xi}_t} \left[ \|\nabla f_{S_t} (\bm{x}) - \nabla f_S (\bm{x})\|_2^2 \Big| \hat{\bm{\xi}}_{t-1} \right]\\
&= 
\mathbb{E}_{\bm{\xi}_t} 
\left[ \left\| 
\frac{1}{b_t} \sum_{i\in [b_t]} \nabla f_{\xi_{t,i}} (\bm{x}) - \nabla f_S (\bm{x}) \right\|_2^2 \Bigg| \hat{\bm{\xi}}_{t-1} \right]\\
&= 
\frac{1}{b_t^2} \mathbb{E}_{\bm{\xi}_t} 
\left[ \left\| 
\sum_{i \in [b_t]} \left( 
\nabla f_{\xi_{t,i}} (\bm{x}) - \nabla f_S (\bm{x}) 
\right) 
\right\|_2^2 \Bigg| \hat{\bm{\xi}}_{t-1} \right].
\end{align*}
{
Let $\bm{v}_k := \nabla f_k(\bm{x}) - \nabla f_S(\bm{x})$ $(k \in [n])$ and
$s^2 := \frac{1}{n}\sum_{k\in[n]}\|\bm{v}_k\|_2^2$.
Then, $\sum_{k\in[n]} \bm{v}_k = \bm{0}$, and Assumption \ref{assum:1}(A2)(ii) implies that $s^2 \leq \sigma^2$.
Expanding the squared norm gives
\begin{align*}
\mathbb{V}_{\bm{\xi}_t}\left[\nabla f_{S_t}(\bm{x}) \Big| \hat{\bm{\xi}}_{t-1}\right]
&= \frac{1}{b_t^2}\Bigg(\sum_{i\in[b_t]}\mathbb{E}_{\bm{\xi}_t}\left[\|\bm{v}_{\xi_{t,i}}\|_2^2 \Big| \hat{\bm{\xi}}_{t-1}\right]\\
&\qquad\quad + \sum_{\substack{i,j\in[b_t]\\ i\neq j}}\mathbb{E}_{\bm{\xi}_t}\left[\langle \bm{v}_{\xi_{t,i}}, \bm{v}_{\xi_{t,j}}\rangle_2 \Big| \hat{\bm{\xi}}_{t-1}\right]\Bigg).
\end{align*}
Since $\bm{\xi}_t$ is independent of $\bm{\xi}_{t-1}$ and each $\xi_{t,i}$ is uniformly distributed on $[n]$, we have, for all $i \in [b_t]$,
\begin{align*}
\mathbb{E}_{\bm{\xi}_t}\left[\|\bm{v}_{\xi_{t,i}}\|_2^2 \Big| \hat{\bm{\xi}}_{t-1}\right]
= \frac{1}{n}\sum_{k\in[n]}\|\bm{v}_k\|_2^2 = s^2.
\end{align*}
Moreover, since $\xi_{t,1}, \ldots, \xi_{t,b_t}$ are drawn without replacement,
for all $i, j \in [b_t]$ with $i \neq j$, the pair $(\xi_{t,i}, \xi_{t,j})$ is uniformly distributed over $\{(k,l) \in [n]^2 : k \neq l\}$.
Hence, for all $i, j \in [b_t]$ with $i \neq j$,
\begin{align*}
\mathbb{E}_{\bm{\xi}_t}\left[\langle \bm{v}_{\xi_{t,i}}, \bm{v}_{\xi_{t,j}}\rangle_2 \Big| \hat{\bm{\xi}}_{t-1}\right]
&= \frac{1}{n(n-1)}\sum_{\substack{k,l\in[n]\\ k\neq l}}\langle \bm{v}_k, \bm{v}_l\rangle_2\\
&= \frac{1}{n(n-1)}\left(\left\|\sum_{k\in[n]}\bm{v}_k\right\|_2^2 - \sum_{k\in[n]}\|\bm{v}_k\|_2^2\right)\\
&= -\frac{s^2}{n-1}.
\end{align*}
Therefore,
\begin{align*}
\mathbb{V}_{\bm{\xi}_t}\left[\nabla f_{S_t}(\bm{x}) \Big| \hat{\bm{\xi}}_{t-1}\right]
&= \frac{1}{b_t^2}\left(b_t s^2 - \frac{b_t(b_t-1)}{n-1}\,s^2\right)\\
&= \frac{n-b_t}{n-1}\cdot\frac{s^2}{b_t}\\
&\leq \frac{n-b_t}{n-1}\cdot\frac{\sigma^2}{b_t}\\
&\leq \frac{\sigma^2}{b_t},
\end{align*}
}
which completes the proof.

We will use the following proposition to prove Theorem \ref{thm:1}.

\begin{proposition}\citep[3.2.6, (10)]{doi:10.1137/1.9780898719468}\label{prop:2}
Let $f \colon \mathbb{R}^d \to \mathbb{R}$ be twice differentiable. Then, for all $\bm{x},\bm{y} \in \mathbb{R}^d$,
\begin{align*}
\nabla f(\bm{y}) = \nabla f(\bm{x}) + \int_0^1 \nabla^2 f (\bm{x} + t (\bm{y} - \bm{x}))(\bm{y} - \bm{x}) \mathrm{d}t. 
\end{align*}
\end{proposition}

\subsection{Proof of Theorem \ref{thm:1}}
We will use Propositions \ref{prop:1} and \ref{prop:2} to prove Theorem \ref{thm:1}.

Let $t \in \mathbb{N} \cup \{0\}$ and $b < n$ and suppose that $\bm{x}_t$ generated by Algorithm \ref{algo:1} satisfies $\nabla f_{S_t} (\bm{x}_t) \neq \bm{0}$ and $\nabla f_{S} (\bm{x}_t) \neq \bm{0}$. Then, we have 
\begin{align}\label{key_3}
\begin{split}
\|\hat{\bm{\omega}}_t\|_2^2
&= 
\left\|\nabla \hat{f}_{S_t,\rho}^{\mathrm{SAM}}(\bm{x}_t) -  \nabla \hat{f}_{S,\rho}^{\mathrm{SAM}}(\bm{x}_t) \right\|_2^2\\
&\underset{(\ref{sam_s_t})}=
\left\|
\nabla f_{S_t} \left(\bm{x}_t + \rho \frac{\nabla f_{S_t} (\bm{x}_t)}{\|\nabla f_{S_t} (\bm{x}_t)\|_2} \right)
- 
\nabla f_{S} \left(\bm{x}_t + \rho \frac{\nabla f_{S} (\bm{x}_t)}{\|\nabla f_{S} (\bm{x}_t)\|_2} \right)
\right\|_2^2\\
&= 
\Bigg\|
\nabla f_{S_t} (\bm{x}_t) + \int_0^1 \nabla^2 f_{S_t} \left(\bm{x}_t + \rho s \frac{\nabla f_{S_t} (\bm{x}_t)}{\|\nabla f_{S_t} (\bm{x}_t)\|_2} \right) \rho \frac{\nabla f_{S_t} (\bm{x}_t)}{\|\nabla f_{S_t} (\bm{x}_t)\|_2} \mathrm{d}s\\
&\quad -
\left(
\nabla f_{S} (\bm{x}_t) + \int_0^1 \nabla^2 f_{S} \left(\bm{x}_t + \rho s \frac{\nabla f_{S} (\bm{x}_t)}{\|\nabla f_{S} (\bm{x}_t)\|_2} \right) \rho \frac{\nabla f_{S} (\bm{x}_t)}{\|\nabla f_{S} (\bm{x}_t)\|_2}
\mathrm{d}s 
\right) \Bigg\|_2^2,
\end{split} 
\end{align}
where the third equation comes from Proposition \ref{prop:2}. From $\|\bm{x} + \bm{y}\|_2^2 \leq 2 \|\bm{x}\|_2^2 + 2 \|\bm{y}\|_2^2$ ($\bm{x},\bm{y} \in \mathbb{R}^d$), we have 
\begin{align*}
\|\hat{\bm{\omega}}_t\|_2^2
&\leq
2 \|\nabla f_{S_t} (\bm{x}_t) - \nabla f_{S} (\bm{x}_t)\|_2^2\\
&\quad 
+ 
4 \left\| 
\int_0^1 \nabla^2 f_{S_t} \left(\bm{x}_t + \rho s \frac{\nabla f_{S_t} (\bm{x}_t)}{\|\nabla f_{S_t} (\bm{x}_t)\|_2} \right) \rho \frac{\nabla f_{S_t} (\bm{x}_t)}{\|\nabla f_{S_t} (\bm{x}_t)\|_2} \mathrm{d}s
\right\|_2^2\\
&\quad +
4 \left\| 
\int_0^1 \nabla^2 f_{S} \left(\bm{x}_t + \rho s \frac{\nabla f_{S} (\bm{x}_t)}{\|\nabla f_{S} (\bm{x}_t)\|_2} \right) \rho \frac{\nabla f_{S} (\bm{x}_t)}{\|\nabla f_{S} (\bm{x}_t)\|_2}
\mathrm{d}s
\right\|_2^2,
\end{align*}
which, together with the property of $\|\cdot \|_2$, implies that 
\begin{align}\label{key_1_1}
\begin{split}
\|\hat{\bm{\omega}}_t\|_2^2
&\leq
2 \|\nabla f_{S_t} (\bm{x}_t) - \nabla f_{S} (\bm{x}_t)\|_2^2\\
&\quad 
+ 
4  
\left( \rho \int_0^1 \left\|\nabla^2 f_{S_t} \left(\bm{x}_t + \rho s \frac{\nabla f_{S_t} (\bm{x}_t)}{\|\nabla f_{S_t} (\bm{x}_t)\|_2} \right) \right\|_2  \mathrm{d}s \right)^2\\
&\quad +
4 \left( 
\rho \int_0^1 \left\| \nabla^2 f_{S} \left(\bm{x}_t + \rho s \frac{\nabla f_{S} (\bm{x}_t)}{\|\nabla f_{S} (\bm{x}_t)\|_2} \right) \right\|_2 
\mathrm{d}s
\right)^2.
\end{split}
\end{align}
Meanwhile, the triangle inequality and the $L_i$--smoothness of $f_i$ (see (A1)) ensure that, for all $\bm{x}, \bm{y} \in \mathbb{R}^d$,
\begin{align*}
\|\nabla f_{S_t}(\bm{x}) - \nabla f_{S_t}(\bm{y}) \|_2
&= 
\left\|
\frac{1}{b_t} \sum_{i\in [b_t]} 
(\nabla f_{\xi_{t,i}}(\bm{x})
-
\nabla f_{\xi_{t,i}}(\bm{y}))
\right\|_2
\leq
\frac{1}{b_t} \sum_{i\in [b_t]} \left\| 
\nabla f_{\xi_{t,i}}(\bm{x})
-
\nabla f_{\xi_{t,i}}(\bm{y})
\right\|_2\\
&\leq
\frac{1}{b_t} \sum_{i\in [b_t]} L_{\xi_{t,i}} \|\bm{x} - \bm{y}\|_2
\leq
\frac{1}{b_t} \sum_{i\in [n]} L_{i} \|\bm{x} - \bm{y}\|_2,
\end{align*}
which implies that, for all $\bm{x} \in \mathbb{R}^d$, $\|\nabla^2 f_{S_t}(\bm{x})\|_2 \leq b_t^{-1} \sum_{i\in [n]} L_{i}$. A discussion similar to the one showing that $\nabla f_{S_t}$ is $b_t^{-1} \sum_{i\in [n]} L_{i}$--smooth ensures that $\nabla f_{S}$ is $n^{-1} \sum_{i\in [n]} L_{i}$--smooth, which in turn implies that, for all $\bm{x} \in \mathbb{R}^d$, $\|\nabla^2 f_{S}(\bm{x})\|_2 \leq n^{-1} \sum_{i\in [n]} L_{i}$. Accordingly, (\ref{key_1_1}) guarantees that
\begin{align}\label{key_2_1}
\|\hat{\bm{\omega}}_t\|_2^2
\leq
2 \|\nabla f_{S_t} (\bm{x}_t) - \nabla f_{S} (\bm{x}_t)\|_2^2 
+ 
\frac{4 \rho^2}{b_t^2} \bigg(\sum_{i \in [n]} L_i \bigg)^2 
+ 
\frac{4 \rho^2}{n^2} \bigg(\sum_{i \in [n]} L_i \bigg)^2. 
\end{align}
Taking the expectation with respect to $\bm{\xi}_t$ conditioned on $\bm{\xi}_{t-1} = \hat{\bm{\xi}}_{t-1}$ on both sides of (\ref{key_2_1}) ensures that
\begin{align*}
\mathbb{E}_{\bm{\xi}_t}[ \|\hat{\bm{\omega}}_t\|_2^2| \hat{\bm{\xi}}_{t-1}]
\leq 
2 \mathbb{E}_{\bm{\xi}_t}[\|\nabla f_{S_t} (\bm{x}_t) - \nabla f_{S} (\bm{x}_t)\|_2^2 | \hat{\bm{\xi}}_{t-1}]
+ 
\frac{4 \rho^2}{b_t^2} \bigg(\sum_{i \in [n]} L_i \bigg)^2 
+ 
\frac{4 \rho^2}{n^2} \bigg(\sum_{i \in [n]} L_i \bigg)^2,
\end{align*}
which, together with Proposition \ref{prop:1}, implies that
\begin{align*}
\mathbb{E}_{\bm{\xi}_t}[ \|\hat{\bm{\omega}}_t\|_2^2| \hat{\bm{\xi}}_{t-1}]
\leq 
\frac{2 \sigma^2}{b_t}
+ 
\frac{4 \rho^2}{b_t^2} \bigg(\sum_{i \in [n]} L_i \bigg)^2 
+ 
\frac{4 \rho^2}{n^2} \bigg(\sum_{i \in [n]} L_i \bigg)^2.
\end{align*}
Since $\bm{\xi}_t$ is independent of $\bm{\xi}_{t-1}$, we have 
\begin{align*}
\mathbb{E}_{\bm{\xi}_{t-1}} \mathbb{E}_{\bm{\xi}_t}[ \|\hat{\bm{\omega}}_t\|_2^2]
=
\mathbb{E}_{\bm{\xi}_{t-1}}[\mathbb{E}_{\bm{\xi}_t}[ \|\hat{\bm{\omega}}_t\|_2^2| \bm{\xi}_{t-1}]]
\leq
\frac{2 \sigma^2}{b_t}
+ 
\frac{4 \rho^2}{b_t^2} \bigg(\sum_{i \in [n]} L_i \bigg)^2 
+ 
\frac{4 \rho^2}{n^2} \bigg(\sum_{i \in [n]} L_i \bigg)^2,
\end{align*}
which, together with $\mathbb{E} = \mathbb{E}_{\bm{\xi}_0} \mathbb{E}_{\bm{\xi}_1} \cdots \mathbb{E}_{\bm{\xi}_t}$, implies that 
\begin{align}\label{key_4}
\mathbb{E}[ \|\hat{\bm{\omega}}_t\|_2^2]
\leq
\frac{2 \sigma^2}{b_t}
+ 
\frac{4 \rho^2}{b_t^2} \bigg(\sum_{i \in [n]} L_i \bigg)^2 
+ 
\frac{4 \rho^2}{n^2} \bigg(\sum_{i \in [n]} L_i \bigg)^2.
\end{align}
Suppose that $\bm{x}_t$ generated by Algorithm \ref{algo:1} satisfies either $\nabla f_{S_t} (\bm{x}_t) = \bm{0}$ or $\nabla f_{S} (\bm{x}_t) = \bm{0}$. Let $\nabla f_{S_t} (\bm{x}_t) = \bm{0}$. A discussion similar to the one for obtaining (\ref{key_3}) and (\ref{key_1_1}), together with (\ref{sam_s_t}), ensures that
\begin{align*}
\|\hat{\bm{\omega}}_t\|_2^2
&= 
\left\|\nabla \hat{f}_{S_t,\rho}^{\mathrm{SAM}}(\bm{x}_t) -  \nabla \hat{f}_{S,\rho}^{\mathrm{SAM}}(\bm{x}_t) \right\|_2^2\\
&=
\left\|
\nabla f_{S_t} (\bm{x}_t + \bm{u} )
- 
\nabla f_{S} \left(\bm{x}_t + \rho \frac{\nabla f_{S} (\bm{x}_t)}{\|\nabla f_{S} (\bm{x}_t)\|_2} \right)
\right\|_2^2\\
&= 
\Bigg\|
\nabla f_{S_t} (\bm{x}_t) + \int_0^1 \nabla^2 f_{S_t} (\bm{x}_t + s \bm{u}) \bm{u} \mathrm{d}s\\
&\quad -
\left(
\nabla f_{S} (\bm{x}_t) + \int_0^1 \nabla^2 f_{S} \left(\bm{x}_t + \rho s \frac{\nabla f_{S} (\bm{x}_t)}{\|\nabla f_{S} (\bm{x}_t)\|_2} \right) \rho \frac{\nabla f_{S} (\bm{x}_t)}{\|\nabla f_{S} (\bm{x}_t)\|_2}
\mathrm{d}s 
\right) \Bigg\|_2^2, 
\end{align*}
which, together with $\|\bm{u}\|_2 \leq \rho$, implies that 
\begin{align*}
\|\hat{\bm{\omega}}_t\|_2^2
&\leq
2 \|\nabla f_{S_t} (\bm{x}_t) - \nabla f_{S} (\bm{x}_t)\|_2^2\\
&\quad 
+ 
4 \left(\rho \int_0^1 \left\|\nabla^2 f_{S_t} (\bm{x}_t + s \bm{u}) \right\|_2  \mathrm{d}s \right)^2\\
&\quad +
4 \left( 
\rho \int_0^1 \left\| \nabla^2 f_{S} \left(\bm{x}_t + \rho s \frac{\nabla f_{S} (\bm{x}_t)}{\|\nabla f_{S} (\bm{x}_t)\|_2} \right) \right\|_2 
\mathrm{d}s
\right)^2.
\end{align*}
Hence, the same discussion as in (\ref{key_2_1}) leads to the finding that
\begin{align*}
\|\hat{\bm{\omega}}_t\|_2^2
\leq
2 \|\nabla f_{S_t} (\bm{x}_t) - \nabla f_{S} (\bm{x}_t)\|_2^2 
+ 
\frac{4 \rho^2}{b_t^2} \bigg(\sum_{i \in [n]} L_i \bigg)^2 
+ 
\frac{4 \rho^2}{n^2} \bigg(\sum_{i \in [n]} L_i \bigg)^2. 
\end{align*}
Accordingly, Proposition \ref{prop:1} and a discussion similar to the one showing (\ref{key_4}) imply that (\ref{key_4}) holds in the case of $\nabla f_{S_t} (\bm{x}_t) = \bm{0}$. Moreover, this ensures that (\ref{key_4}) holds in the case of $\nabla f_{S} (\bm{x}_t) = \bm{0}$. Therefore, we have
\begin{align}\label{key_5}
\mathbb{E}[ \|\hat{\bm{\omega}}_t\|_2]
\leq
\sqrt{\frac{2 \sigma^2}{b_t}
+ 
\frac{4 \rho^2}{b_t^2} \bigg(\sum_{i \in [n]} L_i \bigg)^2 
+ 
\frac{4 \rho^2}{n^2} \bigg(\sum_{i \in [n]} L_i \bigg)^2}.
\end{align}
We reach the desired result for when $b_t < n$ in Theorem \ref{thm:1} from $\|\bm{\omega}_t \|_2 \leq \| \hat{\bm{\omega}}_t \|_2 + |\alpha| G_{\perp}$ and (\ref{key_5}). Moreover, we reach the desired result for when $b_t = n$ from $\|\hat{\bm{\omega}}_t\|_2^2 = 0$. This completes the proof.

\subsection{Proof of Theorem \ref{thm:2}}
Let $t \in \mathbb{N} \cup \{0\}$ and $b < n$ and suppose that $\bm{x}_t$ generated by Algorithm \ref{algo:1} satisfies $\nabla f_{S_t} (\bm{x}_t) \neq \bm{0}$ and $\nabla f_{S} (\bm{x}_t) \neq \bm{0}$. From $|\alpha| \|\nabla f_{S_t \perp}(\bm{x}_t)\|_2 \leq \|\hat{\bm{\omega}}_t\|_2$, we have 
\begin{align*}
\|\bm{\omega}_t\|_2
\geq
\|\hat{\bm{\omega}}_t \|_2 - |\alpha| \|\nabla f_{S_t \perp}(\bm{x}_t)\|_2
\geq 
\|\hat{\bm{\omega}}_t \|_2 - |\alpha| G_{\perp}.
\end{align*}
From (\ref{key_3}), we have
\begin{align}\label{key:2_3}
\|\hat{\bm{\omega}}_t\|_2
&= 
\left\|\nabla \hat{f}_{S_t,\rho}^{\mathrm{SAM}}(\bm{x}_t) -  \nabla \hat{f}_{S,\rho}^{\mathrm{SAM}}(\bm{x}_t) \right\|_2 \nonumber\\
&\geq 
\Bigg| \|\nabla f_{S_t} (\bm{x}_t) - \nabla f_{S} (\bm{x}_t)\|_2 
- \Bigg\| \int_0^1 \nabla^2 f_{S_t} \left(\bm{x}_t + \rho s \frac{\nabla f_{S_t} (\bm{x}_t)}{\|\nabla f_{S_t} (\bm{x}_t)\|_2} \right) \rho \frac{\nabla f_{S_t} (\bm{x}_t)}{\|\nabla f_{S_t} (\bm{x}_t)\|_2} \mathrm{d}s \nonumber \\
&\quad -
\int_0^1 \nabla^2 f_{S} \left(\bm{x}_t + \rho s \frac{\nabla f_{S} (\bm{x}_t)}{\|\nabla f_{S} (\bm{x}_t)\|_2} \right) \rho \frac{\nabla f_{S} (\bm{x}_t)}{\|\nabla f_{S} (\bm{x}_t)\|_2}
\mathrm{d}s 
\Bigg\|_2 \Bigg| =: |A_t|.
\end{align}
When $A_t \geq 0$, 
\begin{align*}
\|\hat{\bm{\omega}}_t\|_2
&\geq 
\|\nabla f_{S_t} (\bm{x}_t) - \nabla f_{S} (\bm{x}_t)\|_2 
- \Bigg\| \int_0^1 \nabla^2 f_{S_t} \left(\bm{x}_t + \rho s \frac{\nabla f_{S_t} (\bm{x}_t)}{\|\nabla f_{S_t} (\bm{x}_t)\|_2} \right) \rho \frac{\nabla f_{S_t} (\bm{x}_t)}{\|\nabla f_{S_t} (\bm{x}_t)\|_2} \mathrm{d}s\\
&\quad -
\int_0^1 \nabla^2 f_{S} \left(\bm{x}_t + \rho s \frac{\nabla f_{S} (\bm{x}_t)}{\|\nabla f_{S} (\bm{x}_t)\|_2} \right) \rho \frac{\nabla f_{S} (\bm{x}_t)}{\|\nabla f_{S} (\bm{x}_t)\|_2}
\mathrm{d}s 
\Bigg\|_2\\
&\geq 
\|\nabla f_{S_t} (\bm{x}_t) - \nabla f_{S} (\bm{x}_t)\|_2
- 
\rho \left( \frac{1}{b_t} + \frac{1}{n} \right) \sum_{i \in [n]} L_i,
\end{align*}
where the second inequality comes from (\ref{key_1_1}) and (\ref{key_2_1}). A similar discussion to the one in (\ref{key_4}), together with Proposition \ref{prop:1}, implies that there exists $c_t \in (0,1]$ such that
\begin{align*}
\mathbb{E}[\|\hat{\bm{\omega}}_t\|_2]
\geq 
\frac{c_t \sigma}{\sqrt{b_t}}
- 
\rho \left( \frac{1}{b_t} + \frac{1}{n} \right) \sum_{i \in [n]} L_i.
\end{align*}
Accordingly, we have 
\begin{align}\label{key:2_1}
\mathbb{E}[\|\bm{\omega}_t\|_2]
\geq
\frac{c_t \sigma}{\sqrt{b_t}}
- 
\rho \left( \frac{1}{b_t} + \frac{1}{n} \right) \sum_{i \in [n]} L_i - |\alpha| G_{\perp}.
\end{align}
Furthermore, when $A_t < 0$, we have 
\begin{align*}
\|\hat{\bm{\omega}}_t\|_2
&\geq 
\Bigg\| \int_0^1 \nabla^2 f_{S_t} \left(\bm{x}_t + \rho s \frac{\nabla f_{S_t} (\bm{x}_t)}{\|\nabla f_{S_t} (\bm{x}_t)\|_2} \right) \rho \frac{\nabla f_{S_t} (\bm{x}_t)}{\|\nabla f_{S_t} (\bm{x}_t)\|_2} \mathrm{d}s\\
&\quad -
\int_0^1 \nabla^2 f_{S} \left(\bm{x}_t + \rho s \frac{\nabla f_{S} (\bm{x}_t)}{\|\nabla f_{S} (\bm{x}_t)\|_2} \right) \rho \frac{\nabla f_{S} (\bm{x}_t)}{\|\nabla f_{S} (\bm{x}_t)\|_2}
\mathrm{d}s 
\Bigg\|_2
- \|\nabla f_{S_t} (\bm{x}_t) - \nabla f_{S} (\bm{x}_t)\|_2\\
&\geq 
\Bigg |
\Bigg\| \int_0^1 \nabla^2 f_{S_t} \left(\bm{x}_t + \rho s \frac{\nabla f_{S_t} (\bm{x}_t)}{\|\nabla f_{S_t} (\bm{x}_t)\|_2} \right) \rho \frac{\nabla f_{S_t} (\bm{x}_t)}{\|\nabla f_{S_t} (\bm{x}_t)\|_2} \mathrm{d}s \Bigg\|_2\\
&\quad -
\Bigg\| \int_0^1 \nabla^2 f_{S} \left(\bm{x}_t + \rho s \frac{\nabla f_{S} (\bm{x}_t)}{\|\nabla f_{S} (\bm{x}_t)\|_2} \right) \rho \frac{\nabla f_{S} (\bm{x}_t)}{\|\nabla f_{S} (\bm{x}_t)\|_2}
\mathrm{d}s 
\Bigg\|_2 \Bigg| - \|\nabla f_{S_t} (\bm{x}_t) - \nabla f_{S} (\bm{x}_t)\|_2,
\end{align*}
which, together with (\ref{key_1_1}) and (\ref{key_2_1}), implies that there exists $d_t \in (0,1]$ such that
\begin{align*}
\|\hat{\bm{\omega}}_t\|_2
\geq 
\rho \left( \frac{d_t}{b_t} - \frac{1}{n}  \right) \sum_{i\in [n]} L_i - \|\nabla f_{S_t} (\bm{x}_t) - \nabla f_{S} (\bm{x}_t)\|_2.
\end{align*}
A similar discussion to the one in (\ref{key_4}), together with Proposition \ref{prop:1}, implies that
\begin{align*}
\mathbb{E}[\|\hat{\bm{\omega}}_t\|_2]
\geq 
\rho \left( \frac{d_t}{b_t} - \frac{1}{n}  \right) \sum_{i\in [n]} L_i 
- \frac{\sigma}{\sqrt{b_t}}.
\end{align*}
Hence, 
\begin{align}\label{key:2_2}
\mathbb{E}[\|\bm{\omega}_t\|_2]
\geq
\rho \left( \frac{d_t}{b_t} - \frac{1}{n}  \right) \sum_{i\in [n]} L_i 
- \frac{\sigma}{\sqrt{b_t}} - |\alpha| G_{\perp}.
\end{align}
Suppose that $\bm{x}_t$ generated by Algorithm \ref{algo:1} satisfies $\nabla f_{S_t} (\bm{x}_t) = \bm{0}$ or $\nabla f_{S} (\bm{x}_t) = \bm{0}$. Then, a discussion similar to the one proving Theorem \ref{thm:1} under $\nabla f_{S_t} (\bm{x}_t) = \bm{0} \lor \nabla f_{S} (\bm{x}_t) = \bm{0}$ ensures that (\ref{key:2_1}) and (\ref{key:2_2}) hold. When $b_t = n$, we have $\bm{\omega}_t = \nabla \hat{f}_{S,\rho}^{\mathrm{SAM}}(\bm{x}_t) - \nabla \hat{f}_{S,\rho}^{\mathrm{SAM}}(\bm{x}_t) + \alpha \nabla f_{S \perp}(\bm{x}_t) = \alpha \nabla f_{S \perp}(\bm{x}_t)$, which implies that $\mathbb{E}[\|\bm{\omega}_t\|_2] = |\alpha| \mathbb{E}[\|\nabla f_{S \perp}(\bm{x}_t)\|_2]$.

\section{General convergence analysis of GSAM and its proof}
\label{appendix_b}

\begin{theorem}
\label{thm:3}
\textup{\textbf{($\epsilon$--approximation of GSAM with an increasing batch size and decaying learning rate)}} Consider the sequence $(\bm{x}_t)$ generated by the mini-batch GSAM algorithm (Algorithm \ref{algo:1}) with an increasing batch size $b_t \in (0,n]$ and a decaying learning rate $\eta_t \in [\underline{\eta}, \overline{\eta}] \subset [0,+\infty)$ in the case that there are positive numbers $H_1 (\underline{\eta}, \overline{\eta})$, $H_2 (\underline{\eta}, \overline{\eta})$, and $H_3 (\underline{\eta}, \overline{\eta})$ such that, for all $T \geq 1$,
\begin{align}\label{eta_thm_3_1}
\frac{T}{\sum_{t=0}^{T-1} \eta_t} \leq H_1 (\underline{\eta}, \overline{\eta}) 
\quad\text{and}\quad 
\frac{\sum_{t=0}^{T-1} \eta_t^2}{\sum_{t=0}^{T-1} \eta_t} 
\leq H_2 (\underline{\eta}, \overline{\eta}) + \frac{H_3 (\underline{\eta}, \overline{\eta})}{T}.
\end{align}
Let us assume that there exists a positive number $G$ such that $\max \{\sup_{t\in \mathbb{N} \cup \{0\}} \|\nabla f_S (\bm{x}_{t} + \hat{\bm{\epsilon}}_{S_{t},\rho}(\bm{x}_{t}))\|_2, \sup_{t\in \mathbb{N} \cup \{0\}} \|\nabla \hat{f}^{\mathrm{SAM}}_{S_t,\rho} (\bm{x}_t)\|_2, \sup_{t\in \mathbb{N} \cup \{0\}} \|\nabla \hat{f}^{\mathrm{SAM}}_{S,\rho} (\bm{x}_t)\|_2, G_{\perp} \} \leq G$, where $G_{\perp} := \sup_{t\in \mathbb{N}\cup \{0\}}\|\nabla f_{S_t \perp}(\bm{x}_t)\|_2 < + \infty$ (Theorem \ref{thm:1}). Let $\epsilon > 0$ be the precision and let $b_0 > 0$, $\alpha \in \mathbb{R}$, and $\rho \geq 0$ such that 
{
\begin{align}
&H_1 
\leq
\frac{\epsilon^2}{12 \sigma \rho C } \left( \frac{G}{\sqrt{b_0}} + \frac{3 \sigma \rho C }{nb_0} \sum_{i\in [n]} L_i  \right)^{-1}, 
\text{ } 
(|\alpha| + 1)^2 H_2 \leq 
\frac{n^3 \epsilon^2}{6 G^2 \sum_{i\in [n]} L_i \{n^2 + 4\rho^2 C^2(\sum_{i\in [n]} L_i)^2 \}}, \nonumber\\
&\rho (|\alpha| + 1) 
\leq 
\frac{n \sqrt{b_0} \epsilon^2}{6G (\sum_{i\in [n]} L_i)(CG \sqrt{b_0}+ B \sigma)}, 
\text{ }
\rho^2
\leq 
\frac{n^2 b_0^2 \epsilon^4}{576 G^2(n^2 + b_0^2)(\sum_{i \in [n]} L_i)^2}, \label{eta_thm_3_2}
\end{align}
}
where $B$ and $C$ are nonnegative constants. Then, there exists $t_0 \in \mathbb{N}$ such that, for all $T \geq t_0$, 
\begin{align*}
&\min_{t \in [0:T-1]}\mathbb{E}\left[ \left\|\nabla \hat{f}_{S,\rho}^{\mathrm{SAM}}(\bm{x}_t) \right\|_2 \right]
\leq \epsilon.
\end{align*}
\end{theorem} 

Let us start with a brief outline of the proof strategy of Theorem \ref{thm:3}, with an emphasis on the main difficulty that has to be overcome. The flow of our proof is almost the same as in Theorem 5.1 of \citep{zhuang2022surrogate}, indicating that GSAM using a decaying learning rate, $\eta_t = \eta_0 /\sqrt{t}$, and a perturbation amplitude, $\rho_t = \rho_0 /\sqrt{t}$, proportional to $\eta_t$ satisfies
\begin{align*}
\frac{1}{T} \sum_{t=1}^{T} \mathbb{E}\left[ \left\|\nabla \hat{f}_{S,\rho_t}^{\mathrm{SAM}}(\bm{x}_t) \right\|_2^2 \right]
\leq \frac{C_1 + C_2 \log T}{\sqrt{T}},
\end{align*}
where $C_1$ and $C_2$ are positive constants. First, from the smoothness condition (A1) of $f_S$ and the descent lemma, we prove the inequality (Proposition \ref{prop:b_1}) that is satisfied for GSAM. Next, using the Cauchy--Schwarz inequality and the triangle inequality, we provide upper bounds of the terms $X_t$ (Proposition \ref{prop:x_t}), $Y_t$ (Proposition \ref{prop:y_t}), and $Z_t$ (Proposition \ref{prop:z_t}) in Proposition \ref{prop:b_1}. The main issue in Theorem \ref{thm:3} is to evaluate the full gradient $\nabla \hat{f}_{S,\rho}^{\mathrm{SAM}}(\bm{x}_t)$ using the mini-batch gradient $\nabla \hat{f}_{S_t,\rho}^{\mathrm{SAM}}(\bm{x}_t)$. The difficulty comes from the fact that the unbiasedness of $\nabla \hat{f}_{S_t,\rho}^{\mathrm{SAM}}(\bm{x}_t)$ does not hold (i.e., $\mathbb{E}[\nabla \hat{f}_{S_t,\rho}^{\mathrm{SAM}}(\bm{x}_t)] \neq \nabla \hat{f}_{S,\rho}^{\mathrm{SAM}}(\bm{x}_t)$, although (A2)(i) holds). However, we can resolve this issue using Theorem \ref{thm:1}. In fact, in order to evaluate the upper bound of $X_t$, we can use Theorem \ref{thm:1} indicating the upper bound of $\|\hat{\bm{\omega}}_t\|_2 = \|\nabla \hat{f}_{S,\rho}^{\mathrm{SAM}}(\bm{x}_t) - \nabla \hat{f}_{S_t,\rho}^{\mathrm{SAM}}(\bm{x}_t)\|_2$. Another issue that has to be overcome in order to prove Theorem \ref{thm:3} is to evaluate the upper bound of $\min_{t \in [0:T-1]}\mathbb{E}[ \|\nabla \hat{f}_{S,\rho}^{\mathrm{SAM}}(\bm{x}_t) \|_2^2 ]$ using a learning rate $\eta_t \in [\underline{\eta}, \overline{\eta}]$. We can resolve this issue by using $\min_{t \in [0:T-1]}\mathbb{E}[ \|\nabla \hat{f}_{S,\rho}^{\mathrm{SAM}}(\bm{x}_t) \|_2^2 ] \leq \sum_{t=0}^{T-1} \eta_t \mathbb{E}[ \|\nabla \hat{f}_{S,\rho}^{\mathrm{SAM}}(\bm{x}_t) \|_2^2 ]/\sum_{t=0}^{T-1} \eta_t$. As a result, we can provide an upper bound of $\min_{t \in [0:T-1]}\mathbb{E}[ \|\nabla \hat{f}_{S,\rho}^{\mathrm{SAM}}(\bm{x}_t) \|_2^2 ]$. Finally, we set $H_1$, $H_2$, $\alpha$, and $\rho$ such that the upper bound of $\min_{t \in [0:T-1]}\mathbb{E}[ \|\nabla \hat{f}_{S,\rho}^{\mathrm{SAM}}(\bm{x}_t) \|_2^2 ]$ is less than or equal to $\epsilon^2$.

\subsection{Lemma and propositions}
The following lemma, called the descent lemma, holds.

\begin{lemma}
[Descent lemma]
\label{lem:descent}\citep[Lemma 5.7]{beck2017} 
Let $f \colon \mathbb{R}^d \to \mathbb{R}$ be $L$--smooth. Then, we have that, for all $\bm{x},\bm{y} \in \mathbb{R}^d$, 
\begin{align*}
f(\bm{y}) \leq f(\bm{x}) + \langle \nabla f(\bm{x}), \bm{y} - \bm{x} \rangle_2 + \frac{L}{2}\|\bm{y} - \bm{x}\|_2^2.
\end{align*}
\end{lemma}

Lemma \ref{lem:descent} leads to the following proposition.

\begin{proposition}\label{prop:b_1}
Under Assumption \ref{assum:1}, we have that, for all $t \in \mathbb{N} \cup \{0\}$,
\begin{align*}
&f_S (\bm{x}_{t+1} + \hat{\bm{\epsilon}}_{S_{t+1},\rho}(\bm{x}_{t+1}))\\
&\leq 
f_S (\bm{x}_{t} + \hat{\bm{\epsilon}}_{S_{t},\rho}(\bm{x}_{t}))\\
&\quad + 
\eta_t \underbrace{\left\langle \nabla f_S (\bm{x}_{t} + \hat{\bm{\epsilon}}_{S_{t},\rho}(\bm{x}_{t})), \bm{d}_{t} \right\rangle_2}_{X_t} 
+
\underbrace{\left\langle \nabla f_S (\bm{x}_{t} + \hat{\bm{\epsilon}}_{S_{t},\rho}(\bm{x}_{t})),
\hat{\bm{\epsilon}}_{S_{t+1},\rho}(\bm{x}_{t+1})
- \hat{\bm{\epsilon}}_{S_{t},\rho}(\bm{x}_{t}) \right\rangle_2}_{Y_t}\\
&\quad + \frac{\sum_{i\in[n]} L_i}{n} 
\underbrace{\left\{ \eta_t^2 \|\bm{d}_{t}\|_2^2 + \left\|\hat{\bm{\epsilon}}_{S_{t+1},\rho}(\bm{x}_{t+1})
-  \hat{\bm{\epsilon}}_{S_{t},\rho}(\bm{x}_{t}) \right\|_2^2 \right\}}_{Z_t}.
\end{align*}
\end{proposition}

{\em Proof of Proposition \ref{prop:b_1}:} The $L_i$--smoothness (A1) of $f_i$ and the definition of $f_S$ ensure that, for all $\bm{x},\bm{y} \in \mathbb{R}^d$,
\begin{align*}
\| \nabla f_S (\bm{x}) - \nabla f_S (\bm{y}) \|_2
&= \left\| \frac{1}{n}\sum_{i\in[n]} (\nabla f_i(\bm{x}) - \nabla f_i(\bm{y})) \right\|_2
\leq 
\frac{1}{n} \sum_{i\in[n]} \| \nabla f_i(\bm{x}) - \nabla f_i(\bm{y}) \|_2\\
&\leq 
\frac{1}{n} \sum_{i\in[n]} L_i \| \bm{x} - \bm{y} \|_2,
\end{align*}
which implies that $f_S$ is $(1/n) \sum_{i\in[n]} L_i$--smooth. Lemma \ref{lem:descent} thus guarantees that, for all $t \in \mathbb{N} \cup \{0\}$,
\begin{align*}
&f_S (\bm{x}_{t+1} + \hat{\bm{\epsilon}}_{S_{t+1},\rho}(\bm{x}_{t+1}))\\
&\leq 
f_S (\bm{x}_{t} + \hat{\bm{\epsilon}}_{S_{t},\rho}(\bm{x}_{t}))
+ 
\left\langle \nabla f_S (\bm{x}_{t} + \hat{\bm{\epsilon}}_{S_{t},\rho}(\bm{x}_{t})), (\bm{x}_{t+1} - \bm{x}_{t}) + (\hat{\bm{\epsilon}}_{S_{t+1},\rho}(\bm{x}_{t+1})
-  \hat{\bm{\epsilon}}_{S_{t},\rho}(\bm{x}_{t})) \right\rangle_2\\
&\quad + \frac{\sum_{i\in[n]} L_i}{2 n} 
\left\|(\bm{x}_{t+1} - \bm{x}_{t}) + (\hat{\bm{\epsilon}}_{S_{t+1},\rho}(\bm{x}_{t+1})
-  \hat{\bm{\epsilon}}_{S_{t},\rho}(\bm{x}_{t})) \right\|_2^2,
\end{align*}
which, together with $\|\bm{x} + \bm{y} \|_2^2 \leq 2(\|\bm{x}\|_2^2 + \|\bm{y}\|_2^2)$ and $\bm{x}_{t+1} - \bm{x}_t = \eta_t \bm{d}_t$, implies that
\begin{align*}
&f_S (\bm{x}_{t+1} + \hat{\bm{\epsilon}}_{S_{t+1},\rho}(\bm{x}_{t+1}))\\
&\leq 
f_S (\bm{x}_{t} + \hat{\bm{\epsilon}}_{S_{t},\rho}(\bm{x}_{t}))\\
&\quad + 
\eta_t \left\langle \nabla f_S (\bm{x}_{t} + \hat{\bm{\epsilon}}_{S_{t},\rho}(\bm{x}_{t})), \bm{d}_{t} \right\rangle_2 
+
\langle \nabla f_S (\bm{x}_{t} + \hat{\bm{\epsilon}}_{S_{t},\rho}(\bm{x}_{t})),
\hat{\bm{\epsilon}}_{S_{t+1},\rho}(\bm{x}_{t+1})
- \hat{\bm{\epsilon}}_{S_{t},\rho}(\bm{x}_{t}) \rangle_2\\
&\quad + \frac{\sum_{i\in[n]} L_i}{n} 
\left\{ \eta_t^2 \|\bm{d}_{t}\|_2^2 + \left\|\hat{\bm{\epsilon}}_{S_{t+1},\rho}(\bm{x}_{t+1})
-  \hat{\bm{\epsilon}}_{S_{t},\rho}(\bm{x}_{t}) \right\|_2^2 \right\},
\end{align*}
which completes the proof.

Using Theorem \ref{thm:1}, we provide an upper bound of $\mathbb{E}[X_t]$.

\begin{proposition}\label{prop:x_t}
Suppose that Assumption \ref{assum:1} holds and there exist $G > 0$ and $G_{\perp} > 0$ such that $\max \{\sup_{t\in \mathbb{N} \cup \{0\}} \|\nabla \hat{f}^{\mathrm{SAM}}_{S_t,\rho} (\bm{x}_t)\|_2, \sup_{t\in \mathbb{N} \cup \{0\}} \|\nabla \hat{f}^{\mathrm{SAM}}_{S,\rho} (\bm{x}_t)\|_2 \} \leq G$ and $\sup_{t\in \mathbb{N}\cup \{0\}}\|\nabla f_{S_t \perp}(\bm{x}_t)\|_2 \leq G_{\perp}$. Then, for all $t\in \mathbb{N}\cup \{0\}$,
\begin{align*}
\mathbb{E}[X_t]
&\leq 
- \mathbb{E}\left[ \left\|\nabla \hat{f}_{S,\rho}^{\mathrm{SAM}}(\bm{x}_t) \right\|_2^2 \right]
+ 
G 
\sqrt{4 \rho^2 \left( \frac{1}{b_t^2} + \frac{1}{n^2} \right)
\big(\sum_{i \in [n]} L_i \big)^2 
+ 2 \sigma_t^2} \\
&\quad + 
(G + |\alpha| G_{\perp})
\frac{\rho B \sigma}{n \sqrt{b_t}} \sum_{i\in [n]} L_i,
\end{align*}
where $\sigma_t^2 := \mathbb{E}[\|\nabla f_{S_t} (\bm{x}_t) - \nabla f_S (\bm{x}_t)\|_2^2] \leq \sigma^2/b_t$ and $B \geq 0$ is a constant.
\end{proposition}

{\em Proof:} Let $t \in \mathbb{N} \cup \{0\}$ and $b_t < n$. The definition of $\bm{d}_t = - (\nabla \hat{f}_{S_t,\rho}^{\mathrm{SAM}}(\bm{x}_t) - \alpha \nabla f_{S_t \perp}(\bm{x}_t))$ implies that
\begin{align}\label{x_t}
X_t = 
\underbrace{- \left\langle \nabla f_S (\bm{x}_{t} + \hat{\bm{\epsilon}}_{S_{t},\rho}(\bm{x}_{t})), \nabla \hat{f}_{S_t,\rho}^{\mathrm{SAM}}(\bm{x}_t) \right\rangle_2}_{X_{t,1}}
+ \underbrace{\alpha \left\langle \nabla f_S (\bm{x}_{t} + \hat{\bm{\epsilon}}_{S_{t},\rho}(\bm{x}_{t})), \nabla f_{S_t \perp}(\bm{x}_t) \right\rangle_2}_{X_{t,2}}.
\end{align}
Then, we have 
\begin{align}\label{ineq_x_t_1}
\begin{split}
X_{t,1}
&= 
- \bigg\{\left\|\nabla \hat{f}_{S,\rho}^{\mathrm{SAM}}(\bm{x}_t) \right\|_2^2
+ 
\left\langle \nabla f_S (\bm{x}_{t} + \hat{\bm{\epsilon}}_{S_{t},\rho}(\bm{x}_{t}))
- \nabla \hat{f}_{S,\rho}^{\mathrm{SAM}}(\bm{x}_t), 
\nabla \hat{f}_{S_t,\rho}^{\mathrm{SAM}}(\bm{x}_t)
\right\rangle_2\\
&\quad + 
\Big\langle \nabla \hat{f}_{S,\rho}^{\mathrm{SAM}}(\bm{x}_t),
\underbrace{\nabla \hat{f}_{S_t,\rho}^{\mathrm{SAM}}(\bm{x}_t) 
- \nabla \hat{f}_{S,\rho}^{\mathrm{SAM}}(\bm{x}_t)}_{- \hat{\bm{\omega}}_t}
\Big\rangle_2 \bigg\}\\
&\leq 
- \left\|\nabla \hat{f}_{S,\rho}^{\mathrm{SAM}}(\bm{x}_t) \right\|_2^2
+ 
\underbrace{\left\| \nabla f_S (\bm{x}_{t} + \hat{\bm{\epsilon}}_{S_{t},\rho}(\bm{x}_{t}))
- \nabla \hat{f}_{S,\rho}^{\mathrm{SAM}}(\bm{x}_t) \right\|_2}_{X_{t,3}} 
\left\|\nabla \hat{f}_{S_t,\rho}^{\mathrm{SAM}}(\bm{x}_t) \right\|_2\\
&\quad + 
\left\| \nabla \hat{f}_{S,\rho}^{\mathrm{SAM}}(\bm{x}_t) \right\|_2 
\|\hat{\bm{\omega}}_t \|_2,
\end{split}
\end{align}
where the second inequality comes from the Cauchy--Schwarz inequality. Suppose that $\nabla f_{S_t} (\bm{x}_t) \neq \bm{0}$ and $\nabla f_{S} (\bm{x}_t) \neq \bm{0}$. The $(1/n) \sum_{i\in [n]} L_i$--smoothness of $f_S$ implies that
\begin{align*}
X_{t,3} 
&= 
\left\|
\nabla f_S \left(\bm{x}_{t} + \rho \frac{\nabla f_{S_t} (\bm{x}_t)}{\|\nabla f_{S_t} (\bm{x}_t)\|_2} \right) 
- 
\nabla f_S \left(\bm{x}_{t} + \rho \frac{\nabla f_{S} (\bm{x}_t)}{\|\nabla f_{S} (\bm{x}_t)\|_2} \right)
\right\|_2\\
&\leq
\frac{\rho}{n} \sum_{i\in [n]} L_i
\left\| \frac{\nabla f_{S_t} (\bm{x}_t)}{\|\nabla f_{S_t} (\bm{x}_t)\|_2} - \frac{\nabla f_S (\bm{x}_t)}{\|\nabla f_S (\bm{x}_t)\|_2} \right\|_2. 
\end{align*}
The discussion in \citep[Pages 15 and 16]{zhuang2022surrogate} implies there exists $B_t \geq 0$ such that 
\begin{align}\label{b_t}
&\left\| \frac{\nabla f_{S_t} (\bm{x}_t)}{\|\nabla f_{S_t} (\bm{x}_t)\|_2} - \frac{\nabla f_S (\bm{x}_t)}{\|\nabla f_S (\bm{x}_t)\|_2} \right\|_2
\leq 
B_t \left\| \nabla f_{S_t} (\bm{x}_t) - \nabla f_S (\bm{x}_t) \right\|_2
\end{align}
Let $B := \sup_{t \in \mathbb{N} \cup \{0\}} B_t$. Then, Proposition \ref{prop:1} ensures that
\begin{align}\label{ineq_x_t_3_1}
\mathbb{E}[X_{t,3}]
\leq 
\frac{\rho B \sigma}{n\sqrt{b_t}} \sum_{i\in [n]} L_i.
\end{align}
Suppose that $\nabla f_{S_t} (\bm{x}_t) = \bm{0}$ or $\nabla f_{S} (\bm{x}_t) = \bm{0}$. Let $\nabla f_{S_t} (\bm{x}_t) = \bm{0}$. The $(1/n) \sum_{i\in [n]} L_i$--smoothness of $f_S$ ensures that
\begin{align*}
X_{t,3} 
&= 
\left\|
\nabla f_S \left(\bm{x}_{t} + \bm{u} \right) 
- 
\nabla f_S \left(\bm{x}_{t} + \rho \frac{\nabla f_{S} (\bm{x}_t)}{\|\nabla f_{S} (\bm{x}_t)\|_2} \right)
\right\|_2
\leq
\frac{1}{n} \sum_{i\in [n]} L_i
\left\| \bm{u} - \rho \frac{\nabla f_S (\bm{x}_t)}{\|\nabla f_S (\bm{x}_t)\|_2} \right\|_2,
\end{align*}
which, together with $\|\bm{u}\|_2 \leq \rho$, implies there exists $C_t \geq 0$ such that 
\begin{align*}
X_{t,3}
\leq 
\frac{\rho C_t}{n} \sum_{i\in [n]} L_i
\left\| \frac{\nabla f_{S_t} (\bm{x}_t)}{\|\nabla f_{S_t} (\bm{x}_t)\|_2} - \frac{\nabla f_S (\bm{x}_t)}{\|\nabla f_S (\bm{x}_t)\|_2} \right\|_2.
\end{align*}
Hence, Proposition \ref{prop:1} implies that (\ref{ineq_x_t_3_1}) holds. A discussion similar to the case where $\nabla f_{S_t} (\bm{x}_t) = \bm{0}$ ensures that (\ref{ineq_x_t_3_1}) holds for $\nabla f_{S} (\bm{x}_t) = \bm{0}$. Taking the total expectation on both sides of (\ref{ineq_x_t_1}), together with (\ref{ineq_x_t_3_1}) and Theorem \ref{thm:1}, yields 
\begin{align}\label{x_t_1}
\mathbb{E}[X_{t,1}]
&\leq 
- \mathbb{E}\left[ \left\|\nabla \hat{f}_{S,\rho}^{\mathrm{SAM}}(\bm{x}_t) \right\|_2^2 \right]
+ 
G 
\sqrt{4 \rho^2 \left( \frac{1}{b_t^2} + \frac{1}{n^2} \right)
\big(\sum_{i \in [n]} L_i \big)^2 
+ 2 \sigma_t^2}\nonumber \\
&\quad + 
\frac{\rho BG \sigma}{n \sqrt{b_t}} \sum_{i\in [n]} L_i.
\end{align}
The Cauchy--Schwarz inequality implies that 
\begin{align*}
X_{t,2} 
&= \alpha \left\langle \nabla f_S (\bm{x}_{t} + \hat{\bm{\epsilon}}_{S_{t},\rho}(\bm{x}_{t})) - \nabla \hat{f}^{\mathrm{SAM}}_{S,\rho}(\bm{x}_t) + \nabla \hat{f}^{\mathrm{SAM}}_{S,\rho}(\bm{x}_t), \nabla f_{S_t \perp}(\bm{x}_t) \right\rangle_2\\
&\leq
|\alpha| X_{t,3} \left\|\nabla f_{S_t \perp}(\bm{x}_t) \right\|_2
+ \alpha \left\langle \nabla \hat{f}^{\mathrm{SAM}}_{S,\rho}(\bm{x}_t), \nabla f_{S_t \perp}(\bm{x}_t) \right\rangle_2\\
&\leq 
|\alpha| G_{\perp} X_{t,3} 
+ \alpha \left\langle \nabla \hat{f}^{\mathrm{SAM}}_{S,\rho}(\bm{x}_t), \nabla f_{S_t \perp}(\bm{x}_t) \right\rangle_2,
\end{align*}
which, together with $\mathbb{E}_{\bm{\xi}_t} [\nabla f_{S_t \perp}(\bm{x}_t)|\bm{\xi}_{t-1}] = \nabla f_{S \perp} (\bm{x}_t)$, $\langle \nabla \hat{f}^{\mathrm{SAM}}_{S,\rho}(\bm{x}_t), \nabla f_{S \perp} (\bm{x}_t) \rangle_2 = 0$ and (\ref{ineq_x_t_3_1}), implies that 
\begin{align}\label{x_t_2}
\mathbb{E}[X_{t,2}]
\leq
\frac{|\alpha| \rho B G_{\perp}  \sigma}{n \sqrt{b_t}} \sum_{i\in [n]} L_i.
\end{align}
Accordingly, (\ref{x_t}), (\ref{x_t_1}), and (\ref{x_t_2}) guarantee that
\begin{align*}
\mathbb{E}[X_t]
&\leq 
- \mathbb{E}\left[ \left\|\nabla \hat{f}_{S,\rho}^{\mathrm{SAM}}(\bm{x}_t) \right\|_2^2 \right]
+ 
G 
\sqrt{4 \rho^2 \left( \frac{1}{b_t^2} + \frac{1}{n^2} \right)
\big(\sum_{i \in [n]} L_i \big)^2 
+ 2 \sigma_t^2} \nonumber \\
&\quad + 
(G + |\alpha| G_{\perp})
\frac{\rho B \sigma}{n \sqrt{b_t}} \sum_{i\in [n]} L_i,
\end{align*}
which completes the proof. 

\begin{proposition}\label{prop:y_t}
Suppose that the assumptions in Proposition \ref{prop:x_t} hold and there exists $G > 0$ such that $\max \{ \sup_{t\in \mathbb{N} \cup \{0\}} \|\nabla f_S (\bm{x}_{t} + \hat{\bm{\epsilon}}_{S_{t},\rho}(\bm{x}_{t}))\|_2, \sup_{t\in \mathbb{N} \cup \{0\}} \|\nabla \hat{f}^{\mathrm{SAM}}_{S_t,\rho} (\bm{x}_t)\|_2, \\ \sup_{t\in \mathbb{N} \cup \{0\}} \|\nabla \hat{f}^{\mathrm{SAM}}_{S,\rho} (\bm{x}_t)\|_2, G_{\perp} \} \leq G$. Then, for all $t\in \mathbb{N}\cup \{0\}$,
\begin{align*}
\mathbb{E}[Y_{t}]
\leq
\rho C G
\left\{  
\frac{\eta_t (|\alpha| + 1) G}{n} \sum_{i\in [n]} L_i 
+ 
\frac{2 \sigma}{\sqrt{b_t}}
\right\}, 
\end{align*}
where $C \geq 0$ is a constant. 
\end{proposition}

{\em Proof:} Let $t \in \mathbb{N} \cup \{0\}$. The Cauchy--Schwarz inequality ensures that 
\begin{align}\label{y_t}
Y_t \leq 
G \left\| \hat{\bm{\epsilon}}_{S_{t+1},\rho}(\bm{x}_{t+1})
- \hat{\bm{\epsilon}}_{S_{t},\rho}(\bm{x}_{t}) \right \|_2 =: G Y_{t,1}.
\end{align}
Suppose that $\nabla f_{S_{t+1}} (\bm{x}_{t+1}) \neq \bm{0}$ and $\nabla f_{S_t} (\bm{x}_t) \neq \bm{0}$. The discussion in \citep[Pages 15 and 16]{zhuang2022surrogate} (see (\ref{b_t})) implies that there exists $C_t \geq 0$ such that 
\begin{align}\label{y_t_1}
&Y_{t,1}
= \rho \left\| \frac{\nabla f_{S_{t+1}} (\bm{x}_{t+1})}{\|\nabla f_{S_{t+1}} (\bm{x}_{t+1})\|_2} - \frac{\nabla f_{S_t} (\bm{x}_t)}{\|\nabla f_{S_t} (\bm{x}_t)\|_2} \right\|_2
\leq 
\rho C_t \left\| \nabla f_{S_{t+1}} (\bm{x}_{t+1}) - \nabla f_{S_t} (\bm{x}_t) \right\|_2.
\end{align}
Let $C := \sup_{t\in \mathbb{N} \cup \{0\}} C_t$. The triangle inequality gives
\begin{align*}
&\left\| \nabla f_{S_{t+1}} (\bm{x}_{t+1}) - \nabla f_{S_t} (\bm{x}_t) \right\|_2\\
&\leq
\left\| \nabla f_{S_{t+1}} (\bm{x}_{t+1}) - \nabla f_{S} (\bm{x}_{t+1}) \right\|_2
+
\left\|\nabla f_{S} (\bm{x}_{t+1}) - \nabla f_S (\bm{x}_t) \right\|_2
+
\left\|\nabla f_S (\bm{x}_t) - \nabla f_{S_t} (\bm{x}_t) \right\|_2,
\end{align*}
which, together with the $(1/n) \sum_{i\in [n]} L_i$--smoothness of $f_S$, $\bm{x}_{t+1} - \bm{x}_t = \eta_t \bm{d}_t$, (\ref{y_t}), and (\ref{y_t_1}), implies that
\begin{align*}
Y_{t,1} 
\leq 
\rho C \left\{  
\frac{\eta_t}{n} \sum_{i\in [n]} L_i \| \bm{d}_t \|_2
+ 
\left\| \nabla f_{S_{t+1}} (\bm{x}_{t+1}) - \nabla f_{S} (\bm{x}_{t+1}) \right\|_2
+
\left\|\nabla f_{S_t} (\bm{x}_t) - \nabla f_{S} (\bm{x}_t) \right\|_2
\right\}.
\end{align*}
Moreover, the Cauchy--Schwarz inequality and the definitions of $G$ and $G_{\perp}$ ensure that 
\begin{align}\label{d_t}
\begin{split}
\|\bm{d}_t \|_2^2
&= 
\left\|
\nabla \hat{f}_{S_t,\rho}^{\mathrm{SAM}}(\bm{x}_t) - \alpha \nabla f_{S_t \perp}(\bm{x}_t)
\right\|_2^2\\
&= 
\left\|
\nabla \hat{f}_{S_t,\rho}^{\mathrm{SAM}}(\bm{x}_t)
\right\|_2^2
-2 \alpha 
\left\langle 
\nabla \hat{f}_{S_t,\rho}^{\mathrm{SAM}}(\bm{x}_t), 
\nabla f_{S_t \perp}(\bm{x}_t)
\right\rangle_2
+ |\alpha|^2 \left\|
\nabla f_{S_t \perp}(\bm{x}_t) 
\right\|_2^2\\
&\leq
G^2 + 2 |\alpha| G G_{\perp} + |\alpha|^2 G_{\perp}^2
\leq (|\alpha| + 1)^2 G^2.
\end{split} 
\end{align}
Accordingly, we have 
\begin{align*}
Y_{t,1}
\leq 
\rho C
\left\{  
\frac{\eta_t (|\alpha| + 1) G}{n} \sum_{i\in [n]} L_i 
+ 
\left\| \nabla f_{S_{t+1}} (\bm{x}_{t+1}) - \nabla f_{S} (\bm{x}_{t+1}) \right\|_2
+
\left\|\nabla f_{S_t} (\bm{x}_t) - \nabla f_{S} (\bm{x}_t) \right\|_2
\right\}, 
\end{align*}
which, together with Proposition \ref{prop:1}, guarantees that 
\begin{align}\label{y_t_1_1}
\mathbb{E}[Y_{t,1}]
\leq
\rho C
\left\{  
\frac{\eta_t (|\alpha| + 1) G}{n} \sum_{i\in [n]} L_i 
+ 
\frac{2 \sigma}{\sqrt{b_t}}
\right\}.
\end{align}
Hence, from (\ref{y_t}),
\begin{align*}
\mathbb{E}[Y_{t}]
\leq
\rho C G
\left\{  
\frac{\eta_t (|\alpha| + 1) G}{n} \sum_{i\in [n]} L_i 
+ 
\frac{2 \sigma}{\sqrt{b_t}}
\right\}.
\end{align*}
We can show that Proposition \ref{prop:y_t} holds for the case where $\nabla f_{S_{t+1}} (\bm{x}_{t+1}) = \bm{0}$ or $\nabla f_{S_t} (\bm{x}_t) = \bm{0}$ by proving Proposition \ref{prop:x_t}.

\begin{proposition}\label{prop:z_t}
Suppose that the assumptions in Proposition \ref{prop:y_t} hold. Then, for all $t\in \mathbb{N}\cup \{0\}$,
{
\begin{align*}
\mathbb{E}[Z_{t}]
\leq
\eta_t^2 (|\alpha| + 1)^2 G^2 \left\{ 1 + \frac{4\rho^2 C^2}{n^2}\bigg(\sum_{i\in [n]} L_i \bigg)^2 \right\}
+ \frac{6 \rho^2 C^2 \sigma^2}{b_t}.
\end{align*}
}
\end{proposition}

{\em Proof:} Let $t\in \mathbb{N}\cup \{0\}$. From (\ref{d_t}), we have 
\begin{align*}
\eta_t^2 \mathbb{E}[\|\bm{d}_t\|_2]
\leq 
\eta_t^2 (|\alpha| + 1)^2 G^2.
\end{align*}
Suppose that $\nabla f_{S_{t+1}} (\bm{x}_{t+1}) \neq \bm{0}$ and $\nabla f_{S_t} (\bm{x}_t) \neq \bm{0}$. Then, from $\|\bm{x} + \bm{y} \|_2^2 \leq 2(\|\bm{x}\|_2^2 + \|\bm{y}\|_2^2)$,
\begin{align*}
&\left\| \nabla f_{S_{t+1}} (\bm{x}_{t+1}) - \nabla f_{S_t} (\bm{x}_t) \right\|_2^2\\
&\leq
2 \left\| \nabla f_{S_{t+1}} (\bm{x}_{t+1}) - \nabla f_{S} (\bm{x}_{t+1}) \right\|_2^2
+
4\left\|\nabla f_{S} (\bm{x}_{t+1}) - \nabla f_S (\bm{x}_t) \right\|_2^2
+
4\left\|\nabla f_S (\bm{x}_t) - \nabla f_{S_t} (\bm{x}_t) \right\|_2^2.
\end{align*}
A discussion similar to the one showing (\ref{y_t_1_1}) ensures that
{
\begin{align*}
\mathbb{E}[Y_{t,1}^2] 
= \mathbb{E} \left[ \left\| \hat{\bm{\epsilon}}_{S_{t+1},\rho}(\bm{x}_{t+1})
- \hat{\bm{\epsilon}}_{S_{t},\rho}(\bm{x}_{t}) \right \|_2^2
\right]
\leq
2 \rho^2 C^2 \left\{ \frac{2 \eta_t^2 (|\alpha| + 1)^2 G^2}{n^2} \bigg(\sum_{i\in [n]} L_i \bigg)^2 + \frac{3 \sigma^2}{b_t} \right\}.
\end{align*}
}
An argument similar to the one used to prove Proposition \ref{prop:x_t} shows that the above inequality also holds for the case where $\nabla f_{S_{t+1}} (\bm{x}_{t+1}) = \bm{0}$ or $\nabla f_{S_t} (\bm{x}_t) = \bm{0}$. Hence, 
{
\begin{align*}
\mathbb{E}[Z_t]
&\leq
\eta_t^2 (|\alpha| + 1)^2 G^2
+ 
2 \rho^2 C^2 \left\{ \frac{2 \eta_t^2 (|\alpha| + 1)^2 G^2}{n^2} \bigg(\sum_{i\in [n]} L_i \bigg)^2 + \frac{3 \sigma^2}{b_t} \right\}\\
&=
\eta_t^2 (|\alpha| + 1)^2 G^2 \left\{ 1 + \frac{4\rho^2 C^2}{n^2}\bigg(\sum_{i\in [n]} L_i \bigg)^2 \right\}
+ \frac{6 \rho^2 C^2 \sigma^2}{b_t},
\end{align*}
}
which completes the proof. 

{\em Proof of Theorem \ref{thm:3}:}
Let us define $F_{\rho}(t) := f_S (\bm{x}_{t} + \hat{\bm{\epsilon}}_{S_{t},\rho}(\bm{x}_{t}))$. From Proposition \ref{prop:b_1}, Proposition \ref{prop:x_t}, Proposition \ref{prop:y_t}, and Proposition \ref{prop:z_t}, for all $t \in \mathbb{N} \cup \{0\}$, we have
{
\begin{align*}
&\mathbb{E}[F_{\rho}(t+1)]
\leq 
\mathbb{E}[F_{\rho}(t)]
+ 
\eta_t \mathbb{E}[X_t] 
+
\mathbb{E}[Y_t]
+
\frac{\sum_{i\in[n]} L_i}{n} \mathbb{E}[Z_t]\\
&\leq
\mathbb{E}[F_{\rho}(t)]
- 
\eta_t 
\mathbb{E}\left[ \left\|\nabla \hat{f}_{S,\rho}^{\mathrm{SAM}}(\bm{x}_t) \right\|_2^2 \right]
+ 
\eta_t G 
\sqrt{4 \rho^2 \left( \frac{1}{b_t^2} + \frac{1}{n^2} \right)
\bigg(\sum_{i \in [n]} L_i \bigg)^2 
+ 2 \sigma_t^2} \\
&\quad + 
\eta_t (|\alpha|+1)
\frac{\rho B G \sigma}{n \sqrt{b_t}} \sum_{i\in [n]} L_i
+ \rho C G
\left\{  
\frac{\eta_t (|\alpha| + 1) G}{n} \sum_{i\in [n]} L_i 
+ 
\frac{2 \sigma}{\sqrt{b_t}}
\right\}\\
&\quad + \frac{\sum_{i\in[n]} L_i}{n} \left[
\eta_t^2 (|\alpha| + 1)^2 G^2 \left\{ 1 + \frac{4\rho^2 C^2}{n^2}\bigg(\sum_{i\in [n]} L_i \bigg)^2 \right\}
+ \frac{6 \rho^2 C^2 \sigma^2}{b_t} \right],
\end{align*}
}
which implies that
{
\begin{align}\label{key_1}
\begin{split}
\eta_t \mathbb{E}\left[ \left\|\nabla \hat{f}_{S,\rho}^{\mathrm{SAM}}(\bm{x}_t) \right\|_2^2 \right]
&\leq 
\left( \mathbb{E}[F_{\rho}(t)] - \mathbb{E}[F_{\rho}(t+1)]  \right) 
+ 2  \sigma \rho C 
\left(  \frac{ G}{\sqrt{b_t}} + \frac{3 \sigma \rho C  }{n b_t} \sum_{i \in [n]} L_i \right)\\
&\quad + \frac{\eta_t^2 (|\alpha| + 1)^2 G^2}{n} 
 \sum_{i\in[n]} L_i
\left\{ 1 + \frac{4\rho^2 C^2}{n^2}\bigg(\sum_{i\in [n]} L_i \bigg)^2 \right\}\\
&\quad 
+ \eta_t G 
\sqrt{4 \rho^2 \left( \frac{1}{b_t^2} + \frac{1}{n^2} \right)
\big(\sum_{i \in [n]} L_i \big)^2 
+ 2 \sigma_t^2}\\
&\quad 
+ \eta_t \frac{\rho (|\alpha| + 1) G}{n} \sum_{i\in [n]} L_i
\left( CG + \frac{B \sigma}{\sqrt{b_t}}   \right).
\end{split}
\end{align}
}
Let $\epsilon > 0$. From $g (b_t) = \sigma_t^2 := \mathbb{E}[\|\nabla f_{S_t} (\bm{x}_t) - \nabla f_S (\bm{x}_t)\|_2^2] \leq \sigma^2/b_t$ ($t \in \mathbb{N} \cup \{0\}$) (see Proposition \ref{prop:1} and \citep[Theorem 8.6]{Freund:1971aa}) and $g (n) = 0$, the sequence $(b_t)$ of increasing batch sizes implies that there exists $t_0 \in \mathbb{N}$ such that, for all $t \geq t_0$,
{
\begin{align*}
2 \sigma_t^2 \leq \frac{\epsilon^4}{48 G^2}.
\end{align*}
}
Let $T \geq t_0 + 1$. Summing the above inequality from $t = 0$ to $t = T -1$, together with $b_0 \leq b_t$ and $\eta_t \leq \overline{\eta}$ ($t \in \mathbb{N} \cup \{0\}$), ensures that
{
\begin{align*}
\sum_{t=0}^{T-1} \eta_t \mathbb{E}\left[ \left\|\nabla \hat{f}_{S,\rho}^{\mathrm{SAM}}(\bm{x}_t) \right\|_2^2 \right]
&\leq 
\left( \mathbb{E}[F_{\rho}(0)] - f_S^\star \right) 
+ 2 \sigma \rho C   
\left(  \frac{G}{\sqrt{b_0}} + \frac{3 \sigma \rho C }{n b_0} \sum_{i \in [n]} L_i \right) T\\
&\quad + \frac{(|\alpha| + 1)^2 G^2}{n} 
 \sum_{i\in[n]} L_i
\left\{ 1 + \frac{4\rho^2 C^2}{n^2}\bigg(\sum_{i\in [n]} L_i \bigg)^2 \right\}
\sum_{t=0}^{T-1} \eta_t^2\\
&\quad 
+ G 
\sqrt{4 \rho^2 \left( \frac{1}{b_0^2} + \frac{1}{n^2} \right)
\big(\sum_{i \in [n]} L_i \big)^2 
+ \frac{2 \sigma^2}{b_0}}
t_0 \overline{\eta}\\
&\quad + G 
\sqrt{4 \rho^2 \left( \frac{1}{b_0^2} + \frac{1}{n^2} \right)
\big(\sum_{i \in [n]} L_i \big)^2 
+ \frac{\epsilon^4}{49 G^2}}
 \sum_{t=t_0}^{T-1} \eta_t\\ 
&\quad + \frac{\rho (|\alpha| + 1) G}{n} \sum_{i\in [n]} L_i
\left( CG + \frac{B \sigma}{\sqrt{b_0}}   \right) \sum_{t=0}^{T-1} \eta_t,
\end{align*}
}
where $f_S^\star$ is the minimum value of $f_S$ over $\mathbb{R}^d$. Since we have that 
\begin{align*}
\min_{t \in [0:T-1]}\mathbb{E}\left[ \left\|\nabla \hat{f}_{S,\rho}^{\mathrm{SAM}}(\bm{x}_t) \right\|_2^2 \right] 
\leq \frac{\sum_{t=0}^{T-1} \eta_t \mathbb{E}\left[ \left\|\nabla \hat{f}_{S,\rho}^{\mathrm{SAM}}(\bm{x}_t) \right\|_2^2 \right]}{\sum_{t=0}^{T-1} \eta_t},
\end{align*}
we also have that 
{
\begin{align}
\min_{t \in [0:T-1]}\mathbb{E}\left[ \left\|\nabla \hat{f}_{S,\rho}^{\mathrm{SAM}}(\bm{x}_t) \right\|_2^2 \right] 
&\leq 
\frac{\mathbb{E}[F_{\rho}(0)] - f_S^\star}{\sum_{t=0}^{T-1} \eta_t}
+ 2 \sigma \rho C   
\left(  \frac{ G}{\sqrt{b_0}} + \frac{3 \sigma\rho C}{n b_0} \sum_{i \in [n]} L_i \right) \frac{T}{\sum_{t=0}^{T-1} \eta_t} \nonumber\\
&\quad + \frac{(|\alpha| + 1)^2 G^2}{n} 
 \sum_{i\in[n]} L_i
\left\{ 1 + \frac{4\rho^2 C^2}{n^2}\bigg(\sum_{i\in [n]} L_i \bigg)^2 \right\}
\frac{\sum_{t=0}^{T-1} \eta_t^2}{\sum_{t=0}^{T-1} \eta_t} \nonumber\\
&\quad 
+ G 
\sqrt{4 \rho^2 \left( \frac{1}{b_0^2} + \frac{1}{n^2} \right)
\big(\sum_{i \in [n]} L_i \big)^2 
+ \frac{2 \sigma^2}{b_0}}
\frac{t_0 \overline{\eta}}{\sum_{t=0}^{T-1} \eta_t}\nonumber \\
&\quad 
+ G 
\sqrt{4 \rho^2 \left( \frac{1}{b_0^2} + \frac{1}{n^2} \right)
\big(\sum_{i \in [n]} L_i \big)^2 
+ \frac{\epsilon^4}{49 G^2}}\nonumber \\  
&\quad 
+ \frac{\rho (|\alpha| + 1) G}{n} \sum_{i\in [n]} L_i
\left( CG + \frac{B \sigma}{\sqrt{b_0}}   \right).\label{key_2}
\end{align}
}
From (\ref{eta_thm_3_1}), i.e.,
\begin{align*}
\frac{T}{\sum_{t=0}^{T-1} \eta_t} \leq H_1 (\underline{\eta}, \overline{\eta}) 
\quad\text{and}\quad 
\frac{\sum_{t=0}^{T-1} \eta_t^2}{\sum_{t=0}^{T-1} \eta_t} 
\leq H_2 (\underline{\eta}, \overline{\eta}) + \frac{H_3 (\underline{\eta}, \overline{\eta})}{T},
\end{align*}

we have that 
{
\begin{align*}
&\min_{t \in [0:T-1]}\mathbb{E}\left[ \left\|\nabla \hat{f}_{S,\rho}^{\mathrm{SAM}}(\bm{x}_t) \right\|_2^2 \right]\\ 
&\leq 
\underbrace{\frac{H_1(\mathbb{E}[F_{\rho}(0)] - f_S^\star)}{T}
+ G H_1
\sqrt{4 \rho^2 \left( \frac{1}{b_0^2} + \frac{1}{n^2} \right)
\big(\sum_{i \in [n]} L_i \big)^2 
+ \frac{2 \sigma^2}{b_0}}
\frac{t_0 \overline{\eta}}{T}}_{U_1 \leq \frac{\epsilon^2}{6}}\\
&\quad +
\underbrace{\frac{(|\alpha| + 1)^2 G^2}{n} 
 \sum_{i\in[n]} L_i
\left\{ 1 + \frac{4\rho^2 C^2}{n^2}\bigg(\sum_{i\in [n]} L_i \bigg)^2 \right\}
\frac{H_3}{T}}_{U_2 \leq \frac{\epsilon^2}{6}}\\
&\quad +  
\underbrace{\left( \frac{G}{\sqrt{b_0}} + \frac{3 \sigma \rho C }{n b_0} \sum_{i \in [n]} L_i \right) 2 \sigma \rho C   H_1}_{U_3 \leq \frac{\epsilon^2}{6}}
+ \underbrace{\frac{(|\alpha| + 1)^2 G^2 H_2}{n} 
 \sum_{i\in[n]} L_i
\left\{ 1 + \frac{4\rho^2 C^2}{n^2}\bigg(\sum_{i\in [n]} L_i \bigg)^2 \right\}}_{U_4 \leq \frac{\epsilon^2}{6}}\\
&\quad + \underbrace{\frac{\rho (|\alpha| + 1) G}{n} \sum_{i\in [n]} L_i
\left( CG + \frac{B \sigma}{\sqrt{b_0}}   \right)}_{U_5 \leq \frac{\epsilon^2}{6}} 
+ \underbrace{G 
\sqrt{4 \rho^2 \left( \frac{1}{b_0^2} + \frac{1}{n^2} \right)
\big(\sum_{i \in [n]} L_i \big)^2 
+ \frac{\epsilon^4}{48 G^2}}}_{U_6 \leq \frac{\epsilon^2}{6}}.
\end{align*}
}
It is guaranteed that there exists $t_1 \in \mathbb{N}$ such that, for all $T \geq \max \{t_0, t_1\}$, $U_1 \leq \frac{\epsilon^2}{6}$ and $U_2 \leq \frac{\epsilon^2}{6}$. Moreover, if (\ref{eta_thm_3_2}) holds, i.e., 
{
\begin{align*}
&H_1 
\leq
\frac{\epsilon^2}{12 \sigma \rho C } \left( \frac{G}{\sqrt{b_0}} + \frac{3 \sigma \rho C }{nb_0} \sum_{i\in [n]} L_i  \right)^{-1}, 
\text{ } 
(|\alpha| + 1)^2 H_2 \leq 
\frac{n^3 \epsilon^2}{6 G^2 \sum_{i\in [n]} L_i \{n^2 + 4\rho^2 C^2(\sum_{i\in [n]} L_i)^2 \}},\\
&\rho (|\alpha| + 1) 
\leq 
\frac{n \sqrt{b_0} \epsilon^2}{6G (\sum_{i\in [n]} L_i)(CG \sqrt{b_0}+ B \sigma)}, 
\text{ }
\rho^2
\leq 
\frac{n^2 b_0^2 \epsilon^4}{576 G^2(n^2 + b_0^2)(\sum_{i \in [n]} L_i)^2}, 
\end{align*}
then $U_i \leq \frac{\epsilon^2}{6}$ ($i=3,4,5,6$), i.e., 
\begin{align}\label{b_t_n_1}
&\min_{t \in [0:T-1]}\mathbb{E}\left[ \left\|\nabla \hat{f}_{S,\rho}^{\mathrm{SAM}}(\bm{x}_t) \right\|_2 \right]
\leq \epsilon.
\end{align}
}
This completes the proof.

\subsection{Proof of Theorem \ref{thm:3_1}}\label{proof_thm_3_1}
Let $\eta_t = \eta > 0$. Then, we have 
\begin{align*}
\frac{T}{\sum_{t=0}^{T-1} \eta_t} = \frac{1}{\eta} =: H_1
\text{ and } 
\frac{\sum_{t=0}^{T-1} \eta_t^2}{\sum_{t=0}^{T-1} \eta_t}
= \eta  =: H_2,
\end{align*}
which implies that (\ref{eta_thm_3_1}) with $H_3 = 0$ holds. Hence, from (\ref{eta_thm_3_2}), the assertion in Theorem \ref{thm:3_1} holds.

\subsection{Proof of Theorem \ref{thm:3_2}}\label{proof_thm_3_2}
We can prove the following corollary by using Theorem \ref{thm:3}.

\begin{corollary}
\label{cor:1}
\textup{\textbf{($\epsilon$--approximation of GSAM with a constant batch size and decaying learning rate)}}
Consider the sequence $(\bm{x}_t)$ generated by the mini-batch GSAM algorithm (Algorithm \ref{algo:1}) with a constant batch size $b \in (0,n]$ and a decaying learning rate $\eta_t \in [\underline{\eta}, \overline{\eta}] \subset [0,+\infty)$ satisfying that there exist positive numbers $H_1 (\underline{\eta}, \overline{\eta})$, $H_2 (\underline{\eta}, \overline{\eta})$, and $H_3 (\underline{\eta}, \overline{\eta})$ such that, for all $T \geq 1$, (\ref{eta_thm_3_1}) holds. We will assume that there exists a positive number $G$ such that $\max \{ \sup_{t\in \mathbb{N} \cup \{0\}} \|\nabla f_S (\bm{x}_{t} + \hat{\bm{\epsilon}}_{S_{t},\rho}(\bm{x}_{t}))\|_2, \sup_{t\in \mathbb{N} \cup \{0\}} \|\nabla \hat{f}^{\mathrm{SAM}}_{S_t,\rho} (\bm{x}_t)\|_2, \\ \sup_{t\in \mathbb{N} \cup \{0\}} \|\nabla \hat{f}^{\mathrm{SAM}}_{S,\rho} (\bm{x}_t)\|_2, G_{\perp} \} \leq G$, where $G_{\perp} := \sup_{t\in \mathbb{N}\cup \{0\}}\|\nabla f_{S_t \perp}(\bm{x}_t)\|_2 < + \infty$ (Theorem \ref{thm:1}). Let $\epsilon > 0$ be the precision and let $b_0 > 0$, $\alpha \in \mathbb{R}$, and $\rho \geq 0$ such that 
{
\begin{align}
&H_1 
\leq \frac{\epsilon^2}{12 \sigma \rho C} \left(  \frac{ G}{\sqrt{b}} + \frac{3 \sigma \rho C}{n b} \sum_{i \in [n]} L_i \right)^{-1}, 
\text{ } 
(|\alpha| + 1)^2 H_2 \leq 
\frac{n^3 \epsilon^2}{6 G^2 \sum_{i\in [n]} L_i \{n^2 + 4\rho^2 C^2(\sum_{i\in [n]} L_i)^2 \}},\nonumber \\
&\rho (|\alpha| + 1) 
\leq 
\frac{n \sqrt{b} \epsilon^2}{6G (\sum_{i\in [n]} L_i)(CG \sqrt{b}+ B \sigma)}, 
\text{ }
\rho^2
\leq 
\frac{n^2 b^2 \epsilon^4}{576 G^2(n^2 + b^2)(\sum_{i\in [n]} L_i)^2}, \label{h_1_2}
\end{align}
}
where $B$ and $C$ are nonnegative constants. Then, there exists $t_0 \in \mathbb{N}$ such that, for all $T \geq t_0$, 
\begin{align*}
&\min_{t \in [0:T-1]}\mathbb{E}\left[ \left\|\nabla \hat{f}_{S,\rho}^{\mathrm{SAM}}(\bm{x}_t) \right\|_2 \right]
\leq \epsilon.
\end{align*}
\end{corollary}

{\em Proof:} Let $b_t = b$ ($t \in \mathbb{N} \cup \{0\}$). Using inequality (\ref{key_1}) that was used to prove Theorem \ref{thm:3}, we have that, for all $t \in \mathbb{N} \cup \{0\}$,
{
\begin{align*}
\eta_t \mathbb{E}\left[ \left\|\nabla \hat{f}_{S,\rho}^{\mathrm{SAM}}(\bm{x}_t) \right\|_2^2 \right]
&\leq 
\left( \mathbb{E}[F_{\rho}(t)] - \mathbb{E}[F_{\rho}(t+1)]  \right) 
+ 2 \sigma \rho C 
\left(  \frac{ G}{\sqrt{b}} + \frac{3 \sigma \rho C}{n b} \sum_{i \in [n]} L_i \right)\\
&\quad + \frac{\eta_t^2 (|\alpha| + 1)^2 G^2}{n} 
 \sum_{i\in[n]} L_i
\left\{ 1 + \frac{4\rho^2 C^2}{n^2}\bigg(\sum_{i\in [n]} L_i \bigg)^2 \right\}\\
&\quad 
+ \eta_t G 
\sqrt{4 \rho^2 \left( \frac{1}{b^2} + \frac{1}{n^2} \right)
\big(\sum_{i \in [n]} L_i \big)^2 
+ 2 \sigma_t^2}\\
&\quad 
+ \eta_t \frac{\rho (|\alpha| + 1) G}{n} \sum_{i\in [n]} L_i
\left( CG + \frac{B \sigma}{\sqrt{b}}   \right),
\end{align*}
}
which, together with a discussion similar to the one showing (\ref{key_2}), implies that, for all $T \geq 1$, 
{
\begin{align*}
\min_{t \in [0:T-1]}\mathbb{E}\left[ \left\|\nabla \hat{f}_{S,\rho}^{\mathrm{SAM}}(\bm{x}_t) \right\|_2^2 \right]
&\leq
\frac{\mathbb{E}[F_{\rho}(0)] - f_S^\star}{\sum_{t=0}^{T-1} \eta_t}
+ 2 \sigma \rho C 
\left(  \frac{ G}{\sqrt{b}} + \frac{3 \sigma\rho C}{n b} \sum_{i \in [n]} L_i \right) \frac{T}{\sum_{t=0}^{T-1} \eta_t}\\
&\quad + \frac{(|\alpha| + 1)^2 G^2}{n} 
\sum_{i\in[n]} L_i
\left\{ 1 + \frac{4\rho^2 C^2}{n^2}\bigg(\sum_{i\in [n]} L_i \bigg)^2 \right\} \frac{\sum_{t=0}^{T-1} \eta_t^2}{\sum_{t=0}^{T-1} \eta_t}\\
&\quad 
+ G 
\sqrt{4 \rho^2 \left( \frac{1}{b^2} + \frac{1}{n^2} \right)
\big(\sum_{i \in [n]} L_i \big)^2 
+ 2 \sigma_t^2}\\
&\quad 
+ \frac{\rho (|\alpha| + 1) G}{n} \sum_{i\in [n]} L_i
\left( CG + \frac{B \sigma}{\sqrt{b}}   \right).
\end{align*} 
}
Let $\epsilon > 0$. From (\ref{eta_thm_3_1}), 
{
\begin{align*}
&\min_{t \in [0:T-1]}\mathbb{E}\left[ \left\|\nabla \hat{f}_{S,\rho}^{\mathrm{SAM}}(\bm{x}_t) \right\|_2^2 \right]\\
&\leq
\underbrace{\frac{H_1 (\mathbb{E}[F_{\rho}(0)] - f_S^\star)}{T}}_{V_1 \leq \frac{\epsilon^2}{6}} 
+ \underbrace{\frac{(|\alpha| + 1)^2 G^2}{n} 
\sum_{i\in[n]} L_i
\left\{ 1 + \frac{4\rho^2 C^2}{n^2}\bigg(\sum_{i\in [n]} L_i \bigg)^2 \right\} \frac{H_3}{T}}_{V_2 \leq \frac{\epsilon^2}{6}}\\
&\quad 
+ \underbrace{2 \sigma \rho C H_1
\left(  \frac{G}{\sqrt{b}} + \frac{3 \sigma\rho C}{n b} \sum_{i \in [n]} L_i \right)}_{V_3 \leq \frac{\epsilon^2}{6}} 
+ \underbrace{\frac{(|\alpha| + 1)^2 G^2 H_2}{n} 
\sum_{i\in[n]} L_i
\left\{ 1 + \frac{4\rho^2 C^2}{n^2}\bigg(\sum_{i\in [n]} L_i \bigg)^2 \right\}}_{V_4 \leq \frac{\epsilon^2}{6}}\\
&\quad
+ \underbrace{\frac{\rho (|\alpha| + 1) G}{n} \sum_{i\in [n]} L_i
\left( CG + \frac{B \sigma}{\sqrt{b}}   \right)}_{V_5 \leq \frac{\epsilon^2}{6}} 
+ \underbrace{G \sqrt{4 \rho^2 \left( \frac{1}{b^2} + \frac{1}{n^2} \right)
\big(\sum_{i \in [n]} L_i \big)^2 
+ 2 \sigma_t^2}}_{V_6 \leq \frac{\epsilon^2}{6}}.
\end{align*} 
}
There exists $t_2 \in \mathbb{N}$ such that, for all $T \geq t_2$, $V_1 \leq \frac{\epsilon^2}{6}$ and $V_2 \leq \frac{\epsilon^2}{6}$. Moreover, if 
{
\begin{align*}
&H_1 
\leq \frac{\epsilon^2}{12 \sigma \rho C} \left(  \frac{G}{\sqrt{b}} + \frac{3 \sigma\rho C}{n b} \sum_{i \in [n]} L_i \right)^{-1}, 
\text{ } 
(|\alpha| + 1)^2 H_2 \leq 
\frac{n^3 \epsilon^2}{6 G^2 \sum_{i\in [n]} L_i \{n^2 + 4\rho^2 C^2(\sum_{i\in [n]} L_i)^2 \}},\\
&\rho (|\alpha| + 1) 
\leq 
\frac{n \sqrt{b} \epsilon^2}{6G (\sum_{i\in [n]} L_i)(CG \sqrt{b}+ B \sigma)}, 
\text{ }
\rho^2
\leq 
\frac{\epsilon^4}{576 G^2}\frac{n^2 b^2}{(n^2 + b^2)(\sum_{i\in [n]} L_i)^2},
\end{align*}
}
then $V_i \leq \frac{\epsilon^2}{6}$ ($i=3,4,5,6$); i.e., (\ref{b_t_n_1}) holds.

{\em Proof of Theorem \ref{thm:3_2}:} Let $\eta_t$ be the cosine-annealing learning rate defined by (\ref{cosine_lr}). We then have 
\begin{align*}
\sum_{t = 0}^{KE-1} \eta_t
&=
\underline{\eta} KE 
+ 
\frac{\overline{\eta} - \underline{\eta}}{2}
KE 
+
\frac{\overline{\eta} - \underline{\eta}}{2}
\sum_{t = 0}^{KE -1} \cos \left\lfloor \frac{t}{K} \right\rfloor \frac{\pi}{E}. 
\end{align*}
We have 
\begin{align}\label{cos}
\sum_{t = 0}^{KE -1} \cos \left\lfloor \frac{t}{K} \right\rfloor \frac{\pi}{E}
=
\sum_{t = 0}^{KE} \cos \left\lfloor \frac{t}{K} \right\rfloor \frac{\pi}{E}
- \cos \pi 
= 
(K-1) + 1
=
K. 
\end{align}
We thus have 
\begin{align*}
\sum_{t = 0}^{KE-1} \eta_t
&=
\underline{\eta} KE 
+ 
\frac{\overline{\eta} - \underline{\eta}}{2}
KE 
+
\frac{\overline{\eta} - \underline{\eta}}{2}
K\\
&= 
\frac{1}{2}
\{ (\underline{\eta} + \overline{\eta}) KE 
+ (\overline{\eta} - \underline{\eta}) K \}\\
&\geq 
\frac{(\underline{\eta} + \overline{\eta})KE}{2}.
\end{align*}
Moreover, we have 
\begin{align*}
\sum_{t = 0}^{KE-1} \eta_t^2
&=
\underline{\eta}^2 KE 
+ \underline{\eta} (\overline{\eta} - \underline{\eta})
\sum_{t = 0}^{KE -1} \left(1 + \cos \left\lfloor \frac{t}{K} \right\rfloor \frac{\pi}{E}   \right)\\
&\quad
+ \frac{(\overline{\eta} - \underline{\eta})^2}{4}
\sum_{t = 0}^{KE -1} \left(1 + \cos \left\lfloor \frac{t}{K} \right\rfloor \frac{\pi}{E}   \right)^2,
\end{align*}
which implies that
\begin{align*}
\sum_{t = 0}^{KE-1} \eta_t^2
&= 
\underline{\eta} \overline{\eta} KE 
+ \frac{(\overline{\eta} - \underline{\eta})^2}{4} KE 
+ \underline{\eta} (\overline{\eta} - \underline{\eta}) 
\sum_{t = 0}^{KE -1} \cos \left\lfloor \frac{t}{K} \right\rfloor \frac{\pi}{E}\\
&\quad 
+ \frac{(\overline{\eta} - \underline{\eta})^2}{2} 
\sum_{t = 0}^{KE -1} \cos \left\lfloor \frac{t}{K} \right\rfloor \frac{\pi}{E}
+ \frac{(\overline{\eta} - \underline{\eta})^2}{4}
\sum_{t = 0}^{KE -1} \cos^2 \left\lfloor \frac{t}{K} \right\rfloor \frac{\pi}{E}.
\end{align*}
From 
\begin{align*}
\sum_{t = 0}^{KE} \cos^2 \left\lfloor \frac{t}{K} \right\rfloor \frac{\pi}{E}
&= 
\frac{1}{2} \sum_{t = 0}^{KE}
\left(
1 + 
\cos 2 \left\lfloor \frac{t}{K} \right\rfloor \frac{\pi}{E}
\right)\\
&=
\frac{1}{2} ( KE +  1 ) + \frac{1}{2}\\
&= \frac{KE}{2} + 1,
\end{align*}
we have
\begin{align*}
\sum_{t = 0}^{KE -1} \cos^2 \left\lfloor \frac{t}{K} \right\rfloor \frac{\pi}{E}
= 
\frac{KE}{2} + 1 - \cos^2 \pi 
=
\frac{KE}{2}.
\end{align*}
Moreover, from (\ref{cos}),
\begin{align*}
\sum_{t = 0}^{KE-1} \eta_t^2
&= 
\frac{(\underline{\eta} + \overline{\eta})^2}{4} KE 
+ \underline{\eta} (\overline{\eta} - \underline{\eta})
+ \frac{(\overline{\eta} - \underline{\eta})^2}{2}
+ \frac{(\overline{\eta} - \underline{\eta})^2}{4}\frac{KE}{2}\\
&=
\frac{3 \underline{\eta}^2 + 2 \underline{\eta} \overline{\eta} + 3 \overline{\eta}^2}{8} KE
+ \frac{(\overline{\eta} - \underline{\eta})(\overline{\eta} + \underline{\eta})}{2}.
\end{align*}
Hence, we have 
\begin{align*}
\frac{KE}{\sum_{t = 0}^{KE-1} \eta_t}
\leq 
\frac{2 KE}{(\underline{\eta} + \overline{\eta})KE}
< 
\frac{2}{\underline{\eta} + \overline{\eta}} =: H_1
\end{align*}
and
\begin{align*}
\frac{\sum_{t = 0}^{KE-1} \eta_t^2}{\sum_{t = 0}^{KE-1} \eta_t}
&\leq
\underbrace{\frac{(3 \underline{\eta}^2 + 2 \underline{\eta} \overline{\eta} + 3 \overline{\eta}^2)}{4(\underline{\eta} + \overline{\eta})}}_{H_2} 
+ \frac{1}{KE} \underbrace{(\overline{\eta} - \underline{\eta})}_{H_3}.
\end{align*}
Accordingly, (\ref{eta_thm_3_1}) holds. From (\ref{h_1_2}), we have
{
\begin{align*}
&\frac{2}{\underline{\eta} + \overline{\eta}}
\leq
\frac{\epsilon^2}{12 \sigma \rho C} \left( \frac{ G}{\sqrt{b}} + \frac{3 \sigma\rho C}{n b} \sum_{i\in [n]} L_i  \right)^{-1},\\ 
&(|\alpha| + 1)^2 \frac{(3 \underline{\eta}^2 + 2 \underline{\eta} \overline{\eta} + 3 \overline{\eta}^2)}{4(\underline{\eta} + \overline{\eta})} \leq 
\frac{n^3 \epsilon^2}{6 G^2 \sum_{i\in [n]} L_i \{n^2 + 4\rho^2 C^2(\sum_{i\in [n]} L_i)^2 \}}.
\end{align*}
}
In particular, when $\underline{\eta} = 0$, 
{
\begin{align*}
&\frac{2}{\overline{\eta}}
\leq
\frac{\epsilon^2}{12 \sigma \rho C} \left( \frac{G}{\sqrt{b}} + \frac{3 \sigma\rho C}{n b} \sum_{i\in [n]} L_i  \right)^{-1},\\ 
&(|\alpha| + 1)^2 \frac{3 \overline{\eta}}{4} \leq 
\frac{n^3 \epsilon^2}{6 G^2 \sum_{i\in [n]} L_i \{n^2 + 4\rho^2 C^2(\sum_{i\in [n]} L_i)^2 \}}.
\end{align*}
}
Therefore, Corollary \ref{cor:1} leads to the assertion in Theorem \ref{thm:3_2}.

Let $\eta_t$ be the linear learning rate defined by (\ref{linear_lr}). We then have 
\begin{align*}
\sum_{t =0}^{T-1} \eta_t
=
\overline{\eta} T 
+ 
\frac{\underline{\eta} - \overline{\eta}}{T} \frac{(T - 1)T}{2}
=
\frac{1}{2} \{ (\underline{\eta} + \overline{\eta}) T + \overline{\eta} - \underline{\eta} \}
> 
\frac{\underline{\eta} + \overline{\eta}}{2} T, 
\end{align*} 
where the third inequality comes from $\overline{\eta} > \underline{\eta}$. We also have 
\begin{align*}
\sum_{t = 0}^{T-1} \eta_t^2
&=
\left(\frac{\underline{\eta} - \overline{\eta}}{T} \right)^2
\frac{(T-1)T(2T -1)}{6}
+ \frac{2(\underline{\eta} - \overline{\eta})\overline{\eta}}{T}
\frac{(T-1)T}{2}
+ \overline{\eta}^2 T\\
&=
\frac{(\underline{\eta} - \overline{\eta})^2 (T-1)(2T-1)}{6 T}
+ (\underline{\eta} - \overline{\eta})\overline{\eta} (T-1)
+ \overline{\eta}^2 T\\
&< 
\frac{(\underline{\eta} - \overline{\eta})^2 T}{3}
+ (\underline{\eta} - \overline{\eta})\overline{\eta} T
+ \overline{\eta}^2 T\\
&=
\frac{(\underline{\eta} - \overline{\eta})^2 T}{3}
+ \underline{\eta} \overline{\eta} T\\
&= 
\frac{\underline{\eta}^2 + \underline{\eta} \overline{\eta} + \overline{\eta}^2}{3}
T,
\end{align*}
where the third inequality comes from $T-1 < T$ and $2 T -1 < 2 T$. Hence,
\begin{align*} 
\frac{T}{\sum_{t = 0}^{T-1} \eta_t}
< 
\frac{2}{\underline{\eta} + \overline{\eta}} =: H_1
\end{align*}
and
\begin{align*}
\frac{\sum_{t = 0}^{T-1} \eta_t^2}{\sum_{t = 0}^{T-1} \eta_t}
<
\frac{2(\underline{\eta}^2 + \underline{\eta} \overline{\eta} + \overline{\eta}^2)}{3(\underline{\eta} + \overline{\eta})}
=: H_2.
\end{align*}
Accordingly, (\ref{eta_thm_3_1}) holds. From (\ref{h_1_2}), we have that
{
\begin{align*}
&\frac{2}{\underline{\eta} + \overline{\eta}}
\leq
\frac{\epsilon^2}{12 \sigma \rho C} \left( \frac{G}{\sqrt{b}} + \frac{3 \sigma \rho C}{n b} \sum_{i\in [n]} L_i  \right)^{-1},\\ 
&(|\alpha| + 1)^2 \frac{2(\underline{\eta}^2 + \underline{\eta} \overline{\eta} + \overline{\eta}^2)}{3(\underline{\eta} + \overline{\eta})} \leq 
\frac{n^3 \epsilon^2}{6 G^2 \sum_{i\in [n]} L_i \{n^2 + 4\rho^2 C^2(\sum_{i\in [n]} L_i)^2 \}}.
\end{align*}
}
In particular, when $\underline{\eta} = 0$,
{
\begin{align*}
&\frac{2}{\overline{\eta}}
\leq
\frac{\epsilon^2}{12 \sigma \rho C} \left( \frac{G}{\sqrt{b}} + \frac{3 \sigma \rho C}{n b} \sum_{i\in [n]} L_i  \right)^{-1},\\ 
&(|\alpha| + 1)^2 \frac{2 \overline{\eta}}{3} \leq 
\frac{n^3 \epsilon^2}{6 G^2 \sum_{i\in [n]} L_i \{n^2 + 4\rho^2 C^2(\sum_{i\in [n]} L_i)^2 \}}.
\end{align*}
}
Therefore, Corollary \ref{cor:1} leads to the assertion in Theorem \ref{thm:3_2}.

{
\section{Convergence Rates of Theorems \ref{thm:3_1} and \ref{thm:3_2}}\label{sec:rate}
}
{
\subsection{Derivation of the rate}\label{subsec:rate}
From the proof of Theorems \ref{thm:3} (just before (\ref{key_2}))
\[
\min_{t \in [0:T-1]}\mathbf{E}\left[ \left\|\nabla \hat{f}_{S,\rho}^{\mathrm{SAM}}(\bm{x}_t) \right\|_2^2 \right] \le U_1 + U_2 + \cdots + U_6 \le \epsilon^2 .
\]
By definition, $U_1 = M/T$ with $M = O(H_1)$. Taking $H_2$ at the upper limit permitted by \ref{eta_2}, 
a constant learning rate gives $H_1 = O(\epsilon^{-2})$, so $U_1 \le \epsilon^2/6$ holds for $T = O(\epsilon^{-4})$, 
i.e., $\epsilon = O(T^{-1/4})$. By Jensen's inequality and Theorem~3, we obtain
\[
\min_{t\in[0:T-1]} \mathbf{E}\left[ \left\|\nabla \hat{f}^{\mathrm{SAM}}_{S,\rho}(\bm{x}_t) \right\|_2 \right] = O\left(\frac{1}{T^{1/4}}\right).
\]
The same argument yields the rate $O(T^{-1/4})$ for Theorem \ref{thm:3_2}.
}

{
\subsection{Conditions on $\epsilon$, $\rho$, and $b$}\label{subsec:conditions}
\paragraph{Joint condition on $\epsilon$ and $\rho$.}
Both endpoints of the interval in (\ref{eta_2}) are positive, and the interval is non-empty 
if and only if
\begin{align}
72(|\alpha|+1)^{2} G^{2}\sigma\rho C \left( \sum_{i\in[n]} L_i \right)
\left( \frac{G}{\sqrt{b_0}} + \frac{3\sigma\rho C}{n b_0} \sum_{i\in[n]} L_i \right)
\left( n^{2} + 4\rho^{2}C^{2} \Big( \sum_{i\in[n]} L_i \Big)^{2} \right)
\le n^{3}\epsilon^{4}.
\label{eq:joint}
\end{align}
This is a single joint condition on $\epsilon$ and $\rho$, which cannot be chosen independently: 
fixing $\epsilon$ determines an upper bound on $\rho$, whereas fixing $\rho$ determines a lower bound 
on the attainable precision $\epsilon$. In particular, (\ref{eq:joint}) implies $\rho = O(\epsilon^4)$, 
and hence $\rho = O(T^{-1})$ for the rate $O(T^{-1/4})$ in rows (7) and (8) of Table \ref{table:1}. 
Therefore, Theorems \ref{thm:3_1} and \ref{thm:3_2} do not hold for an arbitrary precision $\epsilon > 0$ at a fixed $\rho$.
\paragraph{Increasing batch size (Theorem \ref{thm:3_1}).}
Since the mini-batch $S_t$ is sampled without replacement, $S_t = S$ and $\sigma_t^2 = 0$ whenever $b_t = n$. 
Since the schedule (\ref{mini-batch}) reaches $b_M = n$, there exists $t_0 \in \mathbb{N}$ such that 
$2\sigma_t^2 \le \epsilon^4/(48G^2)$ for all $t \ge t_0$, which is the condition required in the proof of Theorem \ref{thm:3}. 
Hence, no additional restriction on $\epsilon$ arises from the batch size, and Theorem~3 holds 
under (\ref{eq:joint}).
\paragraph{Constant batch size (Theorem \ref{thm:3_2}).}
For a constant batch size $b < n$, $\sigma_t^2$ does not vanish, and the following further condition is required:
\[
\epsilon^4 \ge \frac{96 G^2 \sigma^2}{b}, \quad \text{i.e.,} \quad b = \Omega(\epsilon^{-4}).
\]
This lower bound on $\epsilon$ decreases as $b$ approaches $n$, and the condition is unnecessary when $b = n$ 
since $\sigma_t^2 = 0$. Hence, Theorem \ref{thm:3_2}) holds when $\epsilon$ and $\rho$ satisfy (\ref{eq:joint})
with $b_0$ replaced by $b$, and $b = \Omega(\epsilon^{-4})$.
}

{\section{Training curves}}
\label{train curves}
{Figures \ref{figcifar10_b} and \ref{figcifar10_c} show the training curves of ResNet-18 trained from scratch on CIFAR-10. Figures \ref{fig3} and \ref{fig4} show the training curves of WideResNet-28-10 trained from scratch on CIFAR-100. Figures \ref{t-image_batch} and \ref{t-image_cosine} show the training curves of WideResNet-28-10 trained from scratch on Tiny-ImageNet.}

\begin{figure*}[ht]
\begin{tabular}{cc}
\begin{minipage}[t]{0.5\hsize}
\includegraphics[width=1.0\textwidth]{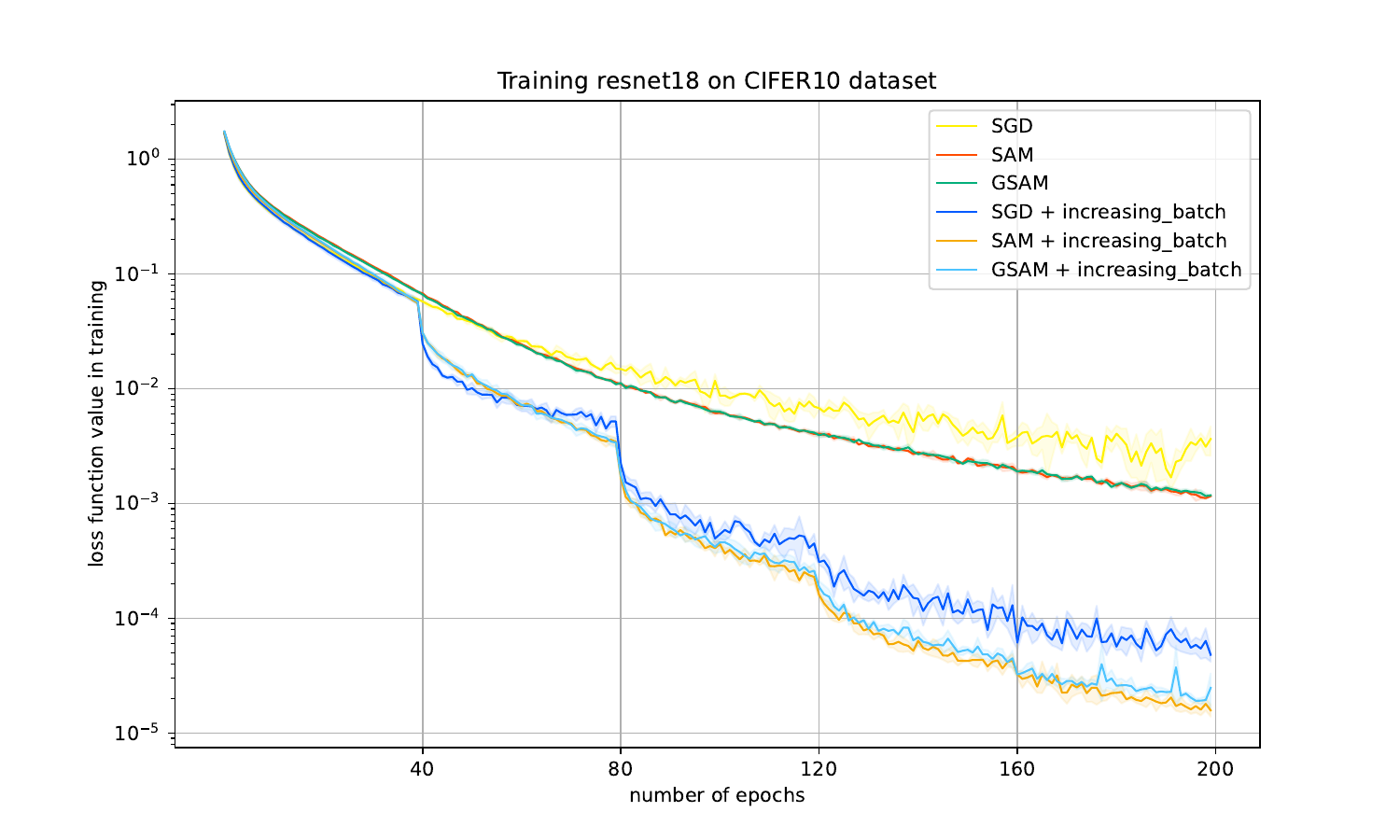}
\end{minipage} &
\begin{minipage}[t]{0.5\hsize}
\includegraphics[width=1.0\textwidth]{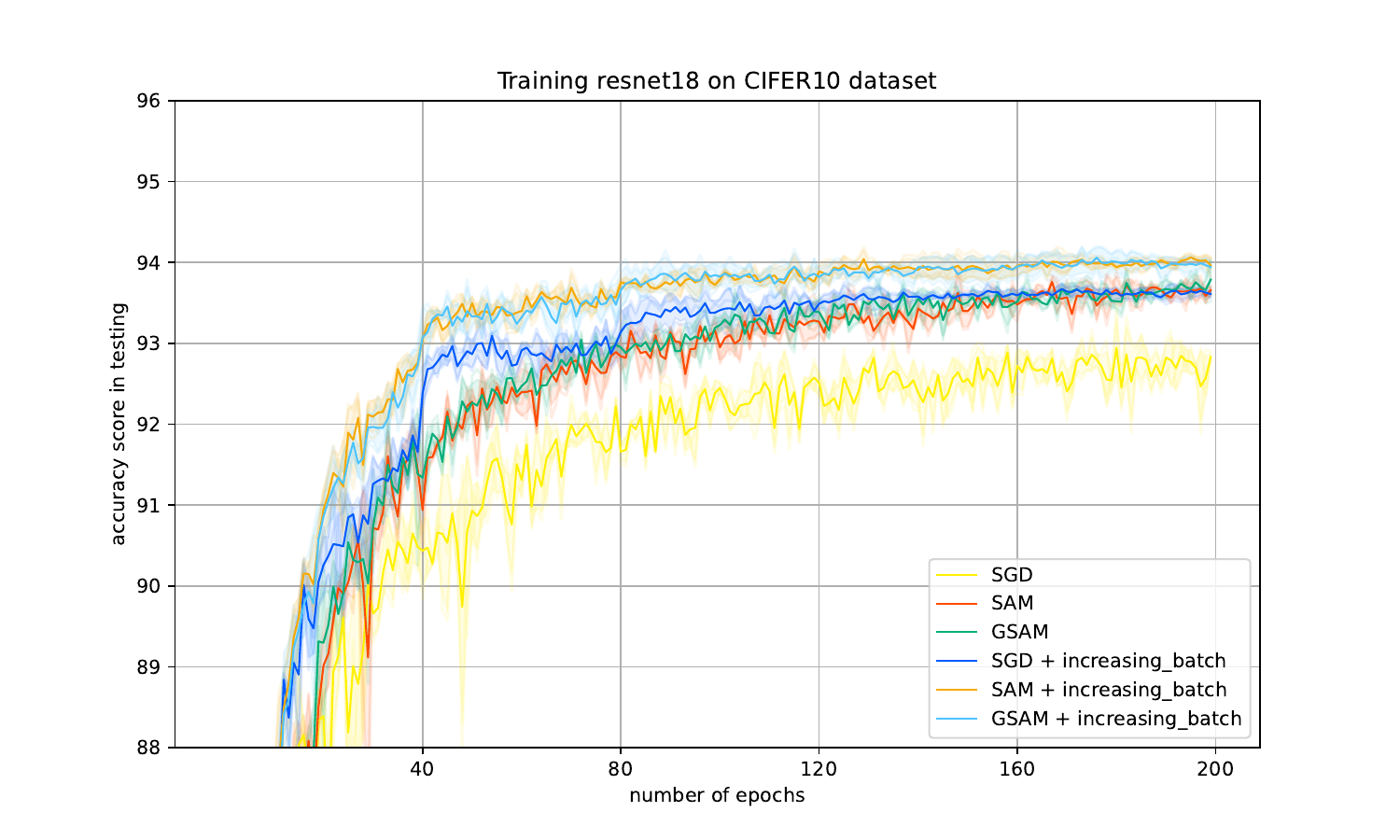}
\end{minipage}
\end{tabular}
\caption{(Left) Loss function value in training and (Right) accuracy score in testing for the algorithms versus the number of epochs in training ResNet-18 on the CIFAR10 dataset. The learning rate of each algorithm was fixed at 0.1. In SGD/SAM/GSAM, the batch size was fixed at 128. In SGD/SAM/GSAM + increasing\_batch, the batch size was set at 16 for the first 40 epochs and then it was doubled every 40 epochs afterwards, i.e., to 32 for epochs 41-80, 64 for epochs 81-120, etc.}
\label{figcifar10_b}
\end{figure*}

\begin{figure*}[ht]
\begin{tabular}{cc}
\begin{minipage}[t]{0.5\hsize}
\includegraphics[width=1.0\textwidth]{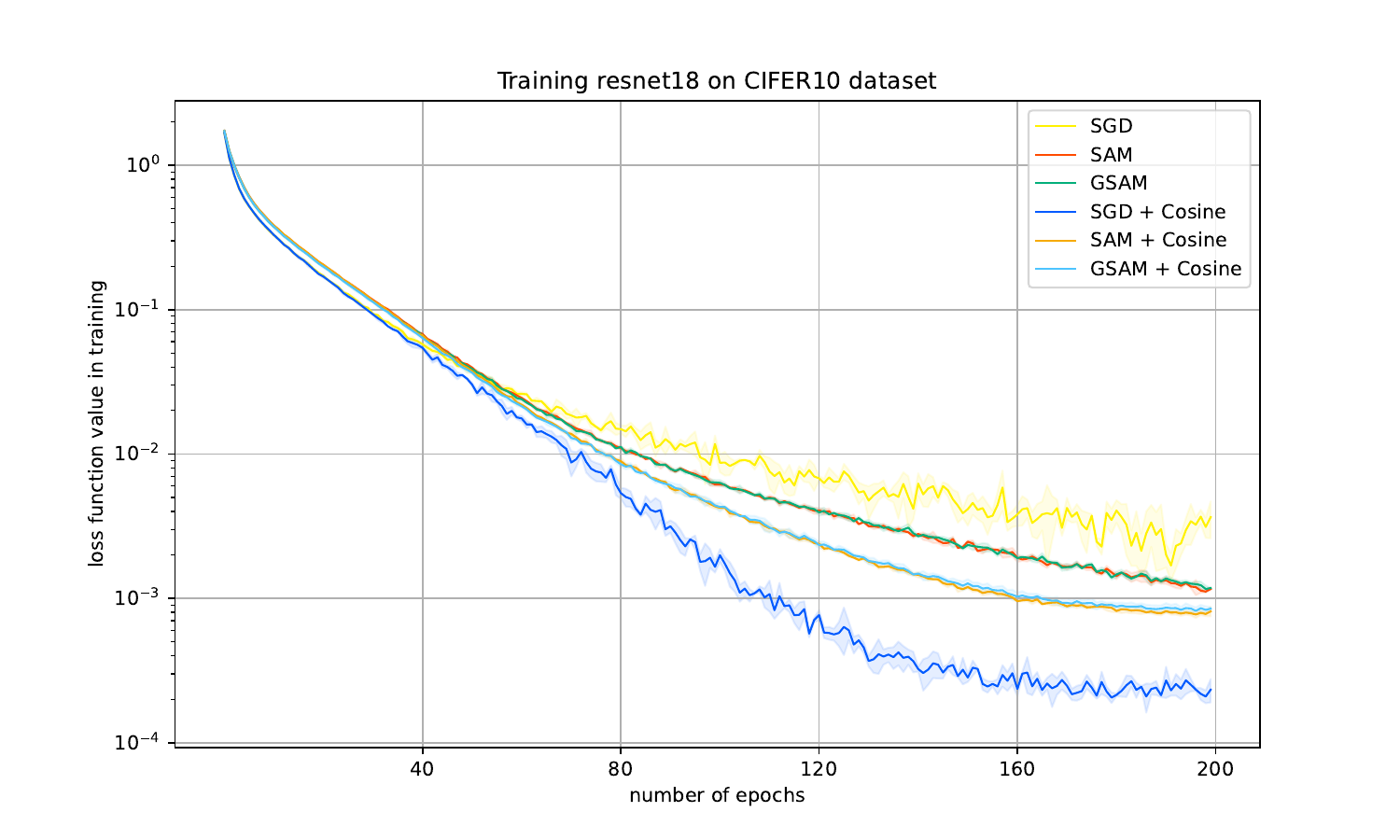}
\end{minipage} &
\begin{minipage}[t]{0.5\hsize}
\includegraphics[width=1.0\textwidth]{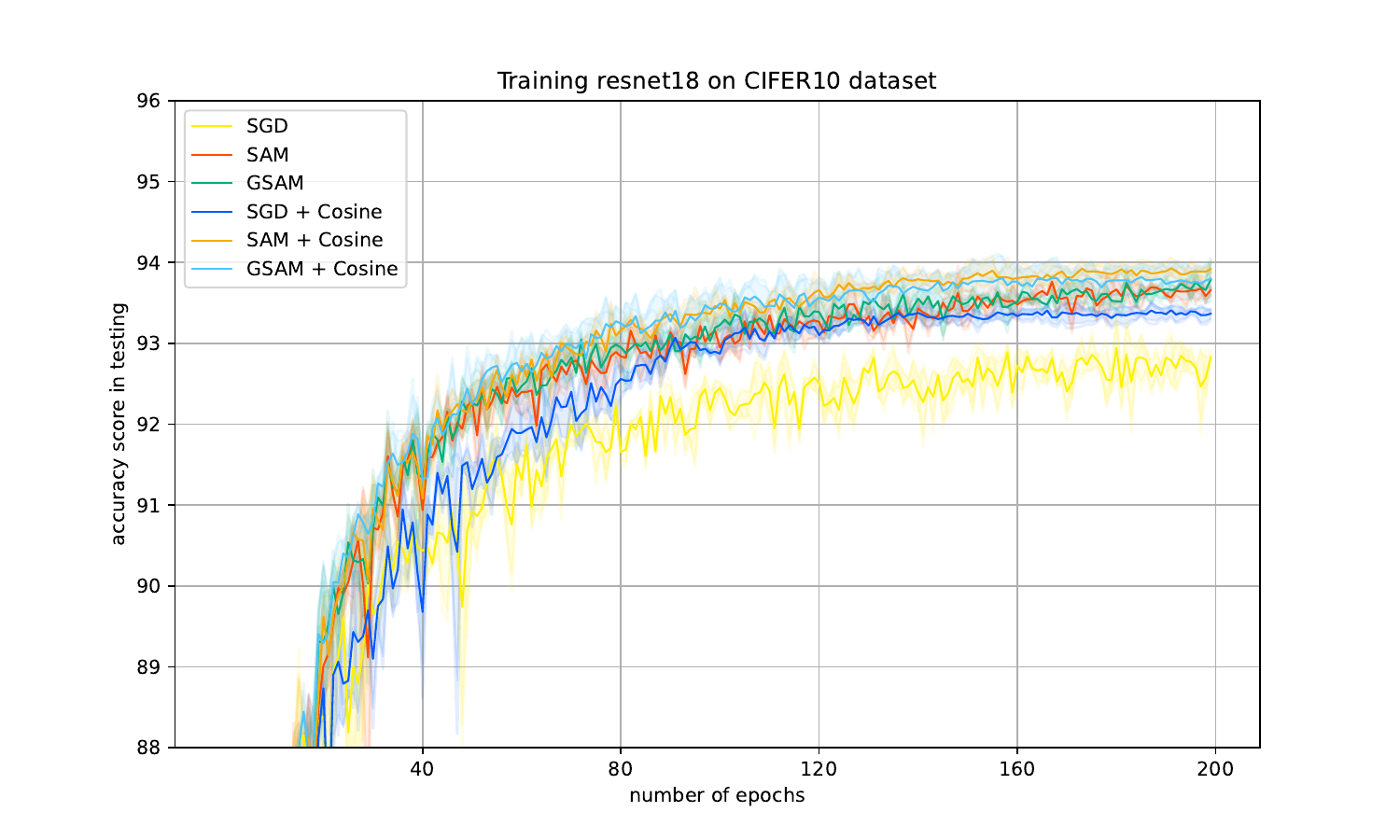}
\end{minipage}
\end{tabular}
\caption{(Left) Loss function value in training and (Right) accuracy score in testing for the algorithms versus the number of epochs in training ResNet-18 on the CIFAR-10 dataset. The batch size of each algorithm was fixed at 128. In SGD/SAM/GSAM, the constant learning rate was fixed at 0.1. In SGD/SAM/GSAM + Cosine, the maximum learning rate was 0.1 and the minimum learning rate was 0.001.}
\label{figcifar10_c}
\end{figure*}

\begin{figure*}[ht]
\begin{tabular}{cc}
\begin{minipage}[t]{0.5\hsize}
\includegraphics[width=1.0\textwidth]{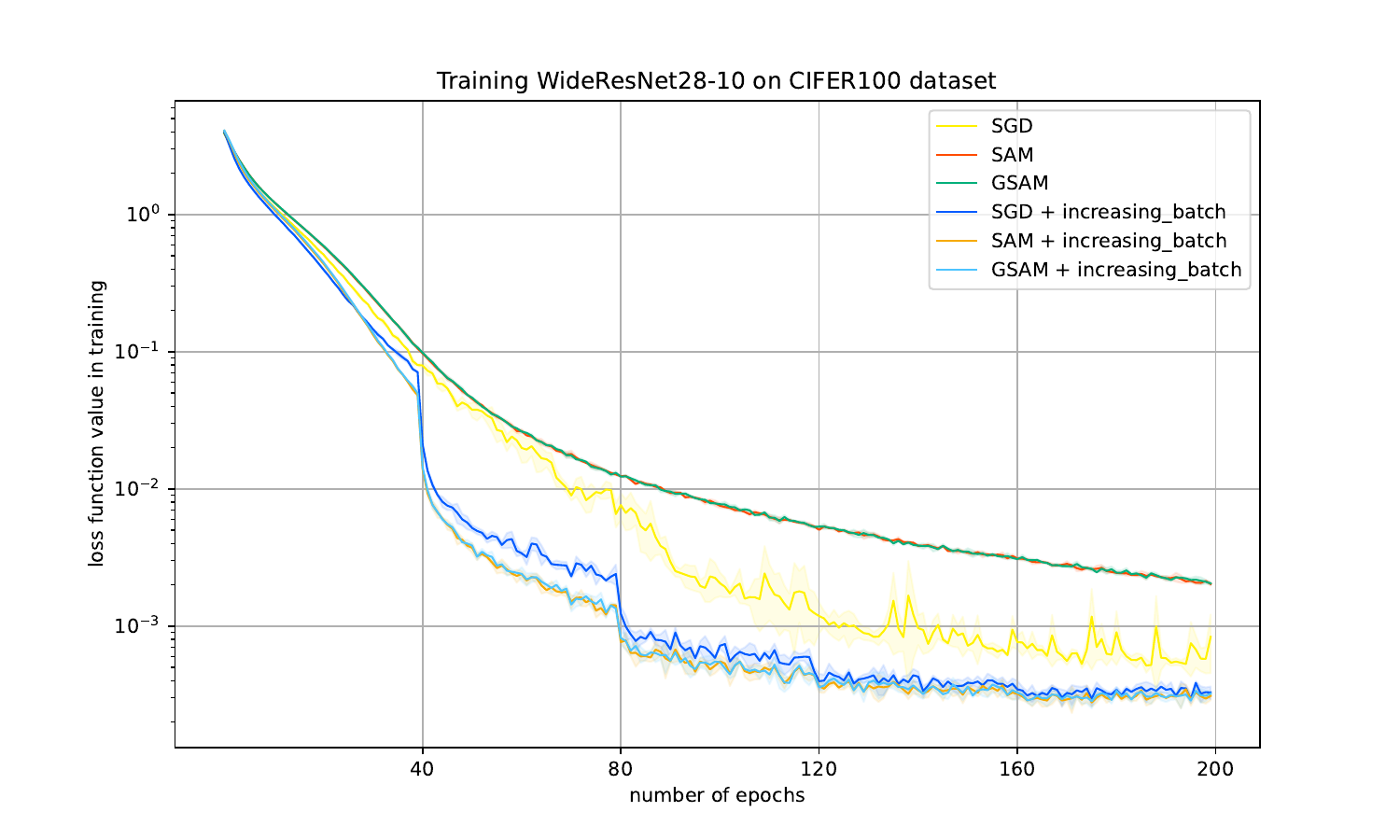}
\end{minipage} &
\begin{minipage}[t]{0.5\hsize}
\includegraphics[width=1.0\textwidth]{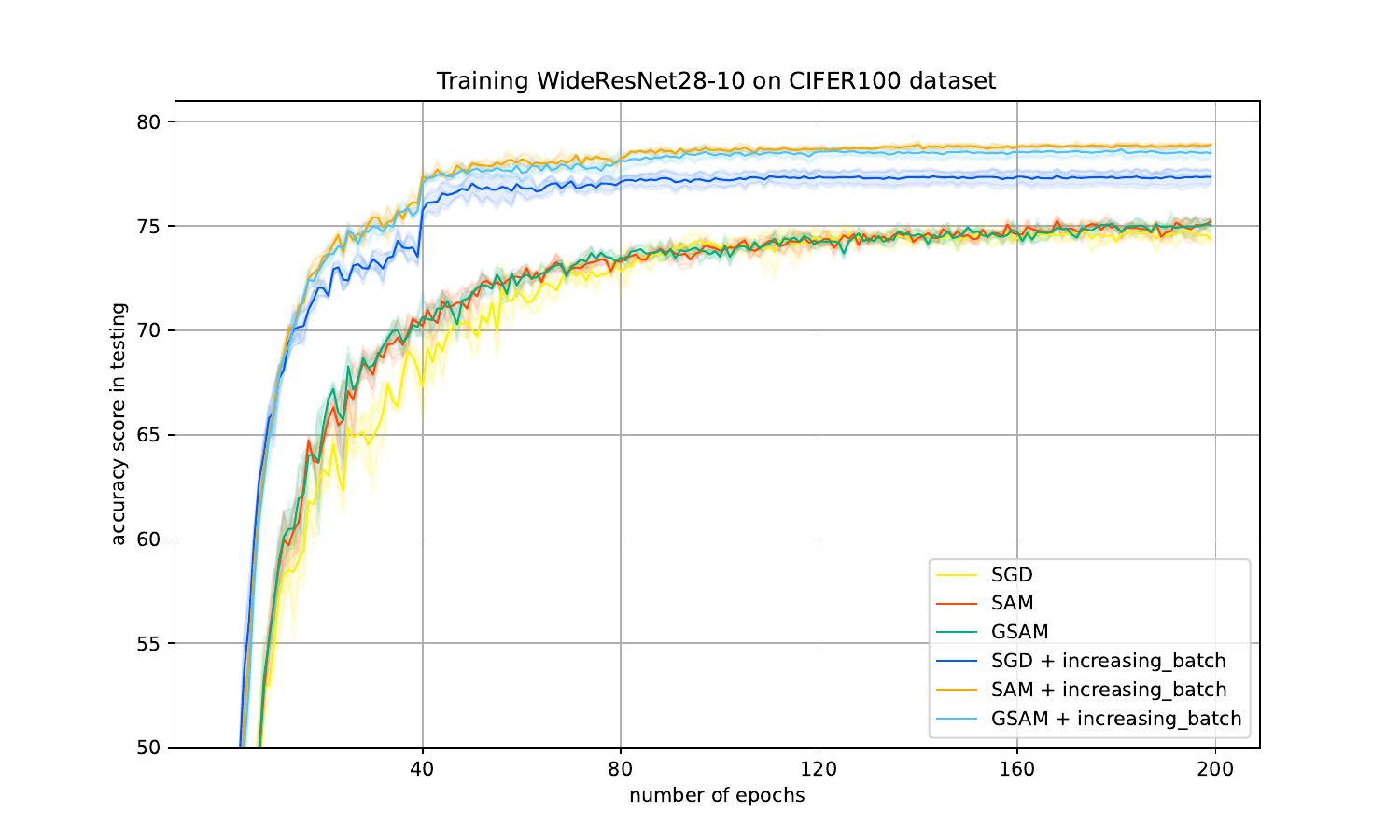}
\end{minipage}
\end{tabular}
\caption{(Left) Loss function value in training and (Right) accuracy score in testing for the algorithms versus the number of epochs in training Wide-ResNet-28-10 on the CIFAR-100 dataset. The learning rate of each algorithm was fixed at 0.1. In SGD/SAM/GSAM, the batch size was fixed at 128. In SGD/SAM/GSAM + increasing\_batch, the batch size was set at 16 for the first 40 epochs and then it was doubled every 40 epochs afterwards, i.e., to 32 for epochs 41-80, 64 for epochs 81-120, etc.}
\label{fig3}
\end{figure*}

\begin{figure*}[ht]
\begin{tabular}{cc}
\begin{minipage}[t]{0.5\hsize}
\includegraphics[width=1.0\textwidth]{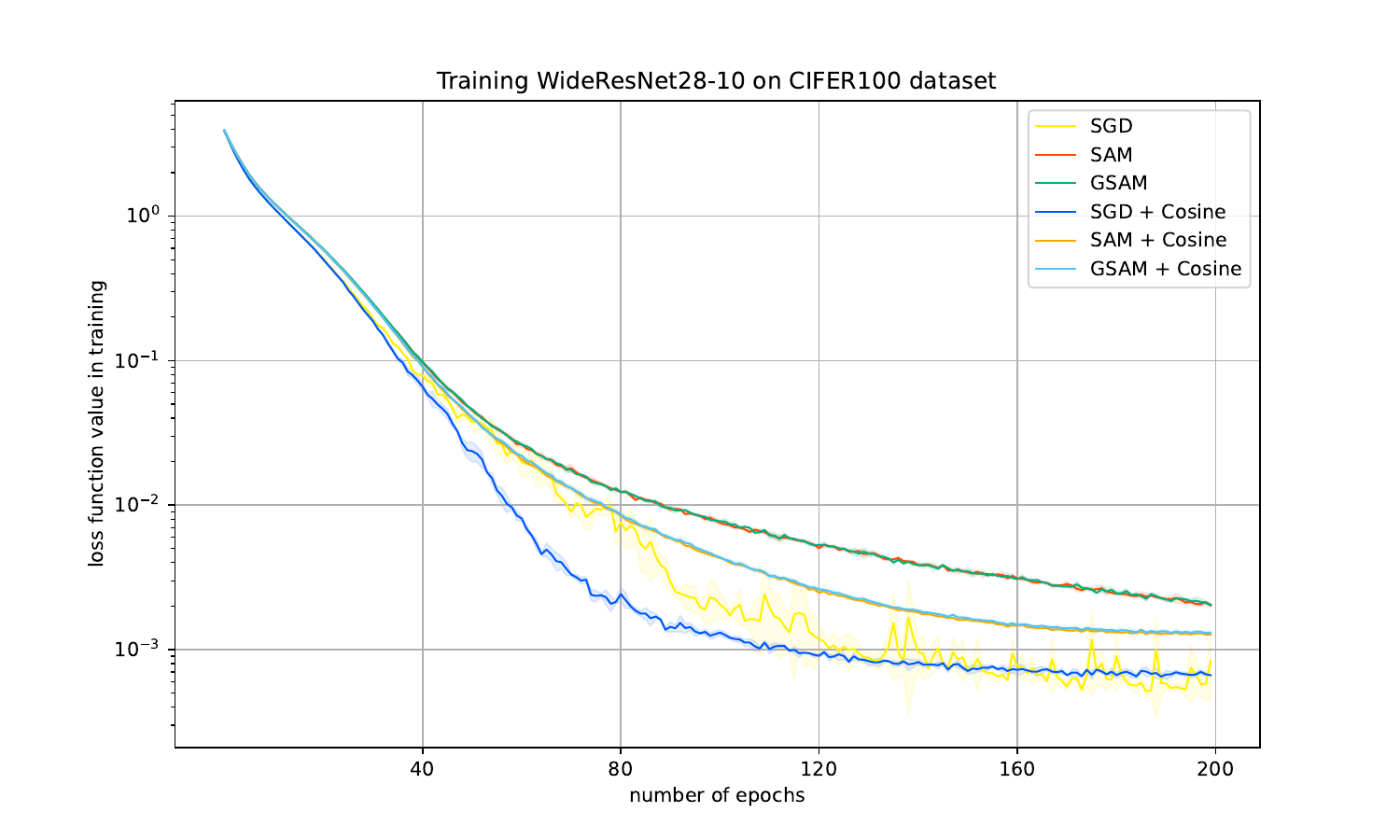}
\end{minipage} &
\begin{minipage}[t]{0.5\hsize}
\includegraphics[width=1.0\textwidth]{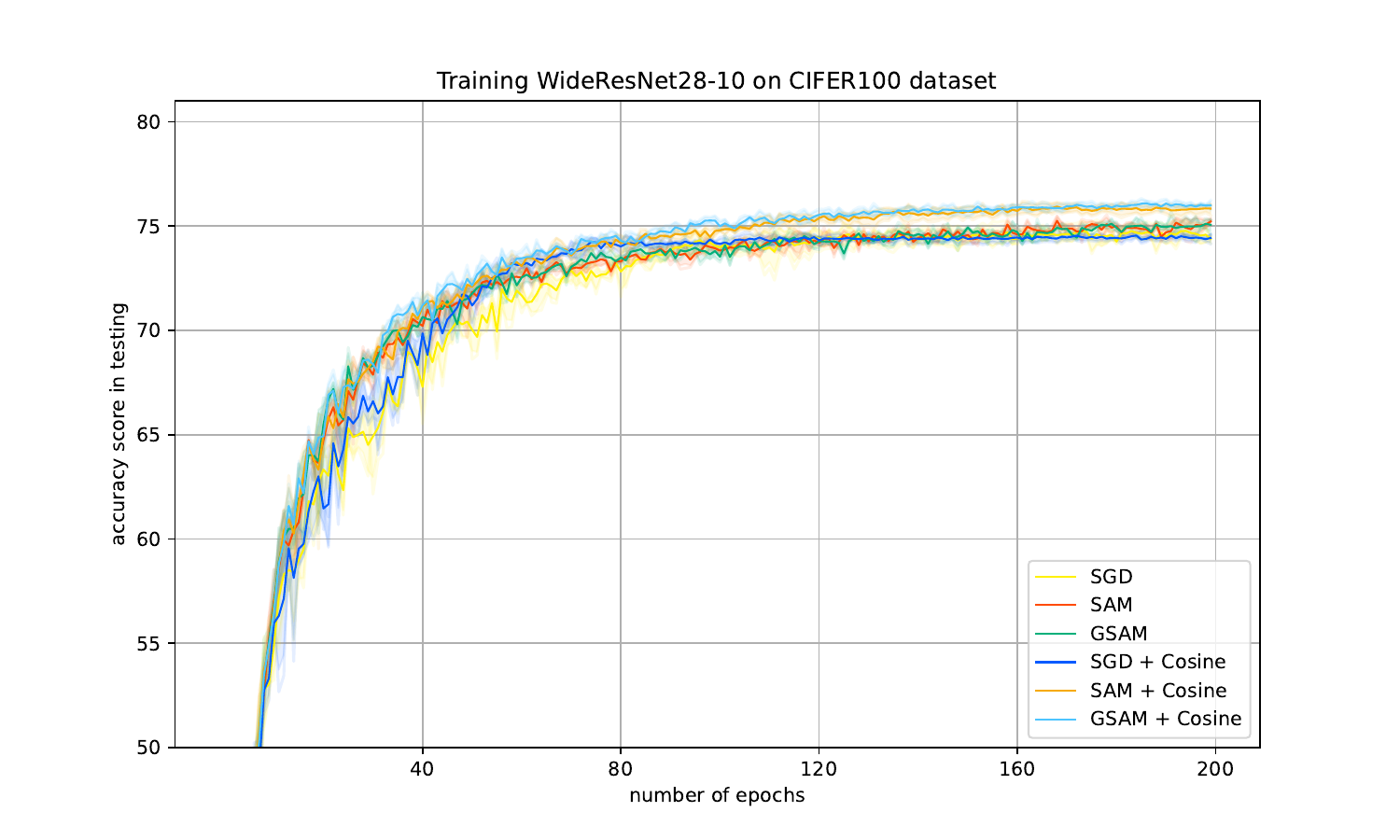}
\end{minipage}
\end{tabular}
\caption{(Left) Loss function value in training and (Right) accuracy score in testing for the algorithms versus the number of epochs in training Wide-ResNet-28-10 on the CIFAR-100 dataset. The batch size of each algorithm was fixed at 128. In SGD/SAM/GSAM, the constant learning rate was fixed at 0.1. In SGD/SAM/GSAM + Cosine, the maximum learning rate was 0.1 and the minimum learning rate was 0.001.}
\label{fig4}
\end{figure*}

\begin{figure*}[ht]
\begin{tabular}{cc}
\begin{minipage}[t]{0.5\hsize}
\includegraphics[width=1.0\textwidth]{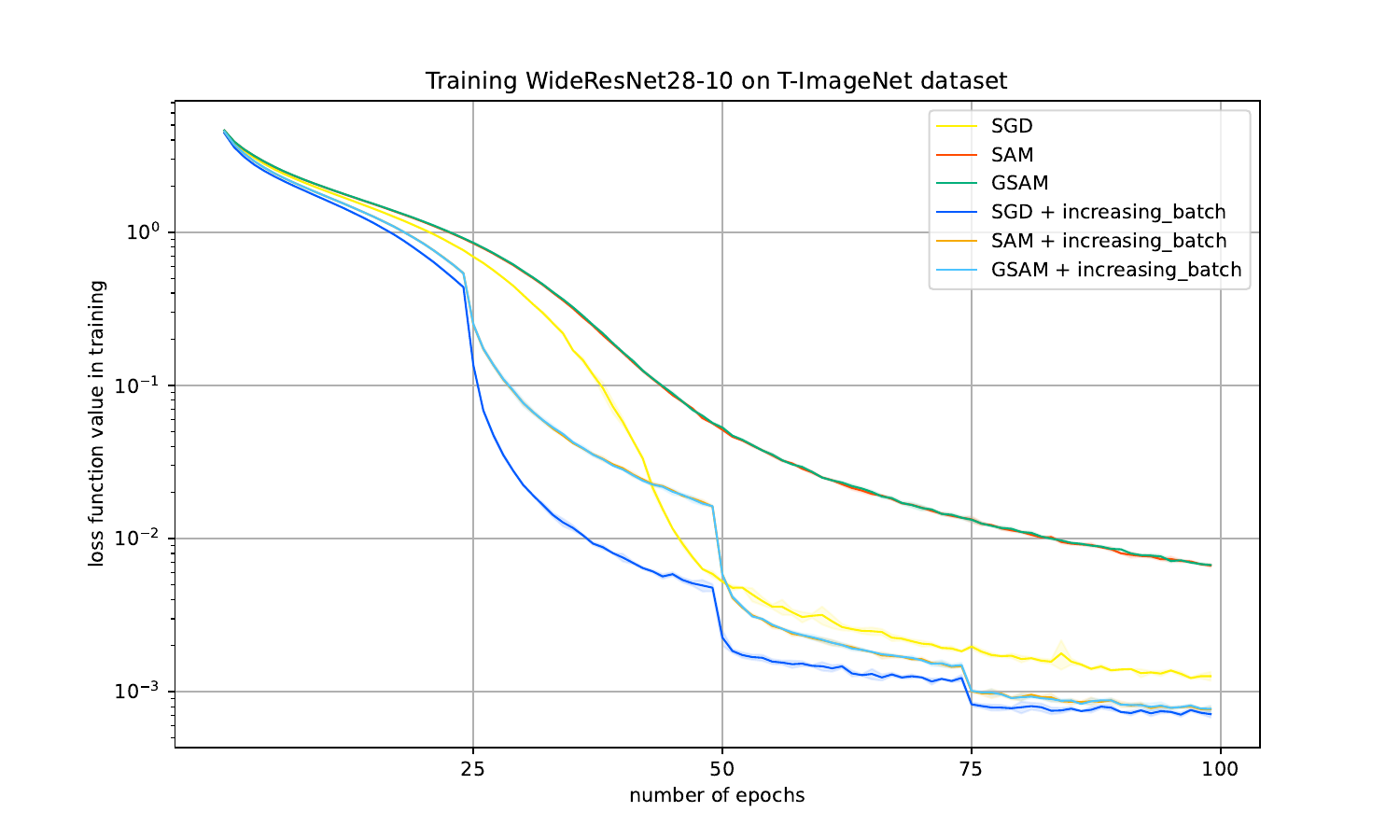}
\end{minipage} &
\begin{minipage}[t]{0.5\hsize}
\includegraphics[width=1.0\textwidth]{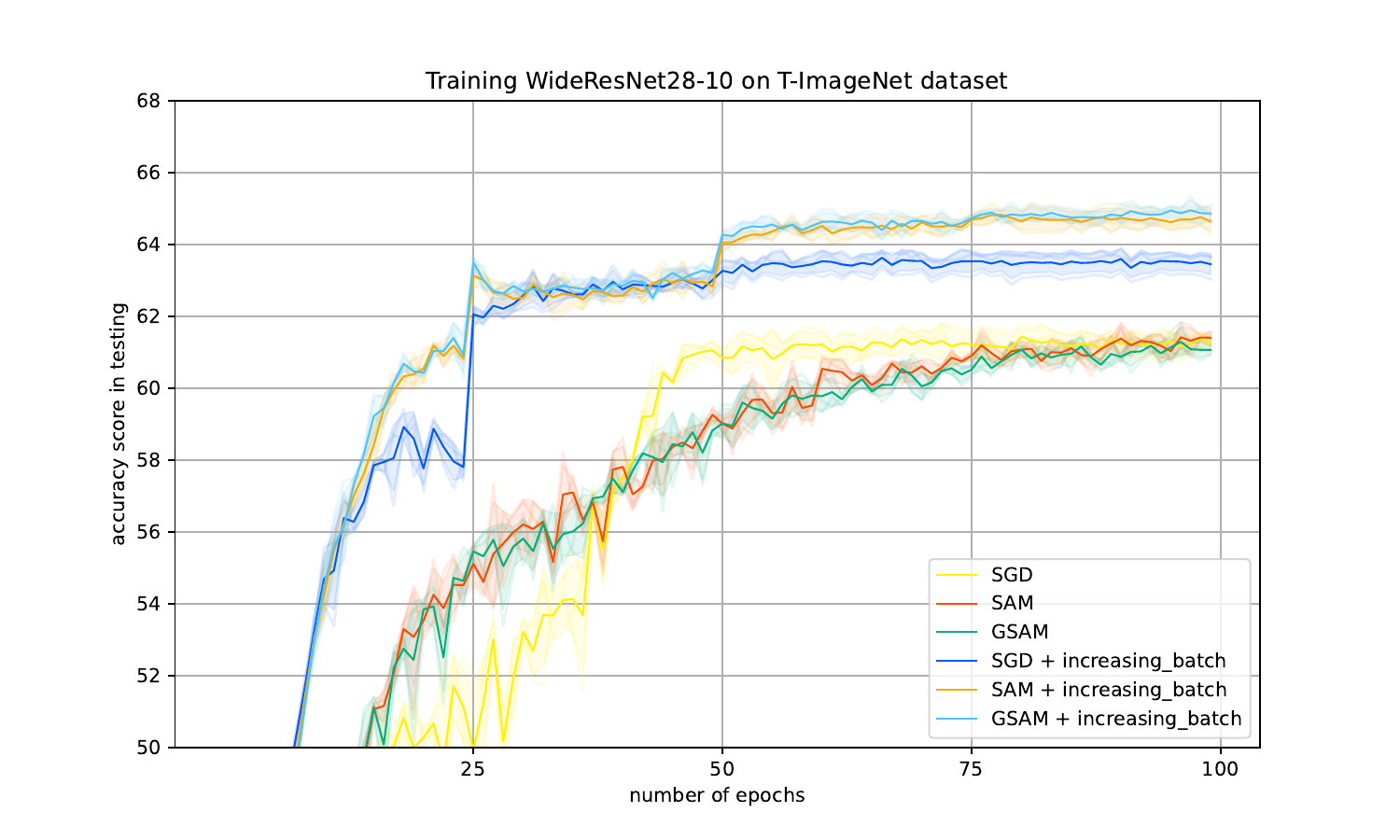}
\end{minipage}
\end{tabular}
\caption{(Left) Loss function value in training and (Right) accuracy score in testing for the algorithms versus the number of epochs in training Wide-ResNet-28-10 on the Tiny-ImageNet dataset. The learning rate of each algorithm was fixed at 0.1. In SGD/SAM/GSAM, the batch size was fixed at 128. In SGD/SAM/GSAM + increasing\_batch, the batch size was set at 16 for the first 25 epochs and then it was doubled every 25 epochs afterwards, i.e., to 32 for epochs 26-50, 64 for epochs 51-75, etc.}
\label{t-image_batch}
\end{figure*}

\begin{figure*}[ht]
\begin{tabular}{cc}
\begin{minipage}[t]{0.5\hsize}
\includegraphics[width=1.0\textwidth]{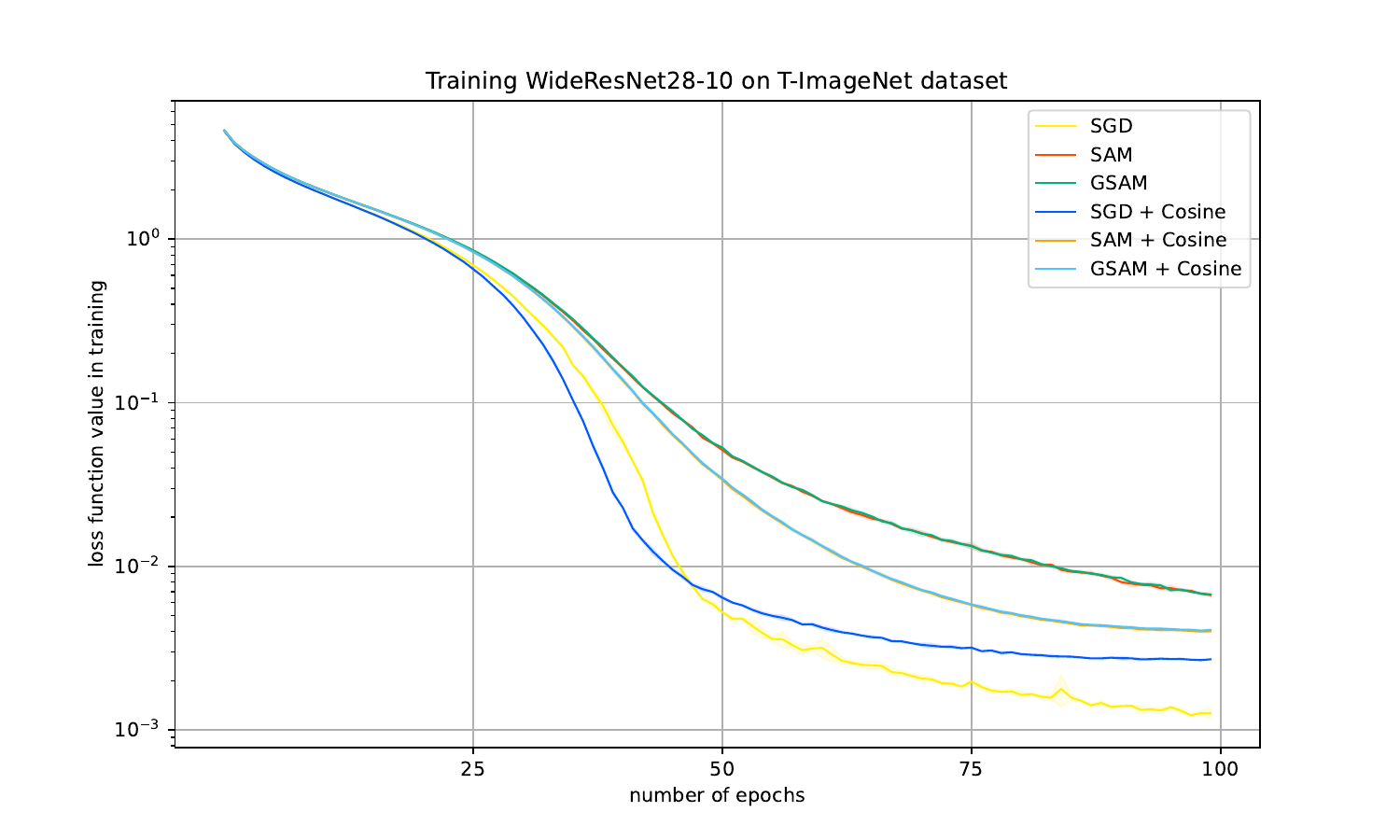}
\end{minipage} &
\begin{minipage}[t]{0.5\hsize}
\includegraphics[width=1.0\textwidth]{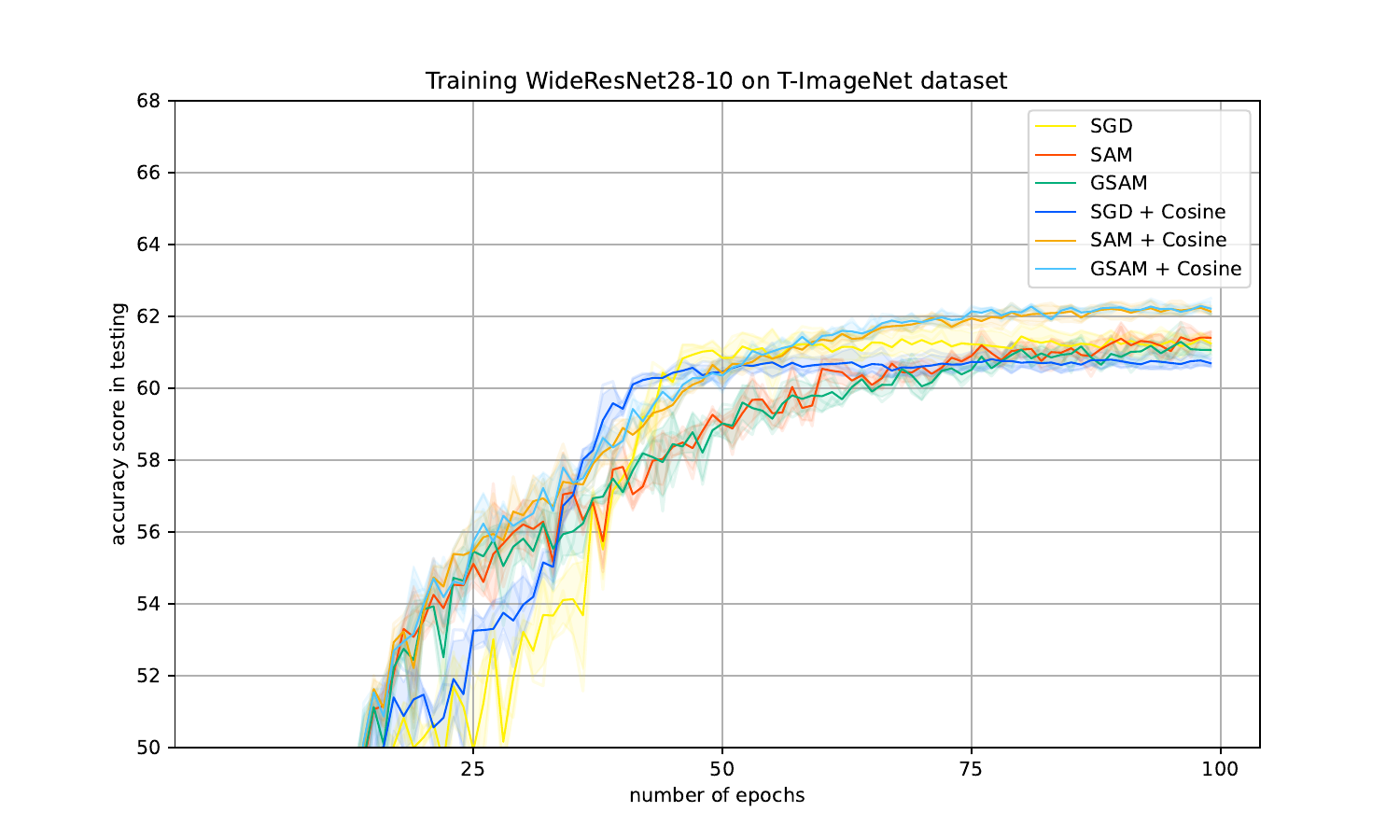}
\end{minipage}
\end{tabular}
\caption{(Left) Loss function value in training and (Right) accuracy score in testing for the algorithms versus the number of epochs in training Wide-ResNet-28-10 on the Tiny-ImageNet dataset. The batch size of each algorithm was fixed at 128. In SGD/SAM/GSAM, the constant learning rate was fixed at 0.1. In SGD/SAM/GSAM + Cosine, the maximum learning rate was 0.1 and the minimum learning rate was 0.001.}
\label{t-image_cosine}
\end{figure*}

\clearpage

{\section{Analysis of batch\_increasing}}
{In batch\_increasing, training starts with smaller batches compared with the original batch size, resulting in a simple increase in the number of iterations. In Figure \ref{fig1_1}, we provide an example where a fixed batch size with approximately the same number of iterations is prepared, demonstrating that batch\_increasing achieves a better approximate solution.}

\begin{figure}[ht]
\begin{tabular}{cc}
\begin{minipage}[t]{0.5\hsize}
\includegraphics[width=1.0\textwidth]{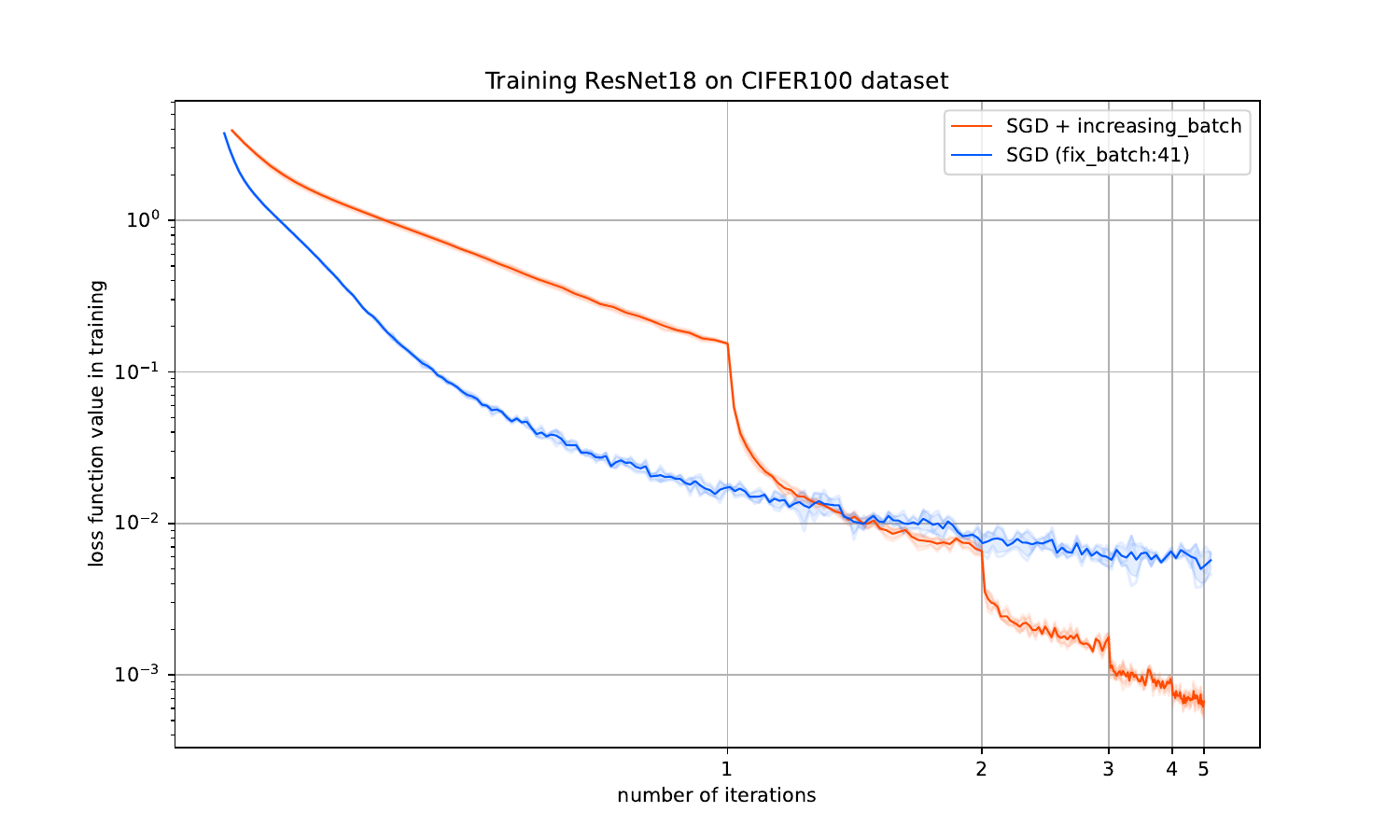}
\end{minipage} &
\begin{minipage}[t]{0.5\hsize}
\includegraphics[width=1.0\textwidth]{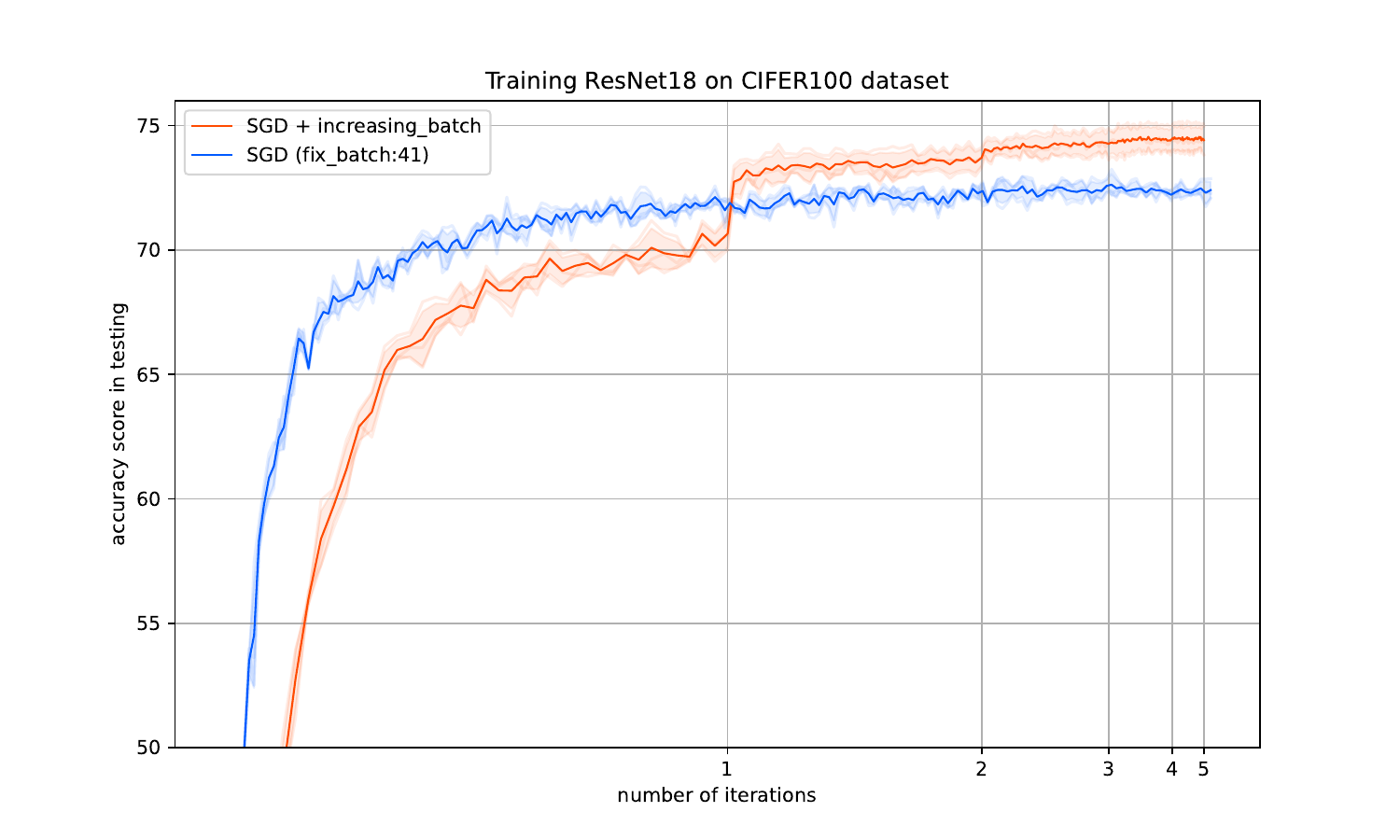}
\end{minipage}
\end{tabular}
\caption{(Left) Loss function value in training and (Right) accuracy score in testing for the batch sizes versus the number of steps in training ResNet-18 on the CIFAR-100 dataset. The learning rate for each batch size was fixed at 0.1. This is a comparison between the case of a varying batch size [16, 32, 64, 128, 256] (iteration: 242,120) and the case of a fixed batch size of 41 (iteration: 243,800).}
\label{fig1_1}
\end{figure}

\begin{table*}[ht]
\caption{Mean test error (Test Error) and worst-case $\ell_{\infty}$ adaptive sharpness (Sharpness) when training ResNet-18 from scratch on CIFAR-100 with different batch sizes.}
\label{}
\centering
\begin{tabular}{c|cccccc}
\hline
& \textbf{batch\_increasing} & \textbf{16} & \textbf{32} & \textbf{64} & \textbf{128} & \textbf{256} \\ \hline
Test Error & \textbf{25.58} & 27.14 & 27.28 & 27.11 & 26.62 & 27.06 \\ \hline
Sharpness & 1.34 &\textbf{ 0.93} & 4.28 & 46.22 & 154.27 & 329.86 \\ \hline
\end{tabular}
\end{table*}

\begin{figure}
\centering
\includegraphics[width=0.4\textwidth]{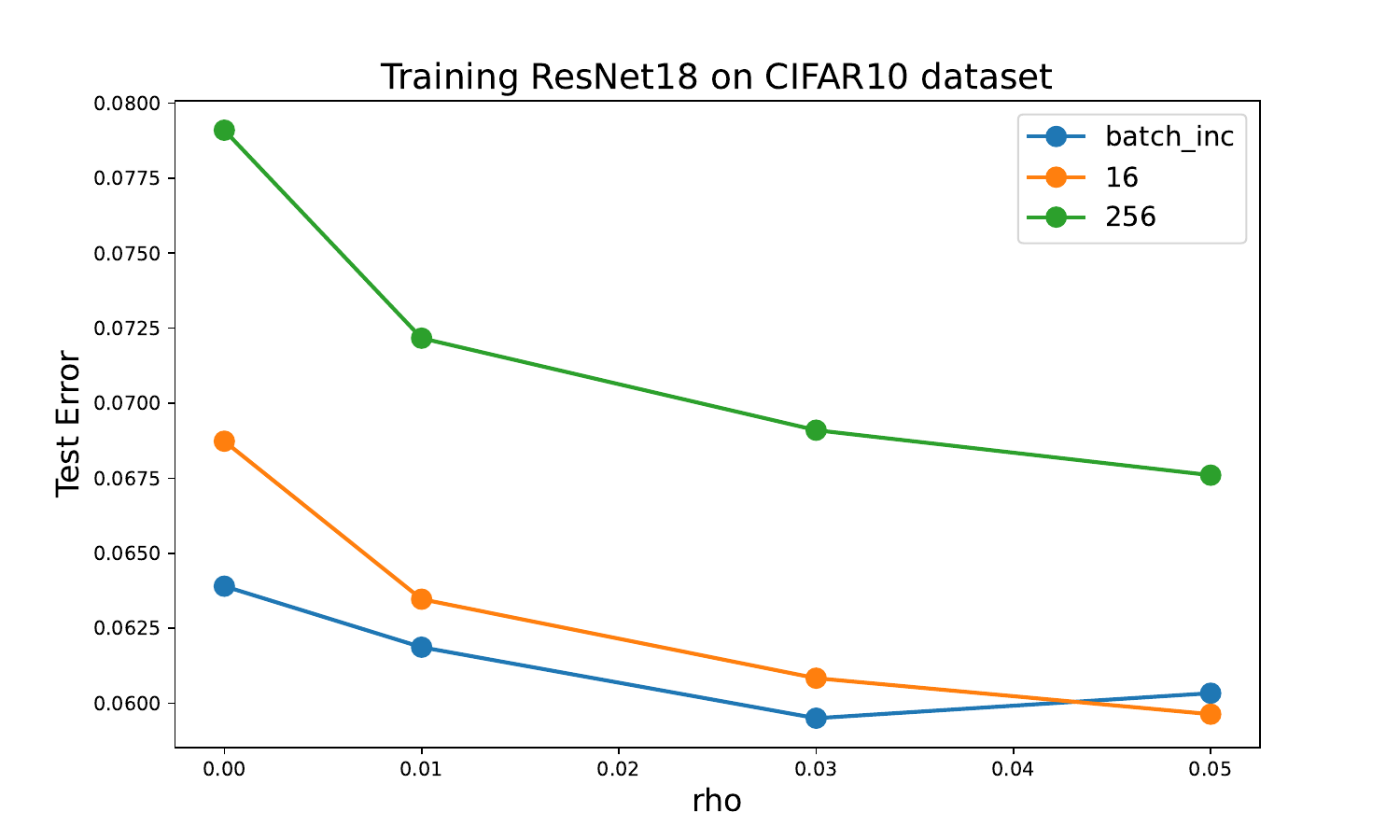}
\caption{{Test errors versus perturbation radius $\rho$ when training ResNet-18 on CIFAR-10 using SAM with a constant batch size, 16 or 256, and an increasing batch size $[16,32, 64, 128, 256]$.}}
\label{cifar10_rho}
\end{figure}

\begin{figure}
\centering
\includegraphics[width=0.4\textwidth]{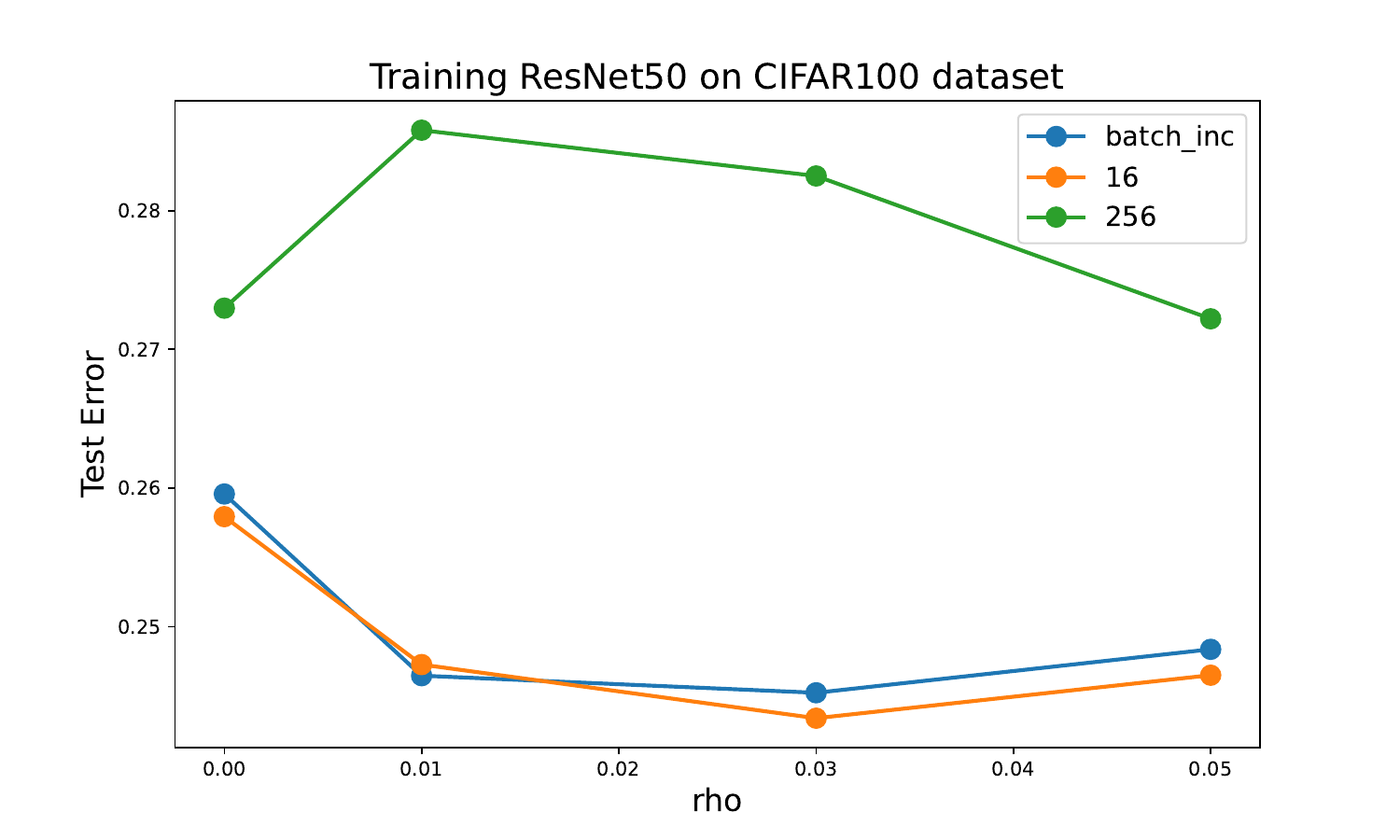}
\caption{{Test errors versus perturbation radius $\rho$ when training ResNet-50 on CIFAR-100 using SAM with constant batch size, 16 or 256, and an increasing batch size $[16,32, 64, 128, 256]$.}}
\label{cifar100_rho}
\end{figure}

{\section{Relationship between batch size and perturbation radius for SAM}}
We investigated the sensitivity to the perturbation radius $\rho$ when using constant/increasing batch sizes. Figures \ref{cifar10_rho} and \ref{cifar100_rho} compare the performances under different perturbation radii for constant batch sizes ($b = 16, 256$), and batch\_increasing (batch\_inc) $[16, 32, 64, 128, 256]$; i.e., the batch size was set to 16 for the first 40 epochs, and then it was doubled every 40 epochs. The learning rate was fixed at 0.1, and the values of $\rho$ were small, such as $\rho = 0, 0.01, 0.03, 0.05$, as discussed in Section "Convergence analysis of GSAM"\ref{sec:2.4}. As was observed in \citep{li2024friendly}, Figures \ref{cifar10_rho} and \ref{cifar100_rho} show that, for a fixed radius $\rho$, using a constant batch size $b = 256$ achieved lower test accuracies compared with using a constant batch size $b = 16$; that is, the larger a constant batch size is, the lower the effectiveness of SAM becomes. This implies that, when $\rho$ is small, SAM should use small constant batch sizes to achieve high test accuracies. Meanwhile, using increasing batch sizes (``batch\_inc" in Figures \ref{cifar10_rho} and \ref{cifar100_rho}) maintains the effectiveness of SAM. This is because they start with small batches (``batch\_inc" in Figures \ref{cifar10_rho} and \ref{cifar100_rho}). Moreover, using large batch sizes reduces computational costs to train DNNs. Therefore, we can conclude that using increasing batch sizes is useful from the viewpoints of both generalization and computational cost.

\section{Limitations}\label{limitation}
{
\paragraph{Theoretical limitations.}
Our convergence results do not hold for an arbitrary precision $\epsilon > 0$ at a fixed perturbation radius $\rho$. 
Theorems~\ref{thm:3_1} and \ref{thm:3_2} require $\epsilon$ and $\rho$ to satisfy the joint condition (\ref{eq:joint}), i.e., $\rho = O(\epsilon^4)$, 
and Theorem \ref{thm:3_2} additionally requires $b = \Omega(\epsilon^{-4})$ (see Appendix \ref{subsec:conditions}). 
Moreover, constants such as $G$, $\sigma$, $C$, and $\sum_{i\in[n]} L_i$ are difficult to estimate for deep neural networks, 
and thus we do not verify whether our experiments satisfy the conditions (\ref{eta_2})--(\ref{eta_3}), (\ref{eta_2_1})--(\ref{eta_3_1}), and (\ref{eq:joint}) numerically. 
We therefore interpret these conditions mainly as qualitative relations among the parameters, 
indicating that $\rho$ and $\eta$ should be small and $b$ should be large.
}

\paragraph{Empirical limitations.}
The experiments in this study examined a limited number of models. While they used multiple datasets, 
including ImageNet, CIFAR, and Tiny-ImageNet, they examined only ResNet18, WideResNet28-10, and ResNet50. 
Here, we primarily selected CNN-based models since our theoretical analysis is based on simple stochastic gradient descent (SGD). 
Hence, we should conduct similar experiments with a larger number of models to support our theoretical results.
{
In addition, we did not empirically examine the combined setting in which an increasing batch size
and a decaying learning rate are used simultaneously.
Although Theorem \ref{thm:3} covers this combined setting and Theorems \ref{thm:3_1} and \ref{thm:3_2} are special cases of it, verifying its practical behavior is left for future work.
Furthermore, while Algorithm \ref{algo:1} covers GSAM, SAM, and SGD in a unified manner through the choice of
$\alpha$ and $\rho$, we did not examine other SAM variants, such as those proposed in (\cite{li2023enhancing,mi2022make,du2022efficient,li2024friendly}).
Investigating whether our results extend to a wider range of SAM variants is also left for future work.
}
\end{document}

%% file: math_commands.tex
\usepackage{amsmath,amsfonts,bm}

\def\eqref#1{equation~\ref{#1}}

\def\1{\bm{1}}

\DeclareMathAlphabet{\mathsfit}{\encodingdefault}{\sfdefault}{m}{sl}
\SetMathAlphabet{\mathsfit}{bold}{\encodingdefault}{\sfdefault}{bx}{n}

\DeclareMathOperator*{\argmax}{arg\,max}

%% file: acml26.bib
@book{Freund:1971aa,
	author = {Freund, J. E.},
	isbn = {9780135622230},
	publisher = {Prentice-Hall},
	series = {Prentice-Hall mathematics series},
	title = {Mathematical Statistics},
	year = {1971}}

@inproceedings{pmlr-v48-hazanb16,
	author = {Hazan, Elad and Levy, Kfir Yehuda and Shalev-Shwartz, Shai},
	booktitle = {Proceedings of The 33rd International Conference on Machine Learning},
	pages = {1833--1841},
	publisher = {PMLR},
	series = {Proceedings of Machine Learning Research},
	title = {On Graduated Optimization for Stochastic Non-Convex Problems},
	volume = {48},
	year = {2016}}

@inproceedings{xie2021a,
	author = {Zeke Xie and Issei Sato and Masashi Sugiyama},
	booktitle = {International Conference on Learning Representations},
	title = {A Diffusion Theory For Deep Learning Dynamics: Stochastic Gradient Descent Exponentially Favors Flat Minima},
	year = {2021}}

@article{DBLP:journals/corr/abs-1903-09900,
	author = {Andrew Hundt and Varun Jain and Gregory D. Hager},
	bibsource = {dblp computer science bibliography, https://dblp.org},
	eprint = {1903.09900},
	eprinttype = {arXiv},
	journal = {CoRR},
	title = {sharp{DARTS}: {F}aster and More Accurate Differentiable Architecture Search},
	volume = {abs/1903.09900},
	year = {2019}}

@inproceedings{pmlr-v37-ioffe15,
	author = {Ioffe, Sergey and Szegedy, Christian},
	booktitle = {Proceedings of the 32nd International Conference on Machine Learning},
	pages = {448--456},
	publisher = {PMLR},
	series = {Proceedings of Machine Learning Research},
	title = {Batch Normalization: Accelerating Deep Network Training by Reducing Internal Covariate Shift},
	volume = {37},
	year = {2015}}

@inproceedings{6952943,
	author = {Wu, Yuting and Holland, Daniel J. and Mantle, Mick D. and Wilson, Andrew G. and Nowozin, Sebastian and Blake, Andrew and Gladden, Lynn F.},
	booktitle = {2014 22nd European Signal Processing Conference (EUSIPCO)},
	pages = {2515-2519},
	title = {A {B}ayesian method to quantifying chemical composition using {NMR}: {A}pplication to porous media systems},
	year = {2014}}

@inproceedings{pmlr-v202-andriushchenko23a,
	author = {Andriushchenko, Maksym and Croce, Francesco and M\"{u}ller, Maximilian and Hein, Matthias and Flammarion, Nicolas},
	booktitle = {Proceedings of the 40th International Conference on Machine Learning},
	month = {23--29 Jul},
	pages = {840--902},
	publisher = {PMLR},
	series = {Proceedings of Machine Learning Research},
	title = {A Modern Look at the Relationship between Sharpness and Generalization},
	volume = {202},
	year = {2023}}

@inproceedings{Jiang2020Fantastic,
	author = {Yiding Jiang and Behnam Neyshabur and Hossein Mobahi and Dilip Krishnan and Samy Bengio},
	booktitle = {International Conference on Learning Representations},
	title = {Fantastic Generalization Measures and Where to Find Them},
	year = {2020}}

@inproceedings{keskar2017on,
	author = {Nitish Shirish Keskar and Dheevatsa Mudigere and Jorge Nocedal and Mikhail Smelyanskiy and Ping Tak Peter Tang},
	booktitle = {International Conference on Learning Representations},
	title = {On Large-Batch Training for Deep Learning: Generalization Gap and Sharp Minima},
	year = {2017}}

@inproceedings{You2020Large,
	author = {Yang You and Jing Li and Sashank Reddi and Jonathan Hseu and Sanjiv Kumar and Srinadh Bhojanapalli and Xiaodan Song and James Demmel and Kurt Keutzer and Cho-Jui Hsieh},
	booktitle = {International Conference on Learning Representations},
	title = {Large Batch Optimization for Deep Learning: Training BERT in 76 minutes},
	year = {2020}}

@inproceedings{NIPS2017_a5e0ff62,
	author = {Hoffer, Elad and Hubara, Itay and Soudry, Daniel},
	booktitle = {Advances in Neural Information Processing Systems},
	title = {Train longer, generalize better: closing the generalization gap in large batch training of neural networks},
	volume = {30},
	year = {2017}}

@misc{goyal2018accuratelargeminibatchsgd,
	archiveprefix = {arXiv},
	author = {Priya Goyal and Piotr Doll{\'a}r and Ross Girshick and Pieter Noordhuis and Lukasz Wesolowski and Aapo Kyrola and Andrew Tulloch and Yangqing Jia and Kaiming He},
	eprint = {1706.02677},
	primaryclass = {cs.CV},
	title={Accurate, Large Minibatch {SGD}: Training {ImageNet} in 1 Hour},
	year = {2018}}

@inproceedings{springer2024sharpnessaware,
	author = {Jacob Mitchell Springer and Vaishnavh Nagarajan and Aditi Raghunathan},
	booktitle = {The Twelfth International Conference on Learning Representations},
	title = {Sharpness-Aware Minimization Enhances Feature Quality via Balanced Learning},
	year = {2024}}

@inproceedings{mollenhoff2023sam,
	author = {Thomas M{\"o}llenhoff and Mohammad Emtiyaz Khan},
	booktitle = {The Eleventh International Conference on Learning Representations},
	title = {{SAM} as an Optimal Relaxation of Bayes},
	year = {2023}}

@inproceedings{chen2023why,
	author = {Zixiang Chen and Junkai Zhang and Yiwen Kou and Xiangning Chen and Cho-Jui Hsieh and Quanquan Gu},
	booktitle = {Thirty-seventh Conference on Neural Information Processing Systems},
	title = {Why Does Sharpness-Aware Minimization Generalize Better Than {SGD}?},
	year = {2023}}

@inproceedings{wen2023how,
	author = {Kaiyue Wen and Tengyu Ma and Zhiyuan Li},
	booktitle = {The Eleventh International Conference on Learning Representations},
	title = {How Sharpness-Aware Minimization Minimizes Sharpness?},
	year = {2023}}

@inproceedings{andriushchenko2023sharpnessaware,
	author = {Maksym Andriushchenko and Dara Bahri and Hossein Mobahi and Nicolas Flammarion},
	booktitle = {Thirty-seventh Conference on Neural Information Processing Systems},
	title = {Sharpness-Aware Minimization Leads to Low-Rank Features},
	year = {2023}}

@inproceedings{du2022efficient,
	author = {Jiawei Du and Hanshu Yan and Jiashi Feng and Joey Tianyi Zhou and Liangli Zhen and Rick Siow Mong Goh and Vincent Tan},
	booktitle = {International Conference on Learning Representations},
	title = {Efficient Sharpness-aware Minimization for Improved Training of Neural Networks},
	year = {2022}}

@inproceedings{sherborne2024tram,
	author = {Tom Sherborne and Naomi Saphra and Pradeep Dasigi and Hao Peng},
	booktitle = {The Twelfth International Conference on Learning Representations},
	title = {{TRAM}: Bridging Trust Regions and Sharpness Aware Minimization},
	year = {2024}}

@inproceedings{wang2024efficient,
	author = {Yili Wang and Kaixiong Zhou and Ninghao Liu and Ying Wang and Xin Wang},
	booktitle = {The Twelfth International Conference on Learning Representations},
	title = {Efficient Sharpness-Aware Minimization for Molecular Graph Transformer Models},
	year = {2024}}

@inproceedings{chen2022when,
	author = {Xiangning Chen and Cho-Jui Hsieh and Boqing Gong},
	booktitle = {International Conference on Learning Representations},
	title = {When Vision Transformers Outperform ResNets without Pre-training or Strong Data Augmentations},
	year = {2022}}

@inproceedings{Liu2020On,
	author = {Liyuan Liu and Haoming Jiang and Pengcheng He and Weizhu Chen and Xiaodong Liu and Jianfeng Gao and Jiawei Han},
	booktitle = {International Conference on Learning Representations},
	title = {On the Variance of the Adaptive Learning Rate and Beyond},
	year = {2020}}

@inproceedings{pmlr-v54-de17a,
	author = {De, Soham and Yadav, Abhay and Jacobs, David and Goldstein, Tom},
	booktitle = {Proceedings of the 20th International Conference on Artificial Intelligence and Statistics},
	pages = {1504--1513},
	publisher = {PMLR},
	series = {Proceedings of Machine Learning Research},
	title = {{Automated Inference with Adaptive Batches}},
	volume = {54},
	year = {2017}}

@misc{balles2016coupling,
	author = {Balles, Lukas and Romero, Javier and Hennig, Philipp},
	description = {Coupling Adaptive Batch Sizes with Learning Rates},
	note = {Thirty-Third Conference on Uncertainty in Artificial Intelligence},
	title = {Coupling Adaptive Batch Sizes with Learning Rates},
	year = 2017}

@article{Byrd:2012aa,
	author = {Byrd, Richard H. and Chin, Gillian M. and Nocedal, Jorge and Wu, Yuchen},
	date = {2012/08/01},
	id = {Byrd2012},
	isbn = {1436-4646},
	journal = {Mathematical Programming},
	number = {1},
	pages = {127--155},
	title = {Sample size selection in optimization methods for machine learning},
	volume = {134},
	year = {2012}}

@inproceedings{li2023enhancing,
	author = {Bingcong Li and Georgios B. Giannakis},
	booktitle = {Thirty-seventh Conference on Neural Information Processing Systems},
	title = {Enhancing Sharpness-Aware Optimization Through Variance Suppression},
	year = {2023}}

@inproceedings{li2024friendly,
	author = {Li, Tao and Zhou, Pan and He, Zhengbao and Cheng, Xinwen and Huang, Xiaolin},
	booktitle = {Proceedings of the IEEE/CVF Conference on Computer Vision and Pattern Recognition},
	title = {Friendly Sharpness-Aware Minimization},
	year = {2024}}

@inproceedings{si2023practical,
	author = {Dongkuk Si and Chulhee Yun},
	booktitle = {Thirty-seventh Conference on Neural Information Processing Systems},
	title = {Practical Sharpness-Aware Minimization Cannot Converge All the Way to Optima},
	year = {2023}}

@inproceedings{mi2022make,
	author = {Peng Mi and Li Shen and Tianhe Ren and Yiyi Zhou and Xiaoshuai Sun and Rongrong Ji and Dacheng Tao},
	booktitle = {Advances in Neural Information Processing Systems},
	title = {Make Sharpness-Aware Minimization Stronger: A Sparsified Perturbation Approach},
	year = {2022}}

@inproceedings{pmlr-v162-andriushchenko22a,
	author = {Andriushchenko, Maksym and Flammarion, Nicolas},
	booktitle = {Proceedings of the 39th International Conference on Machine Learning},
	month = {17--23 Jul},
	pages = {639--668},
	publisher = {PMLR},
	series = {Proceedings of Machine Learning Research},
	title = {Towards Understanding Sharpness-Aware Minimization},
	volume = {162},
	year = {2022}}

@book{doi:10.1137/1.9780898719468,
	author = {Ortega, J. M. and Rheinboldt, W. C.},
	eprint = {https://epubs.siam.org/doi/pdf/10.1137/1.9780898719468},
	publisher = {Society for Industrial and Applied Mathematics},
	title = {Iterative Solution of Nonlinear Equations in Several Variables},
	year = {2000}}

@misc{sato2023using,
	archiveprefix = {arXiv},
	author = {Naoki Sato and Hideaki Iiduka},
	eprint = {2311.08745},
	primaryclass = {id='cs.LG' full_name='Machine Learning' is_active=True alt_name=None in_archive='cs' is_general=False description='Papers on all aspects of machine learning research (supervised, unsupervised, reinforcement learning, bandit problems, and so on) including also robustness, explanation, fairness, and methodology. cs.LG is also an appropriate primary category for applications of machine learning methods.'},
	title = {Using Stochastic Gradient Descent to Smooth Nonconvex Functions: Analysis of Implicit Graduated Optimization with Optimal Noise Scheduling},
	year = {2023},
    url={https://arxiv.org/abs/2311.08745} }

@inproceedings{loshchilov2017sgdr,
	author = {Ilya Loshchilov and Frank Hutter},
	booktitle = {International Conference on Learning Representations},
	title = {{SGDR}: Stochastic Gradient Descent with Warm Restarts},
	year = {2017}}

@inproceedings{foret2021sharpnessaware,
	author = {Pierre Foret and Ariel Kleiner and Hossein Mobahi and Behnam Neyshabur},
	booktitle = {International Conference on Learning Representations},
	title = {Sharpness-aware Minimization for Efficiently Improving Generalization},
	year = {2021}}

@inproceedings{zhuang2022surrogate,
	author = {Juntang Zhuang and Boqing Gong and Liangzhe Yuan and Yin Cui and Hartwig Adam and Nicha C Dvornek and sekhar tatikonda and James s Duncan and Ting Liu},
	booktitle = {International Conference on Learning Representations},
	title = {Surrogate Gap Minimization Improves Sharpness-Aware Training},
	year = {2022}}

@book{beck2017,
	address = {Philadelphia, PA},
	author = {Beck, Amir},
	doi = {10.1137/1.9781611974997},
	eprint = {https://epubs.siam.org/doi/pdf/10.1137/1.9781611974997},
	publisher = {Society for Industrial and Applied Mathematics},
	title = {First-Order Methods in Optimization},
	year = {2017}}

@inproceedings{wang2021on,
	author = {Xiaoyu Wang and Sindri Magn{\'u}sson and Mikael Johansson},
	booktitle = {Advances in Neural Information Processing Systems},
	title = {On the Convergence of Step Decay Step-Size for Stochastic Optimization},
	year = {2021}}

@article{Ghadimi:2016aa,
	author = {Ghadimi, Saeed and Lan, Guanghui and Zhang, Hongchao},
	date = {2016/01/01},
	id = {Ghadimi2016},
	isbn = {1436-4646},
	journal = {Mathematical Programming},
	number = {1},
	pages = {267--305},
	title = {Mini-batch stochastic approximation methods for nonconvex stochastic composite optimization},
	volume = {155},
	year = {2016}}

@article{doi:10.1137/120880811,
	author = {Ghadimi, Saeed and Lan, Guanghui},
	eprint = {https://doi.org/10.1137/120880811},
	journal = {SIAM Journal on Optimization},
	number = {4},
	pages = {2341-2368},
	title = {Stochastic First- and Zeroth-Order Methods for Nonconvex Stochastic Programming},
	volume = {23},
	year = {2013}}

@inproceedings{l.2018dont,
	author = {Samuel L. Smith and Pieter-Jan Kindermans and Quoc V. Le},
	booktitle = {International Conference on Learning Representations},
	title = {Don't Decay the Learning Rate, Increase the Batch Size},
	year = {2018}}

@article{khaled2022better,
	author = {Ahmed Khaled and Peter Richt{\'a}rik},
	journal = {Transactions on Machine Learning Research},
	title = {Better Theory for {SGD} in the Nonconvex World},
	year = {2023}}

@article{Arjevani:2023aa,
	author = {Arjevani, Yossi and Carmon, Yair and Duchi, John C. and Foster, Dylan J. and Srebro, Nathan and Woodworth, Blake},
	journal = {Mathematical Programming},
	number = {1},
	pages = {165--214},
	title = {Lower bounds for non-convex stochastic optimization},
	volume = {199},
	year = {2023}}

@inproceedings{NEURIPS2019_2557911c,
	author = {Vaswani, Sharan and Mishkin, Aaron and Laradji, Issam and Schmidt, Mark and Gidel, Gauthier and Lacoste-Julien, Simon},
	booktitle = {Advances in Neural Information Processing Systems},
	title = {Painless Stochastic Gradient: {I}nterpolation, Line-Search, and Convergence Rates},
	volume = {32},
	year = {2019}}

@conference{loizou2021,
	author = {Loizou, Nicolas and Vaswani, Sharan and Laradji, Issam and Lacoste-Julien, Simon},
	booktitle = {Proceedings of the 24th International Conference on Artificial Intelligence and Statistics},
	title = {Stochastic Polyak Step-size for {SGD}: {A}n Adaptive Learning Rate for Fast Convergence},
	volume = {130},
	year = {2021}}

@conference{chen2020,
	author = {Chen, Hao and Zheng, Lili and AL Kontar, Raed and Raskutti, Garvesh},
	booktitle = {Advances in Neural Information Processing Systems},
	title = {Stochastic Gradient Descent in Correlated Settings: {A} Study on {G}aussian Processes},
	volume = {33},
	year = {2020}}

@conference{sca2020,
	author = {Scaman, Kevin and Malherbe, C\'edric},
	booktitle = {Advances in Neural Information Processing Systems},
	title = {Robustness Analysis of Non-Convex Stochastic Gradient Descent using Biased Expectations},
	volume = {33},
	year = {2020}}

@article{feh2020,
	author = {Fehrman, Benjamin and Gess, Benjamin and Jentzen, Arnulf},
	journal = {Journal of Machine Learning Research},
	pages = {1--48},
	title = {Convergence Rates for the Stochastic Gradient Descent Method for Non-Convex Objective Functions},
	volume = {21},
	year = {2020}}

@article{robb1951,
	author = {Robbins, Herbert and Monro, Herbert},
	journal = {The Annals of Mathematical Statistics},
	pages = {400--407},
	title = {A Stochastic Approximation Method},
	volume = {22},
	year = {1951}}

@techreport{krizhevsky2009learning,
  title={Learning multiple layers of features from tiny images},
  author={Krizhevsky, Alex},
  year={2009},
  institution={University of Toronto}
}

@article{ILSVRC15, 
    author = {Olga Russakovsky and Jia Deng and Hao Su and Jonathan Krause and Sanjeev Satheesh and Sean Ma and Zhiheng Huang and Andrej Karpathy and Aditya Khosla and Michael Bernstein and Alexander C. Berg and Li Fei-Fei}, 
    title = {{ImageNet Large Scale Visual Recognition Challenge}}, year = {2015},
    journal = {International Journal of Computer Vision (IJCV)}, doi = {10.1007/s11263-015-0816-y},
    volume={115},
    number={3},
    pages={211-252} }

@article{Le2015TinyIV,
    title={Tiny {ImageNet} visual recognition challenge},
    author={Le, Y. and Yang, X.},
    journal={CS 231N},
    volume={7},
    number={7},
    pages={3},
    year={2015}
}
